%% file: main.tex
\documentclass[conference]{IEEEtran}
\IEEEoverridecommandlockouts
\usepackage{cite}
\usepackage{amsmath,amssymb,amsfonts}
\usepackage{algpseudocode}
\usepackage{algorithm}
\usepackage{graphicx}
\usepackage{textcomp}
\usepackage{xcolor}
\usepackage{caption}
\usepackage{subcaption}
\usepackage{hyperref}
\hypersetup{colorlinks, linkcolor=blue, citecolor=cyan}
\usepackage{glossaries}
\usepackage[symbols]{glossaries-extra}
\loadglsentries{glossary}
\makenoidxglossaries
\def\BibTeX{{\rm B\kern-.05em{\sc i\kern-.025em b}\kern-.08em
    T\kern-.1667em\lower.7ex\hbox{E}\kern-.125emX}}
\begin{document}

\title{
\centering\LARGE{\textbf{Multi Object Tracking for Predictive Collision Avoidance}}\\}
\author{\IEEEauthorblockN{Bruk Gebregziabher}
\IEEEauthorblockA{
\textit{University of Zagreb}\\
Zagreb, Croatia \\
bruk.gebregziabher@fer.hr}
\and
\IEEEauthorblockN{Hadush Hailu}
\IEEEauthorblockA{
\textit{Maharishi International University}\\
IA, USA \\
hadush.gebrerufael@miu.edu}
}

\maketitle
\begin{abstract}
The safe and efficient operation of Autonomous Mobile Robots (AMRs) in complex environments, such as manufacturing, logistics, and agriculture, necessitates accurate multi-object tracking and predictive collision avoidance. This paper presents algorithms and techniques for addressing these challenges using Lidar sensor data, emphasizing ensemble Kalman filter.

The developed predictive collision avoidance algorithm employs the data provided by lidar sensors to track multiple objects and predict their velocities and future positions, enabling the AMR to navigate safely and effectively. A modification to the dynamic windowing approach is introduced to enhance the performance of the collision avoidance system. The overall system architecture encompasses object detection, multi-object tracking, and predictive collision avoidance control.

The experimental results, obtained from both simulation and real-world data, demonstrate the effectiveness of the proposed methods in various scenarios, which lays the foundation for future research on global planners, other controllers, and the integration of additional sensors. This thesis contributes to the ongoing development of safe and efficient autonomous systems in complex and dynamic environments.
\end{abstract}

\noindent\rule{7cm}{0.2pt}
\input{sections/intro}
\input{sections/state_of_art}
\input{sections/methodology}

\input{sections/system_architecture}
\input{sections/experiment_setup}
\input{sections/results}
\input{sections/limits}
\input{sections/conclusion}


\bibliographystyle{IEEEtran}

\bibliography{bibliography}
\printnoidxglossary[type=\acronymtype, title=Acronyms]
\input{sections/appendex}


\end{document}

%% file: glossary.tex

\newacronym{kf}{KF}{Kalman Filter}
\newacronym{enkf}{EnKF}{Ensemble Kalman Filter}
\newacronym{agv}{AGV}{Automated Guided Vehicle}
\newacronym{amr}{AMR}{Autonomous Mobile Robot}
\newacronym{cad}{CAD}{Computer Aided Design}
\newacronym{dwa}{DWA}{Dynamic Windowing Approach}
\newacronym{jpda}{JPDA}{Joint Probabilistic Data Association}
\newacronym{icr}{ICR}{Instantaneous Center of Rotation}
\newacronym{mot}{MOT}{Multi Object Tracking}
\newacronym{gnn}{GNN}{Global Neighbor Data Association}
\newacronym{nnda}{NNDA}{Nearest Neighbor Data Association}
\newacronym{pca}{PCA}{Predictive Collision Avoidance}
\newacronym{pcl}{PCL}{Point Cloud Library}
\newacronym{plc}{PLC}{Programmable Logic Controller}
\newacronym{ros2}{ROS 2}{Robot Operating System 2}
\newacronym{stp}{STEP}{Standard for Exchange of Product Model Data}
\newacronym{ttc}{TTC}{Time to Collision }
\newacronym{udp}{UDP}{User Datagram Protocol }
\newacronym{urdf}{URDF}{Unified Robot Description Format}


%% file: sections/intro.tex
\section{Introduction}

\subsection{Background and Motivation}
Collision and obstacle avoidance are critical aspects of the design and operation of autonomous mobile robots. In order to effectively navigate their environment, these robots must be able to detect and avoid collisions with both stationary and moving objects. One key component of an effective collision and obstacle avoidance system is the use of sensors, such as lidar, radar, and ultrasonic sensors, which enable the robot to create a detailed map of its surroundings and detect the presence of obstacles. The ability to track multiple objects and predict potential collisions is a critical aspect of autonomous systems, such as self-driving cars, drones, and robots in industrial settings. The increasing demand for safe and efficient automation has led to a growing interest in the development of robust and reliable multi-object tracking and predictive collision avoidance techniques. These algorithms should be able to handle complex and dynamic environments while ensuring the safety and efficiency of the autonomous system.

\subsection{Objectives}
The main objective of this thesis is to explore methods for multi-object tracking (MOT) and predictive collision avoidance (PCA) that can significantly improve the safety and efficiency of autonomous mobile robots, specifically focusing on industrial \acrshort{agv}s (Automated Guided Vehicles) such as forklifts. This research focuses on the development and evaluation of multi-object tracking and predictive collision avoidance algorithms for autonomous systems operating in dynamic environments. The study will consider both simulated and real-world scenarios, with a focus on industrial settings.

\subsection{Challenges and Limitations}
The primary challenges of multi-object tracking and predictive collision avoidance include:

\begin{enumerate}
    \item Developing algorithms that effectively avoid potential collisions while maintaining a certain level of task accomplishment and time optimization in complex environments.
    \item Ensuring the performance, accuracy, and computational efficiency of the developed methods.
    \item Addressing the heterogeneous nature of robots and humans, which adds more uncertainty to the algorithms.
    \item Adapting the proposed algorithms to diverse environments and scenarios in real-world deployments.
    \item Effectively using the information obtained from lidar sensors, which are the primary sensing modality in this research, to track objects and predict potential collisions.
\end{enumerate}

The limitations of this research include:

\begin{enumerate}
    \item The assumptions made regarding the dynamic behavior of objects in the environment, may not accurately reflect real-world situations.
    \item The potential challenges related to the implementation of the proposed algorithms on real-world systems, such as ensuring robustness to sensor noise and computational efficiency.
    \item The reliance on lidar sensors for object tracking and collision avoidance, which might not be as effective in certain environments or under certain conditions, such as in low-visibility situations or when faced with highly reflective surfaces.
\end{enumerate}

\subsection{Contributions}
The primary contributions of this thesis include a comprehensive review of existing approaches for multi-object tracking and predictive collision avoidance, the development of a novel predictive collision avoidance algorithm, the design and implementation of an innovative \acrshort{pca} system, the evaluation of the proposed \acrshort{mot} and \acrshort{pca} techniques through extensive simulations and real-world experiments, and the identification of future research directions and potential applications of the developed methods.

\autoref{sec:stateart} presents a literature review of the current state of research in multi-object tracking and predictive collision avoidance, including a discussion of the various algorithms and techniques employed in the field. \autoref{sec:methods} details the proposed predictive collision avoidance algorithm, its theoretical foundations, and implementation details. \autoref{sec:architecture} introduces the overall system architecture, which incorporates multi-object tracking and predictive collision avoidance, along with the use of ROS2 and UDP communication with the forklift. \autoref{sec:setup} discusses the experimental setup, including the use of a Gazebo and stage environment for simulation and real-world data from a forklift in an industrial environment, as well as the use of Blender with Phobos to convert CAD files to URDF for the forklift model. \autoref{sec:results} presents the results and discussion of the proposed methods, highlighting the effectiveness of the proposed algorithms in various scenarios. \autoref{sec:limits} discusses the limitations of the study and suggests future research directions, including potential improvements and extensions to the developed methods. \autoref{sec:conclusion} concludes the thesis, summarizing the contributions and providing an outlook on the wider implications of the research.

%% file: sections/state_of_art.tex
\section{Literature Review and Related Works}
\label{sec:stateart}
The field of \glsxtrshort{mot} and \glsxtrshort{pca} has evolved rapidly in the last decade, with numerous advancements in various application areas, such as autonomous vehicles, robotics, surveillance, and traffic management. This section provides an overview of the state-of-the-art methods and technologies in this area, focusing on the key milestones achieved in recent years and the current trends shaping the future of the field.

\subsection{Literature Review}
In addition to classical collision avoidance methods predictive collision avoidance has also been studied extensively, with a focus on integrating object-tracking algorithms with motion prediction models and decision-making algorithms. These approaches have demonstrated good performance in a variety of scenarios, but there is still room for improvement in terms of accuracy and real-time performance. 
\subsubsection{Multi-Object Tracking Techniques}
\paragraph{Data Association Approaches}
Global Neighbor Data Association (GNN): The \glsxtrshort{gnn} method is a popular data association technique and has become a foundation for many MOT approaches. \glsxtrshort{gnn} minimizes the overall distance between detections in consecutive frames by solving the assignment problem using the Hungarian algorithm \cite{kuhn1955hungarian}. Although \glsxtrshort{gnn} is computationally efficient for a smaller number of objects, it can suffer from incorrect associations in complex scenarios, such as occlusions or closely spaced objects.

Joint Probabilistic Data Association (JPDA): To address the limitations of \glsxtrshort{gnn}, Bar-Shalom et al. proposed \glsxtrshort{jpda} \cite{jpda}, which computes the posterior probability of each association hypothesis. By considering all possible associations within a specified search region, \glsxtrshort{jpda} can handle ambiguous situations better than \glsxtrshort{gnn}. However, \glsxtrshort{jpda}'s computational complexity increases with the number of objects and the uncertainty in the measurements.

Multiple Hypothesis Tracking (MHT): Reid introduced MHT to better handle complex scenarios \cite{reid}. MHT maintains multiple association hypotheses simultaneously. MHT builds a tree of possible association hypotheses and prunes low-probability branches to manage computational complexity. While MHT has shown superior performance in handling occlusions and closely spaced objects, its computational complexity remains a challenge, especially for real-time applications.

Deep SORT: Wojke et al. combined deep learning techniques with data association methods to develop Deep SORT \cite{deepsort}, a significant advancement in MOT. Deep SORT uses a CNN-based appearance descriptor to extract features from object detections and employs a Kalman filter for motion prediction. The Hungarian algorithm is then applied for data association based on appearance and motion similarity. Deep SORT has demonstrated improved performance in handling appearance changes and occlusions compared to traditional methods. However, it may struggle in the presence of long-term occlusions and rapid motion changes.

MOTNet: To further improve MOT performance, Sadeghian et al. proposed MOTNet \cite{motnet}, which utilizes a Siamese CNN architecture to model appearance similarity between detections. MOTNet also incorporates an RNN-based motion model to enforce temporal consistency, making it more robust to missed detections and false positives. Although MOTNet has achieved impressive results, it requires extensive training data and can be computationally expensive for large-scale applications.

\paragraph{End-to-End Deep Learning MOT Techniques} 
Recent advancements in deep learning have led to the development of end-to-end MOT techniques, such as FairMOT \cite{fairmot} and Tracktor++ \cite{tracktor}. These methods leverage the power of deep learning to jointly learn object detection, feature extraction, and data association in a single network, leading to more accurate and efficient tracking. FairMOT, for instance, uses an anchor-free object detection network combined with a re-identification head to learn both appearance and motion information simultaneously. Tracktor++ builds upon the existing object detection frameworks like Faster R-CNN and incorporates an appearance embedding model to improve tracking performance. These end-to-end approaches have demonstrated significant improvements in tracking accuracy and real-time performance, making them promising candidates for practical applications.

\subsubsection{Predictive Collision Avoidance}
\paragraph{Model-Based Approaches}
Rapidly-exploring Random Trees (RRT): LaValle (1998) introduced RRT \cite{rrt}, a widely-used motion planning algorithm. RRT incrementally constructs a tree of potential trajectories while considering dynamic constraints, allowing for efficient exploration of the search space. RRT has been successfully applied to various robotic and autonomous vehicle applications. However, it can suffer from sub-optimal solutions and may struggle to find feasible paths in high-dimensional or constrained environments.

Model Predictive Control (MPC): MPC \cite{mpc} is an optimization-based method that iteratively solves a constrained optimization problem over a finite horizon to find an optimal control input sequence. MPC has been successfully applied to collision avoidance in autonomous vehicles and robotics, offering the advantage of explicitly considering constraints and performance objectives. However, MPC relies on accurate models of the environment and the surrounding objects, making it sensitive to modeling errors and uncertainties.

Interaction-Aware Moving Target Model Predictive Control (MTMPC): Zhou et al. (2022) proposed an interaction-aware moving target model predictive control (MTMPC)\cite{9838002} approach for motion planning of autonomous vehicles. This method takes into consideration the interactions between the ego vehicle and other traffic participants, enabling more accurate predictions of their future motion. By incorporating the anticipated motion of other vehicles, the MTMPC can generate collision-free trajectories that account for the dynamic nature of the surrounding environment. This approach has shown promising results in terms of safety and efficiency, as it allows the ego vehicle to adapt its motion plan based on the predicted behavior of other vehicles on the road.

Priority-based collision avoidance: Cai et al. (2007)~\cite{4304002} developed an algorithm for avoiding dynamic objects in multi-robot systems using a priority-based approach. The robots are assigned unique ID numbers as their priority levels. When two robots encounter each other, the lower priority robot takes an action, such as stopping or speed reduction, to avoid the higher priority one. This approach helps prevent deadlocks and enables smooth navigation in a multi-robot environment.

Dynamic-Window Approach (DWA): Fox et al. (1997) introduced the DWA \cite{dwa} as a local motion planning method for mobile robots that combines trajectory generation and obstacle avoidance. The main idea of DWA is to search for a control input (linear and angular velocities) within a dynamically constrained window. The window is determined based on the robot's current velocity, acceleration limits, and the time required to come to a complete stop. DWA generates a set of candidate trajectories by sampling control inputs within the dynamic window and evaluates them based on three criteria: progress towards the goal, clearance from obstacles, and velocity. The control input corresponding to the trajectory with the highest score is selected as the optimal solution. DWA has been successfully applied in various robotic applications and is particularly well-suited for real-time implementations due to its computational efficiency. However, DWA has some limitations. It is primarily a local planner, making it susceptible to local minima and potentially requiring a global planner for more complex environments. Additionally, DWA assumes that the environment is static and may not perform as well in highly dynamic scenarios with multiple moving objects. To overcome this limitation, extensions of DWA, such as the Velocity Obstacle (VO) method (Fiorini and Shiller, 1998) \cite{fiorini1998motion} and the Time-Elastic Band (TEB) approach (Rösmann et al., 2017) \cite{teb}, have been proposed to consider the dynamic nature of the environment.

\paragraph{Learning-Based Approaches}
Social Force Model (SFM): Helbing and Molnar (1995) proposed the SFM \cite{helbing1995social}, a physics-inspired approach that models the interaction between pedestrians using attractive and repulsive forces. SFM considers individual motivations, such as desired speed and direction, and the influence of surrounding agents to predict pedestrian trajectories. Although SFM has been successful in simulating pedestrian behavior in various scenarios, its parameters need to be carefully tuned, and it may not generalize well to non-pedestrian agents or complex environments.

Deep Imitative Models (DIM): Rhinehart et al. (2018) proposed DIM \cite{carla}, which uses deep neural networks to imitate expert behavior for trajectory prediction and motion planning. By learning from demonstrations, DIM can capture complex human-like behavior and generalize to a wide range of scenarios. DIM has shown promising results in predicting pedestrian and vehicle trajectories, as well as planning collision-free paths. However, the quality of the learned model depends on the quality of the expert demonstrations, and the method may struggle to handle situations not encountered during training.

Deep Reinforcement Learning for Collision Avoidance (DRLCA): Chen et al. (2016) proposed DRLCA \cite{drlca}, a deep reinforcement learning-based approach for collision avoidance in autonomous vehicles. DRLCA leverages the power of deep neural networks to learn collision avoidance policies directly from raw sensor data, without the need for hand-crafted features or expert demonstrations. By learning through trial and error, DRLCA can adapt to complex environments and handle previously unknown situations. DRLCA has demonstrated promising results in various simulated scenarios and real-world applications. However, deep reinforcement learning methods can be challenging to train, often requiring a significant amount of data and computational resources.

\subsection{Related Work}
\subsubsection{Multi-Object Tracking and Collision Avoidance }
Danescu et al. introduced a real-time multi-object tracking algorithm that uses 2D LiDAR scans combined with a Rao-Blackwellized particle filter for robust tracking in urban environments \cite{elevationmaps}.

Schulz et al. developed a multi-object tracking method that uses the Expectation Maximization (EM) algorithm  for tracking multiple moving objects in crowded environments \cite{trackingmultiple}. Tipaldi and Arras proposed an adaptive multi-hypothesis tracking approach that utilizes 2D LiDAR data to enable efficient tracking in environments with a high number of dynamic objects \cite{socialrobots}.

Fox et al. proposed the Dynamic Window Approach (DWA) for real-time robot navigation, which has been widely used with 2D LiDAR sensors for both autonomous vehicles and robotics \cite{dwa}. More recently, Zhang et al. presented a deep reinforcement learning-based method for local path planning and collision avoidance using 2D LiDAR data \cite{deep}.

Rösmann et al. introduced the Time-Elastic Band (TEB) method, an extension of the Dynamic Window Approach (DWA), which can handle dynamic environments by incorporating 2D LiDAR data for safe navigation in robotics applications \cite{teb}. The TEB method improves upon DWA by considering both the spatial and temporal aspects of robot trajectories. This allows for more accurate and efficient navigation in dynamic environments with moving obstacles. By optimizing the robot's trajectory using a cost function that considers factors such as path length, clearance from obstacles, and smoothness, the TEB method aims to generate safe and feasible trajectories for robots operating in cluttered and complex environments. This approach is particularly relevant to our work as we focus on developing effective multi-object tracking and predictive collision avoidance techniques for autonomous mobile robots, such as industrial AGVs, in dynamic environments.

A notable work in local navigation is the Adaptive Dynamic Window Approach (ADWA) proposed by Matej Dobrevski and Danijel Skocaj \cite{dobrevskiadaptive2020}. The ADWA is an extension of the original \glsxtrshort{dwa}. The ADWA improves the original DWA by introducing an adaptive search space that changes dynamically based on the robot's current state and the environment. This adaptive approach allows the robot to better handle complex environments with narrow passages and dynamic obstacles. The ADWA also incorporates a more sophisticated cost function that considers not only the distance to the goal and the robot's velocity but also the smoothness of the path and the robot's orientation. Dobrevski and Skocaj conducted experiments using a mobile robot equipped with a LiDAR sensor, demonstrating that the ADWA outperformed the original DWA in terms of obstacle avoidance, navigation efficiency, and overall path quality. Although their work focused on local navigation and did not directly address LiDAR-based object tracking and filtering, it is still a relevant contribution to the field of autonomous navigation in dynamic environments.

\subsubsection{Other Approaches for MOT and Collision Avoidance}

Qi et al. proposed Frustum PointNets, a deep learning-based method for 3D object detection and tracking using 2D LiDAR data along with RGB images \cite{pointnet}. Chen et al. introduced a LiDAR-based end-to-end trainable deep network, named LiDAR-RCNN, for real-time multi-object tracking in autonomous driving scenarios \cite{lidarrcnn}.

Yang, et al. presented a hierarchical approach that combines model predictive control (MPC) with a polygonal distance-based dynamic obstacle avoidance method is proposed to achieve dynamic and accurate collision avoidance,. \cite{mpcdouble}. Gu et al. proposed a deep reinforcement learning-based algorithm for collision avoidance that leverages 2D LiDAR data for training and deployment \cite{realtime}.

Missura and Bennewitz propose an enhanced Dynamic Window Approach (DWA) for mobile robot navigation in dynamic environments \cite{pcadwa}. Their method predicts the motion of other objects, assuming constant velocity over short time intervals, and integrates these predictions into the DWA's cost function. They use polygonal clearance to model the environment, including both static and dynamic obstacles. The modified cost function allows the robot to better assess collision risks and select safer trajectories. The effectiveness of the proposed approach is demonstrated through kinematic simulation, highlighting its potential for improving robotic navigation in dynamic environments.

Multi-object tracking was effectively implemented in \cite{motenkf} using \glsxtrshort{enkf} with the \glsxtrshort{nnda}. In this paper, the result is compared with \glsxtrshort{jpda}. \glsxtrshort{jpda} was found to have less error but was computationally expensive. 

%% file: sections/methodology.tex
\pagestyle{plain}
\section{Methodology}
\label{sec:methods}
\subsection{System Kinematic Model}
\begin{figure}[H]
    \centering
    \includegraphics[width=0.6\textwidth]{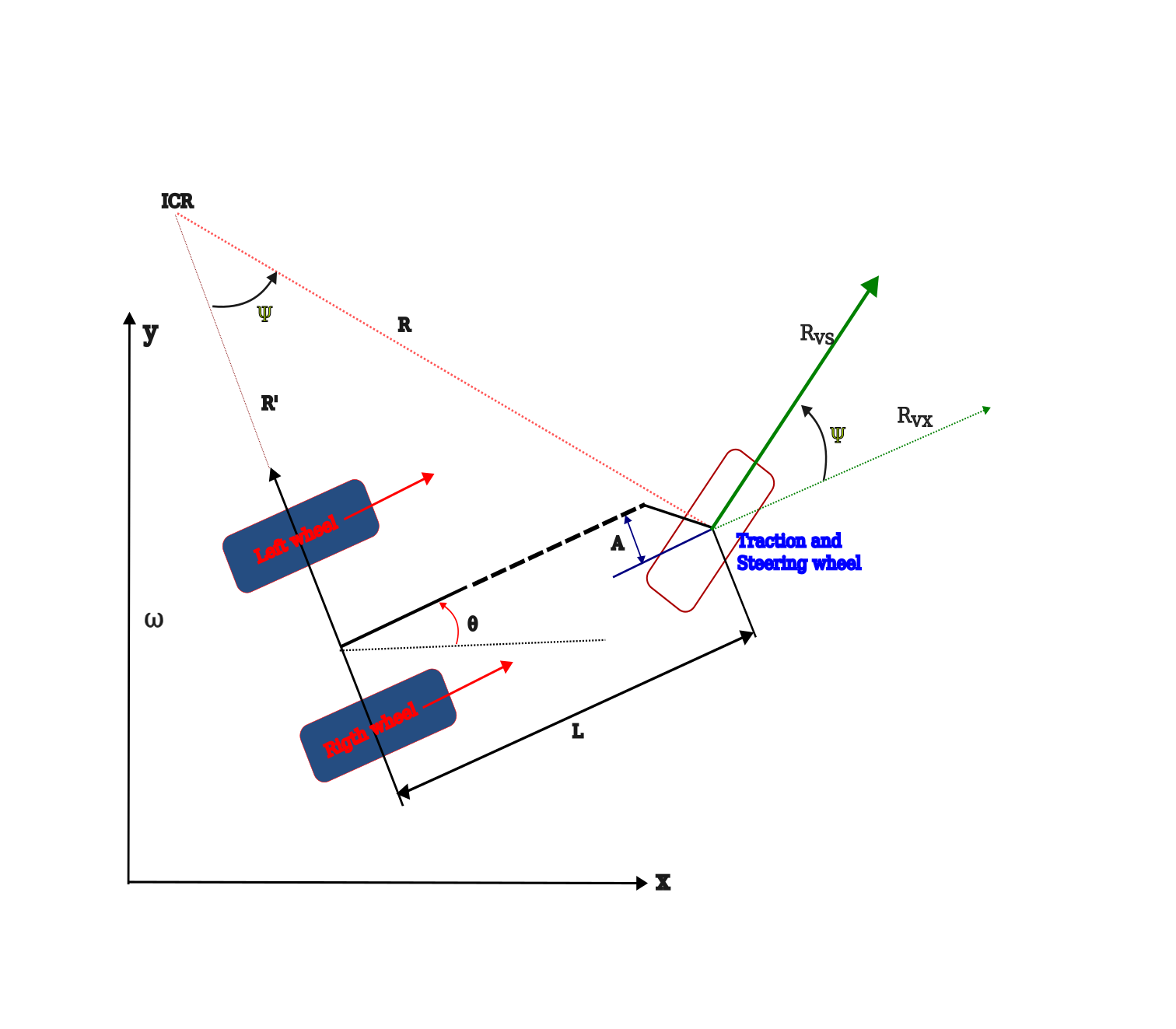}
    \caption{\textit{Tricycle Kinematic Model with front wheel traction and steering} }
    \label{fig:kinemtics}
\end{figure}

For this research, a tricycle kinematic model is used for the robot to estimate the velocity of the robot from wheel encoders. $\psi$ is the steering angle, $\nu_s$ traction wheel linear velocity, $\omega_s$ traction wheel rotational velocity, traction wheel radius r, and $\theta$ is  the orientation of the robot in the map frame. \autoref{fig:kinemtics} shows the kinematics diagram.

\begin{equation}
\nu_s = \omega_s*r; \textbf{  }
\omega = \nu_s/R = \nu_s s_\psi/L 
\end{equation}

\begin{equation}
\begin{bmatrix}
^{\mathcal{R}}\nu_x \\
^{\mathcal{R}}\nu_y \\
^{\mathcal{R}}\dot{\theta}
\end{bmatrix} = 
\begin{bmatrix}
c_\psi - A \cdot s_\psi/L & 0\\
0 & 0\\
0& 1
\end{bmatrix} 
\begin{bmatrix}
\nu_s\\
\omega
\end{bmatrix}
\end{equation}

\begin{equation}
\begin{bmatrix}
^{\mathcal{M}}\dot{x} \\
^{\mathcal{M}}\dot{y} \\
^{\mathcal{M}}\dot{\theta}
\end{bmatrix} = 
\begin{bmatrix}
c_\theta c_\psi & 0\\
s_\theta c_\psi & 0\\
0& 1
\end{bmatrix} 
\begin{bmatrix}
\nu_s\\
\omega
\end{bmatrix}
\end{equation}


This will be used to simulate the motion of the robot for \autoref{subsec:trajectory} any other kinematic model can be used depending on the architecture of the robot.

\subsection{Object Detection}
In this section, we describe the object detection process using a 2D LiDAR sensor, which is an essential component of the proposed mobile robot navigation system. The 2D LiDAR sensor provides range and angle measurements of the surrounding environment, allowing the system to detect and locate objects.

\subsubsection{Data Acquisition}
The 2D LiDAR sensor emits laser pulses and measures the time it takes for the pulses to bounce back after hitting an object. By knowing the speed of light and the time of flight, the sensor calculates the distance to the object. The sensor rotates to cover the entire field of view, providing a set of distance measurements at different angles, which are collectively called a scan.

\subsubsection{Scan Processing}
The raw data from the LiDAR sensor is typically noisy and requires preprocessing to obtain accurate object detection. Common preprocessing steps include filtering out the noise, down-sampling, and converting the polar coordinates (range and angle) into Cartesian coordinates (x, y):

\begin{equation}
\label{eq:polar2cart}
^\mathcal{R}x = r \cos(\theta), \quad ^\mathcal{R}y = r \sin(\theta)
\end{equation}

where $r$ is the range, and $\theta$ is the angle.

In this study, two LiDAR scanners were employed to provide a more comprehensive view of the environment. To effectively process and analyze the data from both scanners, the scans were first synchronized by approximating their timestamps. Then, the point clouds from the two scanners were merged, and the combined data was transformed into the global map frame using the following transformation equation:

\begin{equation}
\begin{bmatrix}
^\mathcal{M}x \\
^\mathcal{M}y \\
\end{bmatrix} = 
\begin{bmatrix}
\cos(\theta_{_{R}}) & -\sin(\theta_{_{R}}) \\
\sin(\theta_{_{R}}) & \cos(\theta_{_{R}})
\end{bmatrix}
\begin{bmatrix}
^\mathcal{R}x \\
^\mathcal{R}y
\end{bmatrix} + 
\begin{bmatrix}
x_{_{R}} \\
y_{_{R}}
\end{bmatrix}
\end{equation}

where $\theta_{_{R}}$ is the yaw angle, $x_{_{R}}$, and $y_{_{R}}$ are the translation offsets of the robot in the global map frame while $^\mathcal{R}x$ 
 and $^\mathcal{R}y$ are the scans in Cartesian coordinated with respect to the robot frame obtained \autoref{eq:polar2cart}. This transformation enables seamless integration of the data, allowing for a more accurate and reliable object detection process.

\subsubsection{Object Segmentation}
After preprocessing the scan data, the next step is to segment the environment into distinct objects. The object segmentation method used in this work is based on the k-d tree algorithm from \glsxtrshort{pcl} \cite{5980567}, as described in \autoref{alg:kdtree-clustering}. Alternative clustering techniques such as mean-shift and DBSCAN were also tested for comparison purposes.

The k-d tree algorithm groups points that are close to each other based on a distance threshold. This method starts by selecting an unprocessed point, creating a new cluster, and iteratively adding neighboring points that are within the threshold distance. The process continues until all points in the scan have been assigned to a cluster.

The k-d tree algorithm was chosen over mean-shift and DBSCAN primarily due to its efficiency. The k-d tree provides a fast and scalable method for finding neighboring points, resulting in reduced computational complexity and faster processing times. While the segmentation accuracy of the k-d tree algorithm is comparable to that of mean-shift and DBSCAN, its superior computational performance makes it a more suitable choice for the proposed mobile robot navigation system. It is worth mentioning both DBSCAN and mean-shift were tested.

\setlength{\textfloatsep}{0.1cm}
\setlength{\floatsep}{0.1cm}
\begin{algorithm}[H]
    \caption{KDtree Euclidean clustering}\label{alg:kdtree-clustering}
    \begin{algorithmic}[1]
        \Require{$P$ : set of points, $k$ : minimum number of points in cluster, $d_{thresh}$ : distance threshold}
        \Ensure{Clusters of points}
        \Procedure{ kdtreeClustering} {$P, k, d_{thresh}$ }
            \State $tree \gets \textbf{buildKdtree}(P)$ \Comment{Build kdtree from $P$}
            \State $visited \gets {p \in P : p.\text{visited} = \text{False}}$ \Comment{Initialize visited set}
            \State $clusters \gets \emptyset$ \Comment{Initialize set of clusters}
            \For{$p \in P$}
                \If{$p.\text{visited} = \text{False}$}
                    \State $cluster \gets \textbf{searchCluster}(p, tree, k, d_{thresh})$ \Comment{Search for points in cluster}
                    \State $visited \gets visited \setminus cluster$ \Comment{Remove points from visited set}
                    \State $clusters \gets clusters \cup {cluster}$ \Comment{Add cluster to set of clusters}
                \EndIf
            \EndFor
            \State \textbf{return} $clusters$
        \EndProcedure
        
        \Procedure{searchCluster}{$p, tree, k, d_{thresh}$}
            \State $cluster \gets \{p\}$ \Comment{Initialize cluster}
            \State $p.\text{visited} \gets \text{True}$ \Comment{Mark point as visited}
            \State $neighbors \gets \text{kdtree\_query}(tree, p, d_{thresh})$ \Comment{Find nearby points}
            \If{$|neighbors| \geq k$}
                \For{$q \in neighbors$}
                    \If{$q.\text{visited} = \text{False}$}
                        \State $q.\text{visited} \gets \text{True}$ \Comment{Mark point as visited}
                        \State $cluster \gets cluster \cup \text{searchCluster}(q, tree, k, d_{thresh})$ \Comment{Recursively search for nearby points}
                    \EndIf
                \EndFor
            \EndIf
            \State \textbf{return} $cluster$
        \EndProcedure
    \end{algorithmic}
\end{algorithm}

\subsubsection{Object Representation}
After segmenting the objects, various representation methods can be employed, such as bounding boxes, circles, or polygons. Initially, the proposed system utilized cluster centroids as the object representation. However, this approach proved to be ineffective, as the centroid location was highly dependent on the point density within the cluster. To address this issue, an alternative method was introduced. This method involves identifying the extreme points of the cluster, calculating the geometric center, and then determining the radius that encompasses the extreme points. This approach yields a more accurate and stable object representation. \autoref{eq:center} shows the method followed which is illustrated in \autoref{fig:center_illu}. The result of this is shown in \autoref{fig:center} we can see the laser scan depicted in white, the cluster centroid in red, and the geometric center in green. Data obtained from 2d lidar and visualized in rviz.

\begin{equation}
    \label{eq:center}
    \begin{aligned}
        (c_x, c_y) &= \left(\frac{x_\text{min} + x_\text{max}}{2}, \frac{y_\text{min} + y_\text{max}}{2}\right), \\
        W &=  x_{max} - x_{min}, \\
        H &=  y_{max} - y_{min},\\
        R &= \ \frac{\sqrt{W^2 + H^2}}{2}\\
    \end{aligned}
\end{equation}
where $x_{\text{min}}$, $x_{\text{max}}$, $y_{\text{min}}$, and $y_{\text{max}}$ are the minimum and maximum x and y coordinates among all the points in the cluster, respectively. W and H are the width and height of the cluster bounding box. R is the radius of the distance from the geometric center to the extreme point

\begin{figure}[H]
    \begin{subfigure}[b]{0.45\linewidth}
         \centering
         \caption{\textit{Proposed solution illustration}}
         \includegraphics[width=1\linewidth]{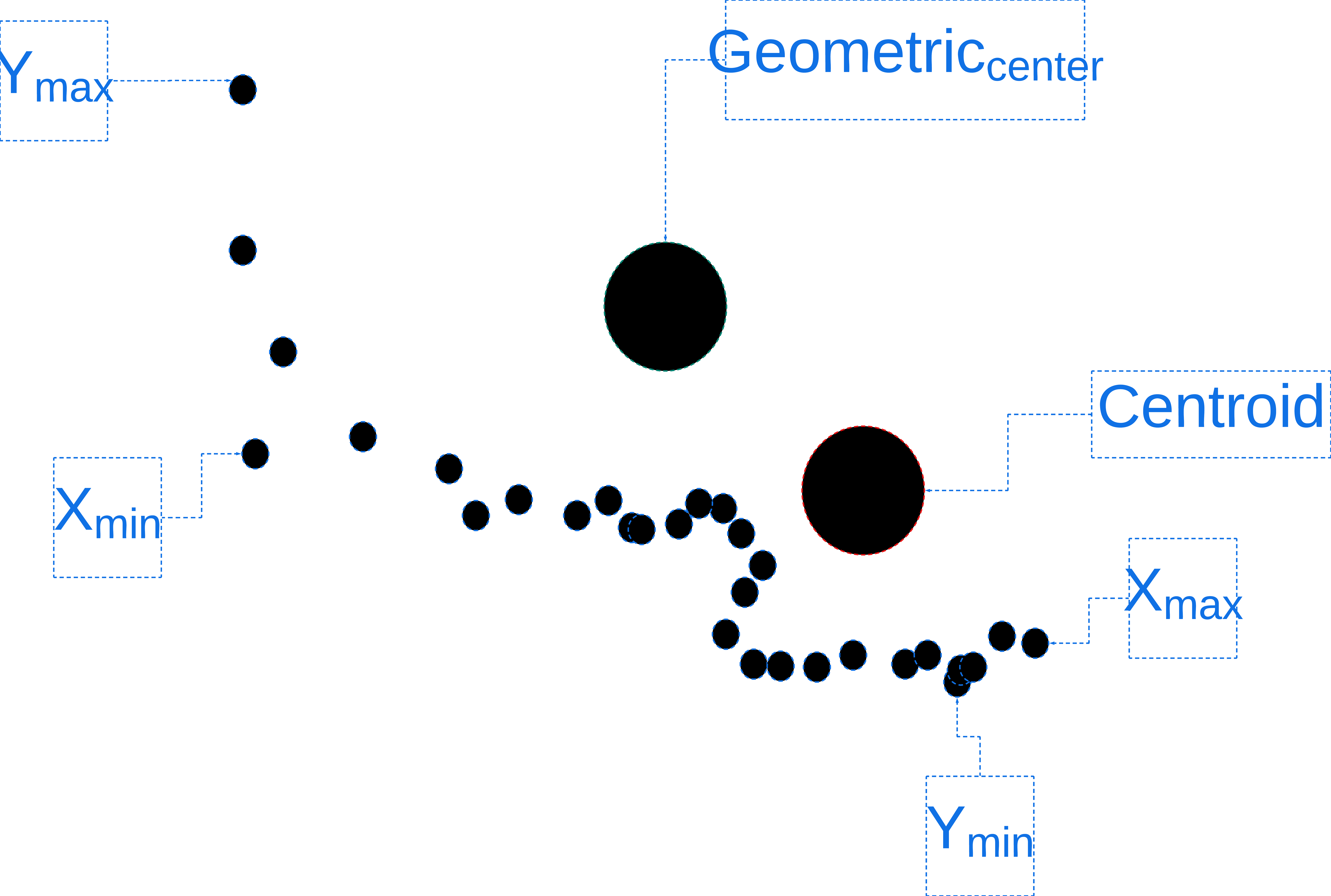}
         \label{fig:center_illu}
    \end{subfigure}
    \begin{subfigure}[b]{0.45\linewidth}
        \centering
        \includegraphics[width=1\linewidth]{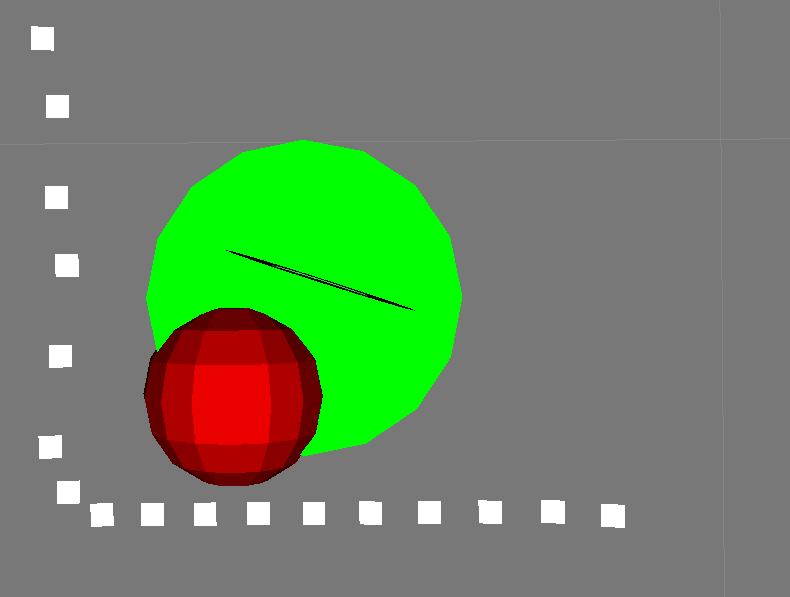}
        \caption{\textit{Obtained center(green) compared with cluster centroid(red) from laser scan} }         
        \label{fig:center}
    \end{subfigure}
\end{figure}
\begin{figure}[H]
    \centering
    \ContinuedFloat
    \begin{subfigure}[b]{0.45\linewidth}
         \centering
        \includegraphics[width=1\linewidth]{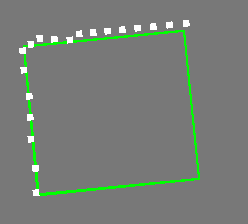}
        \caption{\textit{Obtained bounding box from laser scan} }         
        \label{fig:bounding_box}
    \end{subfigure}
\caption{\textit{Proposed solution and result in Rviz}}
\label{fig:proposed_bounding}
\end{figure}

In summary, the object detection process in our system consists of data acquisition, scan processing, object segmentation, and object representation. By using a 2D LiDAR sensor and appropriate algorithms, the proposed system can effectively detect and track obstacles in the environment, enabling safe and efficient navigation.

\subsection{Object Association}
In the literature review, several data association methods were discussed, focusing on the problem of tracking multiple moving objects. Among these methods, (\glsxtrshort{gnn}) and Greedy Nearest Neighbor (Greedy NN) were chosen for implementation in this work. The primary reasons for selecting these techniques are as follows:
\begin{itemize}
    \item \textbf{Simplicity}: Both \glsxtrshort{gnn} and Greedy NN are relatively simple and easy to implement compared to more complex data association techniques. Their simplicity allows for easier integration into existing systems and faster development and testing.
    
    \item \textbf{Computational Efficiency}: \glsxtrshort{gnn} and Greedy NN offer computationally efficient solutions for data association, making them suitable for real-time applications. Both methods require fewer computations compared to more complex techniques, such as Joint Probabilistic Data Association (JPDA) or Multiple Hypothesis Tracking (MHT).
    
    \item \textbf{Scalability}: Although \glsxtrshort{gnn} and Greedy NN may not be the most robust methods for data association, they can still handle a moderate number of objects and can be adapted to various application domains. Their scalability allows for a balance between computational efficiency and tracking performance.
    
    \item \textbf{Flexibility}: \glsxtrshort{gnn} and Greedy NN can be integrated with various motion models and sensor types, providing a versatile foundation for data association. This flexibility allows for easy adaptation to different tracking scenarios and sensor setups.
\end{itemize}

\begin{algorithm}[H]
\small
\caption{Global Nearest Neighbor (GNN) Algorithm}
\begin{algorithmic}[1]
\Require{$\mathcal{Z}$ : set of observations, $\mathcal{T}$ : set of existing tracks, $dist_{thresh}$ : distance threshold}
\Ensure{$\mathcal{M}$}
\Procedure{GNN}{$\mathcal{Z}, \mathcal{T}, \text{dist\_thresh}$}
    \State $\mathcal{C} \gets \text{cost matrix of size } (|\mathcal{Z}|+1) \times (|\mathcal{T}|+1)$
    \For{$i = 1$ to $|\mathcal{Z}|$}
        \For{$j = 1$ to $|\mathcal{T}|$}
            \State $\mathcal{C}_{i,j} \gets \text{distance}(\mathcal{Z}_i, \mathcal{T}_j)$
            \If{$\mathcal{C}_{i,j} > \text{dist\_thresh}$}
                \State $\mathcal{C}_{i,j} \gets \infty$
            \EndIf
        \EndFor
        \State $\mathcal{C}_{i, |\mathcal{T}|+1} \gets \infty$
    \EndFor
    \For{$j = 1$ to $|\mathcal{T}|$}
        \State $\mathcal{C}_{|\mathcal{Z}|+1, j} \gets \infty$
    \EndFor
    \State $M \gets \text{min-cost matching of } \mathcal{C}$
    \State \Return $\mathcal{M}$
\EndProcedure
\end{algorithmic}
\end{algorithm}
where $\mathcal{Z}$: is the set of measurements (or observations) received from the sensors at the current time step. Each element in $\mathcal{Z}$ corresponds to a detected object or a point of interest.
$\mathcal{T}$: The set of existing tracks (or targets) maintained by the tracking system. Each element in $\mathcal{T}$ represents a target being tracked, with an associated state estimate.

In the context of data association, the min-cost matching of the cost matrix $\mathcal{C}$ refers to finding the optimal assignment of the measurements (observations) to the existing tracks (targets) such that the total cost is minimized. The cost matrix $\mathcal{C}$ is an $m \times n$ matrix, where $m$ is the number of measurements and $n$ is the number of tracks. Each element $c_{ij}$ in the matrix represents the cost associated with assigning measurement $i$ to track $j$. The cost is typically based on a distance metric between the measurements and the predicted states of the tracks.

There are several algorithms for solving the min-cost matching problem, such as the Hungarian algorithm, the auction algorithm, or the Jonker-Volgenant algorithm. These algorithms find the optimal assignment of measurements to tracks that results in the lowest total cost, taking into account all possible assignments.

\begin{equation}
\label{eq:min_cost_match}
\text{minimize} \quad \sum_{i=1}^{m} \sum_{j=1}^{n} c_{ij} x_{ij}
\end{equation}
\begin{equation}
\label{eq:constraint1}
\text{subject to} \quad \sum_{i=1}^{m} x_{ij} = 1, \quad \forall j \in {1, \dots, n}
\end{equation}
\begin{equation}
\label{eq:constraint2}
\sum_{j=1}^{n} x_{ij} = 1, \quad \forall i \in {1, \dots, m}
\end{equation}
\begin{equation}
\label{eq:constraint3}
x_{ij} \in {0, 1}, \quad \forall i \in {1, \dots, m}, \quad \forall j \in {1, \dots, n}
\end{equation}
Here, $x_{ij}$ is a binary variable that indicates whether measurement $i$ is assigned to track $j$. The objective function \autoref{eq:min_cost_match} minimizes the total cost, while the constraints \autoref{eq:constraint1} and \autoref{eq:constraint2} ensure that each measurement is assigned to at most one track and each track is assigned at most one measurement. Constraint \autoref{eq:constraint3} enforces the binary nature of the assignment variables.

On the other hand, the Greedy Nearest Neighbor algorithm iteratively selects the pair of track and measurement with the smallest cost and assigns them to each other. After each assignment, the selected track and measurement are removed from further consideration.
\begin{algorithm}[H]
\small
    \caption{Greedy Nearest Neighbor}
    \begin{algorithmic}[1]
        \Require{$\mathcal{Z}$ : set of observations, $\mathcal{T}$ : set of existing tracks, $\mathcal{C}$ : distance threshold}
        \Ensure{$\mathcal{A}$}
        \Procedure{GreedyNN}{$\mathcal{Z}, \mathcal{T}, \mathcal{C}$}
        \State $\mathcal{U} \gets \mathcal{Z}$
        \State $\mathcal{A} \gets \emptyset$
        \While{$U \neq \emptyset$}
            \State $(i, j) \gets \arg\min_{(i, j) \in U \times \mathcal{T}} \mathcal{C}(i, j)$
            \State $\mathcal{A} \gets \mathcal{A} \cup {(i, j)}$
            \State $\mathcal{U} \gets \mathcal{U} \setminus {i}$
            \State $\mathcal{T} \gets \mathcal{T} \setminus {j}$
        \EndWhile
        \State \textbf{return} $\mathcal{A}$
    \EndProcedure
    \end{algorithmic}
\end{algorithm}

In this context, $\mathcal{Z}$ is the set of measurements, U is the set of unassigned measurements, $\mathcal{A}$ is a set that stores the pairs of matched objects from sets $\mathcal{T}$, and $\mathcal{Z}$. $\mathcal{C}(i, j)$ represents the cost between measurement i and track j. This algorithm starts with all measurements unassigned ($\mathcal{U} = \mathcal{Z}$). In each iteration, it selects the track-measurement pair with the smallest cost (i, j) and assigns them to each other. After the assignment, the selected measurement i and tracked j are removed from the unassigned sets $\mathcal{U}$ and $\mathcal{T}$, respectively. The algorithm continues until all measurements are assigned.

\subsection{Object tracking}
Object tracking is a crucial aspect of mobile robot navigation, and various algorithms exist to address this task. In this research, both the classical Kalman Filter and Ensemble Kalman Filter (\glsxtrshort{enkf}) were implemented and tested, as described in \autoref{alg:enkf}. The Ensemble Kalman Filter is commonly used for weather prediction, where the prediction step is referred to as the "forecast," and the correction step is called the "analysis."

The \glsxtrshort{enkf} was found to provide better results and exhibit greater stability compared to the classical Kalman Filter. Clustered objects from the previous section are treated as independent, non-correlated, n-dimensional states in the \glsxtrshort{enkf}. Each state space is projected onto N-ensembles during the filter initialization. This approach allows the filter to capture uncertainties and nonlinearities in the system more effectively. The ensembles act as particles that are propagated during the forecast and analysis stage.

\begin{algorithm}[H]
\small
    \caption{Ensemble Kalman Filter}
    \label{alg:enkf}
    \begin{algorithmic}[1]
        \Require State vector $\mathbf{\mathcal{X}}_{(0)}$, control vector $\mathbf{u}_{(0)}$, measurement vector $\mathcal{Z}_{(1)}$, measurement noise covariance matrix $R$, model function $f$, observation function $h$, number of ensembles $N$, ensemble perturbation matrix $\mathbf{P}$, and inflation factor $\alpha$
        \Ensure Updated state estimate $\hat{\mathbf{\mathcal{X}}}_{(k)}$ and covariance estimate $\mathbf{P}_{(k)}$
        \Procedure{EnKF}{$\mathcal{X}, \mathcal{Z}$}
            \For{$k \gets 1$ to $K$}
                \For{$i \gets 1$ to $N$}
                    \State Sample a perturbation vector $\boldsymbol{\epsilon}^{(i)}_k$ from the perturbation matrix $\mathbf{P}$
                    \State $\mathbf{\mathcal{X}}_{(k)}^i = f(\mathbf{\mathcal{X}}_{(k-1)}^i,\mathbf{u}_{(k)},\boldsymbol{\epsilon}_{(k)}^i)$ \Comment{state vector for ensemble member $i$ at time $k$}
                    \State $\mathcal{Y}_{(k)}^i = h(\mathbf{\mathcal{X}}_{(k)}^i) + \mathbf{v}_{(k)}^i$,  \Comment{$\mathbf{v}_{(k)}^i$ is a noise vector with mean zero and covariance $R$}
                \EndFor
                \State  $\bar{\mathbf{\mathcal{X}}}_{(k)} = \frac{1}{N} \sum_{i=1}^{N} \mathbf{\mathcal{X}}_{(k)}^i$ \Comment{mean of the ensemble forecasts(prediction)}
                \State  $\bar{\mathbf{\mathcal{Z}}}_{(k)} = \frac{1}{N} \sum_{i=1}^{N} \mathbf{\mathcal{Z}}_{(k)}^i$ \Comment{mean of the ensemble analysis(observation)}
                \State $\mathbf{P}_{(k)} = \frac{1}{N-1} \sum_{i=1}^{N} (\mathcal{X}_{(k)}^i - \bar{\mathcal{X}}_{(k)}) (\mathcal{Z}_{(k)}^i - \bar{\mathcal{Z}}_{(k)})^T$
                \State $\mathbf{K}_{(k)} = \mathbf{P}_{(k)}\mathbf{H}^T  (\mathbf{H} \mathbf{P}_{(k)} \mathbf{H}^T + \alpha \mathbf{R})^{-1}$  
                \State $\hat{\mathbf{\mathcal{X}}}_{(k)} = \bar{\mathbf{\mathcal{X}}}_{(k)} + \mathbf{K}_{(k)} (\mathcal{Z}_{(k)} - \mathcal{Y}_{(k)})$ \Comment{Update the state estimate}
            
            \EndFor
        \EndProcedure
    \end{algorithmic}
\end{algorithm}

By employing an Ensemble Kalman Filter, the proposed object tracking method offers improved accuracy and reliability in estimating object trajectories, which is essential for ensuring robust and safe navigation for mobile robots.
The objects are represented in \autoref{eq:enkfstate} where the superscript represents the ensemble for each state. This equation shows the state vector for one object

\begin{equation}
\label{eq:enkfstate}
\mathcal{X}_k^i = \begin{bmatrix}
x^1 && x^2 && ... && x^N &&\\
y^1 && y^2 && ... && y^N \\
\dot{x}^1 && \dot{x}^2 && ... && \dot{x}^N\\
\dot{y}^1 &&  \dot{y}^2 && ... && \dot{y}^N &&
\end{bmatrix} \textit{where }  i = {1,2,3,...N}
\end{equation}

\begin{figure}[H]
    \centering
    \includegraphics[width=0.45\textwidth]{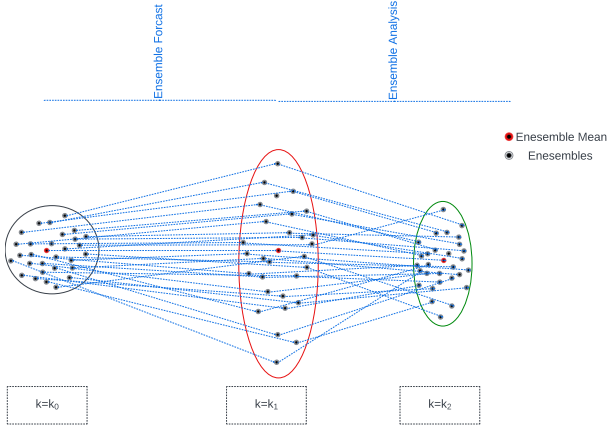}
    \caption{\textit{\glsxtrshort{enkf} Evolution through time for one state} }
    \label{fig:enkf_evo}
\end{figure}

\autoref{fig:enkf_evo} shows an illustration of the forecast and analysis stages in the ensemble Kalman filter. The figure demonstrates how the ensemble generation process, incorporating the calculated radius, is integrated into the overall functioning of the filter.

The radius from \autoref{eq:center} is utilized as a measure of the size or spread of the object clusters, providing essential information to generate ensembles that represent the uncertainty in the object's state. This approach enables the ensemble Kalman filter to effectively estimate the true state of the objects, considering the uncertainties associated with the measurements. In the proposed methodology, the ensemble members are generated by considering the radius calculated for each object. At each time step $k$, the $i$-th ensemble member $\mathcal{X}_k^i$ is obtained by adding a random noise term, $\mathcal{N}(0, R_k^i)$, to the mean of the $i$-th ensemble member from the previous time step, $\bar{\mathcal{X}}_{k-1}^i$. The random noise term is drawn from a normal distribution with mean 0 and covariance $R_k^i$, which is a function of the radius. This process is repeated for all ensemble members $i = \{1,2,3,...N\}$.

The ensemble generation equation is given by:

\begin{equation}
\label{eq:ensemble_generation}
\mathcal{X}_k^i = \bar{\mathcal{X}}_{k}^i + \mathcal{N}(0, R_k) \textit{ where } i = \{1,2,3,...N\}
\end{equation}

By incorporating the radius in the ensemble generation process, the proposed methodology accounts for the uncertainty in the object's position and velocity, resulting in a more accurate representation of the object's state.

\begin{figure}[H]
    \begin{subfigure}[b]{0.25\linewidth}
         \centering
         \includegraphics[width=1.0\linewidth]{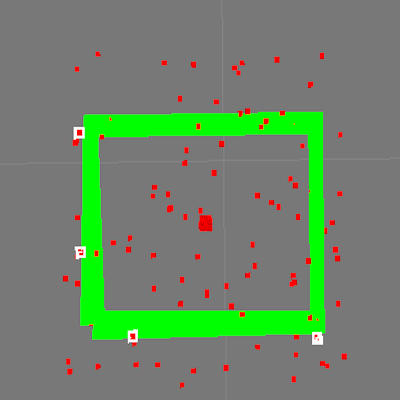}
         \label{fig:ensembles:a}
    \end{subfigure}
    \hfill
    \begin{subfigure}[b]{0.25\linewidth}
         \centering
         \includegraphics[width=1.0\linewidth]{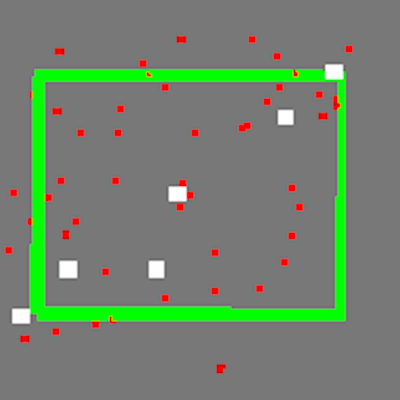}
         \label{fig:ensembles:b}
    \end{subfigure}
\caption{\textit{Sample of ensembles(red) generated for different objects, point clouds(white), and bounding box(green)}}
\label{fig:ensembles}
\end{figure}
\autoref{fig:ensembles} presents a visualization of the ensemble generation process using real data obtained from the LiDAR sensor. The figure highlights the effectiveness of using the calculated radius for ensemble generation in real-world scenarios.
\begin{equation}
\mathcal{X}_{(k)}^i = f(\mathcal{X}_{(k-1)}^i,\mathbf{u}_{(k)},\boldsymbol{\epsilon}_{(k)}^i)
\label{eq:state}
\end{equation}

The holonomic model can be used to represent the function $f$ in the Ensemble Kalman Filter algorithm. In a holonomic model, the object's states x and y can be controlled independently. For two-dimensional holonomic objects, the state vector $\mathcal{X}$ and the function f can be represented using time discretization with a time step $\Delta t$.


\begin{equation}
\label{eq:motion_function}
    f(\mathcal{X}^i_{(k)}, \mathbf{u}_{(k)}, \boldsymbol{\epsilon}_{(k)}) = 
    \begin{bmatrix} x_{(k-1)} + \dot{x}_{(k)} \Delta t + \epsilon_{x_{(k)}} \\ y_{(k-1)} + \dot{y}_{(k)} \Delta t + \epsilon_{y_{(k)}} 
    \end{bmatrix}
\end{equation}
where $\dot{x}$ and $\dot{y}$ are the linear velocities in the $x$ and $y$ directions, respectively., $\boldsymbol{\epsilon}_{(k)} = [\epsilon_{x_{(k)}}, \epsilon_{y_{(k)}}]^T$ represents the noise terms for $x$ and $y$ directions at time step $k$.

In this study, both position and velocity are estimated for the tracked objects. The lidar scan provides direct measurements for the position, while the velocity estimation is derived by differentiating the current position from the previous one and applying a sliding window. The mean ensemble is forecasted using the motion model in \autoref{eq:motion_function}:

\begin{equation}
\dot{x}_{(k)} = \frac{x_{(k)} - x_{(k-1)}}{\Delta t}, \quad \dot{y}_{(k)} = \frac{y_{(k)} - y_{(k-1)}}{\Delta t}
\label{eq:velocity_calculation}
\end{equation}

The velocities are computed by differentiating the position measurements with respect to time, as shown in Equation \autoref{eq:velocity_calculation}:

To reduce the noise in the velocity estimates, a sliding window approach is applied. The window size, denoted by \(W\), determines the number of previous velocity estimates to consider when computing the average velocity. The average velocities, \(\bar{\dot{x}}_{(k)}\) and \(\bar{\dot{y}}_{(k)}\), are calculated using the velocities in the window as shown in Equation \autoref{eq:sliding_window}:

\begin{equation}
\bar{\dot{x}}_{(k)} = \frac{1}{W} \sum_{i=0}^{W-1} \dot{x}_{(k-i)}, \quad \bar{\dot{y}}_{(k)} = \frac{1}{W} \sum_{i=0}^{W-1} \dot{y}_{(k-i)}
\label{eq:sliding_window}
\end{equation}
The choice of the window size, \(W\), in the sliding window approach, has a significant impact on the accuracy and noise characteristics of the velocity estimates. A small window size leads to a more responsive estimate, which can quickly adapt to changes in the true velocity. However, this also results in a higher sensitivity to noise in the velocity measurements, as the estimate is more influenced by short-term fluctuations. On the other hand, a larger window size smooths out the noise in the velocity estimates by averaging over a longer period, but at the expense of reduced responsiveness to rapid changes in the true velocity. In this study, the optimal window size is determined through experimentation and evaluation of the trade-off between noise reduction and responsiveness.


\subsection{Feasible Velocity Generation}
\label{subsec:trajectory}
An Optimal collision avoidance system should be able to generate velocity for a robot not only based on the current state of obstacles (dynamic and static) but considering their velocity. Here the classical dynamic windowing approach \cite{dwa}  is used with a modified cost function that considers the tracked objects' velocity and future position.

Given the robot's current pose $\mathbf{x}$, current velocity commands $v, \omega$, obstacle state $\mathcal{O}$,  time horizon $T$, maximum linear and angular velocity $v_{\text{max}}, \omega_{\text{max}}$, time resolution $\Delta t$, and trajectory $\mathbf{t_i}$ from $N$ number of samples, the dynamic window approach with obstacle avoidance outputs the best velocity commands $v_{\text{cmd}}, \omega_{\text{cmd}}$ that minimizes a scoring function while avoiding collisions with obstacles. the Pseudo-code for this is shown in \autoref{alg:dwa}. The total scoring function is defined in \autoref{eq:score}.

\begin{equation}
    \label{eq:score}
    \begin{split}
        \text{cost}(v_{\text{cmd}}, \omega_{\text{cmd}}) =  w_o \cdot obstacleCost(\mathbf{t_i},  \mathcal{O})\\ + w_{\nu} \cdot speedCost(v_{cmd}, v_{max}) \\+ w_g \cdot goalCost()
    \end{split}
\end{equation}

Where the $w_o$, $w_{\nu}$, and $w_g $ are the weights assigned to each cost. Each of the cost functions and the modifications made the to cost will be further discussed.

\begin{algorithm}[H]
\small
    \caption{Dynamic Window Approach for Robot Navigation with Obstacle Velocities} 
    \label{alg:dwa}
    \begin{algorithmic}[1]
    \Require{x : current position, $\nu$ : current velocity, g : goal, $\mathcal{O}$ : obstacles}
    \Ensure{$\nu*$}
    \Procedure{DynamicWindowApproach}{$x$, $\nu$, g, $\mathcal{O}$}
        \State $\mathcal{V} \gets \Call{GetAdmissibleVelocities}{x, \nu}$
        \State $best\_cmd \gets (0, 0)$
        \State $best\_cost \gets \infty$
        \For{$v \in \mathcal{V}$}
            \For{$\omega \in \mathcal{V}$}
                \State $\mathcal{T} \gets \Call{SimulateTrajectory}{x, v, \omega}$
                \State $cost_{o} \gets \Call{obstacleCost}{\mathcal{T}, \mathcal{O}}$
                \State $cost_{\nu} \gets \Call{speedCost}{v}$
                \State $cost_{g} \gets \Call{goalCost}{\mathcal{T}, g}$
                \State $total\_cost \gets w_o * cost_{o} + w_{\nu} * cost_{\nu} + w_g * cost_{g}$
                \If{$total\_cost < best\_cost$}
                    \State $best\_cost \gets total\_cost$
                    \State $\nu* \gets (v, \omega)$
                \EndIf
            \EndFor
        \EndFor
        \State \textbf{return} $\nu*$
    \EndProcedure
    \end{algorithmic}
\end{algorithm}

Some of the methods discussed in \autoref{sec:stateart} use \glsxtrshort{ttc} to consider dynamic objects' velocity in the cost function. \autoref{eq:obstaclettc} is a common way of calculating obstacle cost by taking into consideration the distance and the \glsxtrshort{ttc} here  where $d(\cdot)$ represents the distance between two points, and $TTC(\cdot)$ represents the \glsxtrshort{ttc} based on the relative velocity between the robot and the obstacle. $\epsilon$ is a small constant to avoid division by zero.

\begin{equation}
\label{eq:obstaclettc}
  \text{obstacleCost}(\mathcal{T}, \mathcal{O}) = \sum_{t \in \mathcal{T}} \sum_{o \in \mathcal{O}} \frac{1}{d(t_{xy}, o_{xy}) + TTC(t_{\dot{x}\dot{y}}, o_{\dot{x}\dot{y}}) + \epsilon}
\end{equation}

The distance between two points is calculated using the Euclidean distance formula below \autoref{eq:eucl} and the TTC in \autoref{eq:ttc}

\begin{equation}
    \label{eq:eucl}
    d(t_{xy}, o_{xy}) = \sqrt{(t_x - o_x)^2 + (t_y - o_y)^2} - R_r - R_o
\end{equation}

where $(t_x, t_y)$ and $(o_x, o_y)$ are the coordinates of the point in trajectory and center of obstacle respectively. $R_r$ and $R_o$ are the respective radii.

The \glsxtrshort{ttc} between the point and the obstacle is calculated as follows:
\begin{equation}
    \label{eq:ttc}
    TTC(t_{\dot{x}\dot{y}}, o_{\dot{x}\dot{y}}) = \frac{d(t_{xy}, o_{xy}))}{\lVert \nu_{r} - o_{\dot{x}\dot{y}} \rVert + \epsilon}
\end{equation}

where $\nu_{r}$ is the robot's velocity, and $o_{\dot{x}\dot{y}}$ is the obstacle's velocity. The $\lVert \cdot \rVert$ notation represents the magnitude (norm) of a vector, and $\epsilon$ is a small constant to avoid division by zero.

In this thesis the common equation in \autoref{eq:obstaclettc} is modified in a way that is more computationally simplistic and effective shown in \autoref{alg:obstaclecost} 
here algorithm calculates the obstacle cost for a given trajectory (t) and obstacles $\mathcal{O}$, representing the likelihood of a collision with obstacles by projecting the obstacles to their future state according to their velocities. A high obstacle cost implies a higher chance of collision, while a low cost suggests a safer trajectory. The algorithm computes the obstacle cost using distance-based calculations considering the future positions of the obstacles.
\begin{enumerate}
    \item Initialize variables: $skip_{n}$ (step size for trajectory iteration), $min_{dist}$ (minimum distance between trajectory and obstacles), dt (time step from config).
    \item Iterate through the trajectory with a step size of $skip_{n}$:
    \begin{itemize}
        \item  Iterate through each row of the obstacle matrix (config.ob).
        \item  Extract the obstacle's position and velocity.
        \item  Iterate through the robot's footprint (config.footprint) to check for collisions:
        (i) Calculate the global position of the footprint point.
        (ii) Update the obstacle's position based on its velocity and time step, considering the future position of the obstacle.
        (iii) Calculate the distance between the footprint point and the updated obstacle position, updating min\_dist if needed.
    \end{itemize}
        
    \item Check for collision by comparing $min_{dist}$ to the obstacle margin (config.obstacle\_margin). If a collision occurs, return infinity as the obstacle cost.

    \item Calculate and return the distance-based cost as the reciprocal of $min_{dist}$.
 
\end{enumerate}

\begin{algorithm}[H]
\small
    \caption{obstacleCost Function}\label{alg:obstaclecost}
    \begin{algorithmic}[1]
        \Function{obstacleCost}{$t, ob, config$}
        
            \State $skip_n \gets 2$  \Comment{points to skip in candidate trajectory mainly for faster computation}
            \State $min_{dist} \gets \infty$
            \State $\Delta t \gets skip\_n * config.dt$
            \ForAll{$ii \in {0, skip_n, ..., |t|}$}
                \ForAll{$i \in {0, 1, ..., ob.rows() - 1}$}
                    \State $o_x \gets ob.(i,0)$  \Comment{obstacle $x$ position}
                    \State $o_y \gets ob.(i,1)$  \Comment{obstacle $y$ position}
                    \State $o_{v_x} \gets ob.(i,2)$ \Comment{obstacle $\dot{x}$ velocity}
                    \State $o_{v_y} \gets ob.(i,3)$ \Comment{obstacle $\dot{y}$ velocity}
                    \ForAll{$j \in {0, 1, ..., |config.footprint[0]| - 1}$}
                        \State $p_x \gets t[ii][0] + config.footprint[j][0] * \cos(t[ii][2]) - config.footprint[j][1] * \sin(t[ii][2])$
                        \State $p_y \gets t[ii][1] + config.footprint[j][0] * \sin(t[ii][2]) + config.footprint[j][1] * \cos(t[ii][2])$\\
                        \State $d_x \gets p_x - (o_x + o_{\dot{x}} * \Delta t * ii)$ \Comment{obstacles projected to future time step}
                        \State $d_y \gets p_y - (o_y + o_{\dot{y}} * \Delta t * ii)$
                        \State $dist \gets \sqrt{d_x * d_x + d_y * d_y}$
                        \If{$dist < min_{dist}$}
                            \State $min_{dist} \gets dist$
                        \EndIf
                    \EndFor
                \EndFor
            \EndFor
            \If{$min_{dist} \leq config.obstacle\_margin$} \Comment{check for minimum distance requirement}
                \State \textbf{print} "Collision course detected"
                \State \Return $\infty$
            \EndIf
            \State $cost \gets \frac{1}{min_{dist}}$
            \State \Return $cost$
        \EndFunction
    \end{algorithmic}
\end{algorithm}

The speedCost function in \autoref{alg:goal_speedcost} calculates the cost based on the difference between the robot's current velocity and its maximum velocity. The cost is the square of this difference, which encourages the robot to travel faster when it is safe to do so.

\begin{algorithm}[hbt!]
    \caption{goalCost and speedCost Functions} \label{alg:goal_speedcost}
    \begin{algorithmic}[1]
        \Procedure{goalCost}{$\mathcal{T}, goal$}
            \State $f_p \gets \Call{GetFinalPoint}{\mathcal{T}}$ \Comment{final point generated from SimulateTrajectory}
            \State $d \gets \Call{Distance}{f_p, goal}$ \Comment{euclidean distance}
            \State $cost \gets d$
            \State \textbf{return} $cost$
        \EndProcedure\\
        
        \Procedure{speedCost}{$v$}
            \State $v_{max} \gets \Call{GetMaxVelocity}{}$
            \State $v_{diff} \gets v_{max} - v$
            \State $cost \gets v_{diff}$
            \State \textbf{return} $cost$
        \EndProcedure
    \end{algorithmic}
\end{algorithm}

The goalCost function in \autoref{alg:goal_speedcost} calculates the cost based on the distance between the final point of the trajectory and the goal. The cost is the square of this distance, which encourages the robot to move closer to the goal.

\begin{figure}[H]
    \begin{subfigure}[b]{0.4\textwidth}
         \centering
         \includegraphics[width=0.45\textwidth]{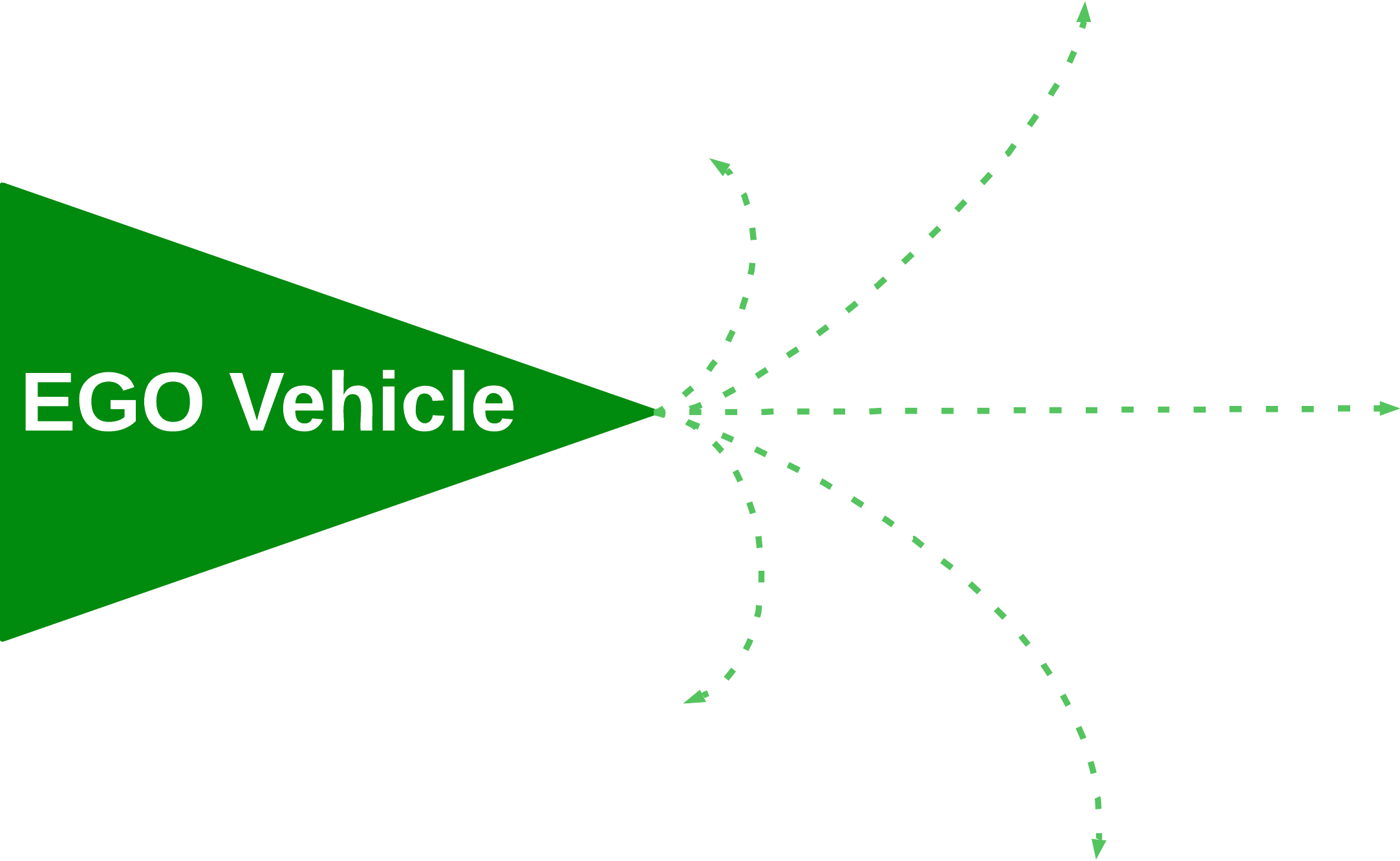}
         \caption{candidate trajectories generated using dynamic windowing}
         \label{fig:obstacle:a}
    \end{subfigure}
    \begin{subfigure}[b]{0.5\textwidth}
         \centering
         \caption{candidate trajectory that would result in a collision}
         \includegraphics[width=0.45\textwidth]{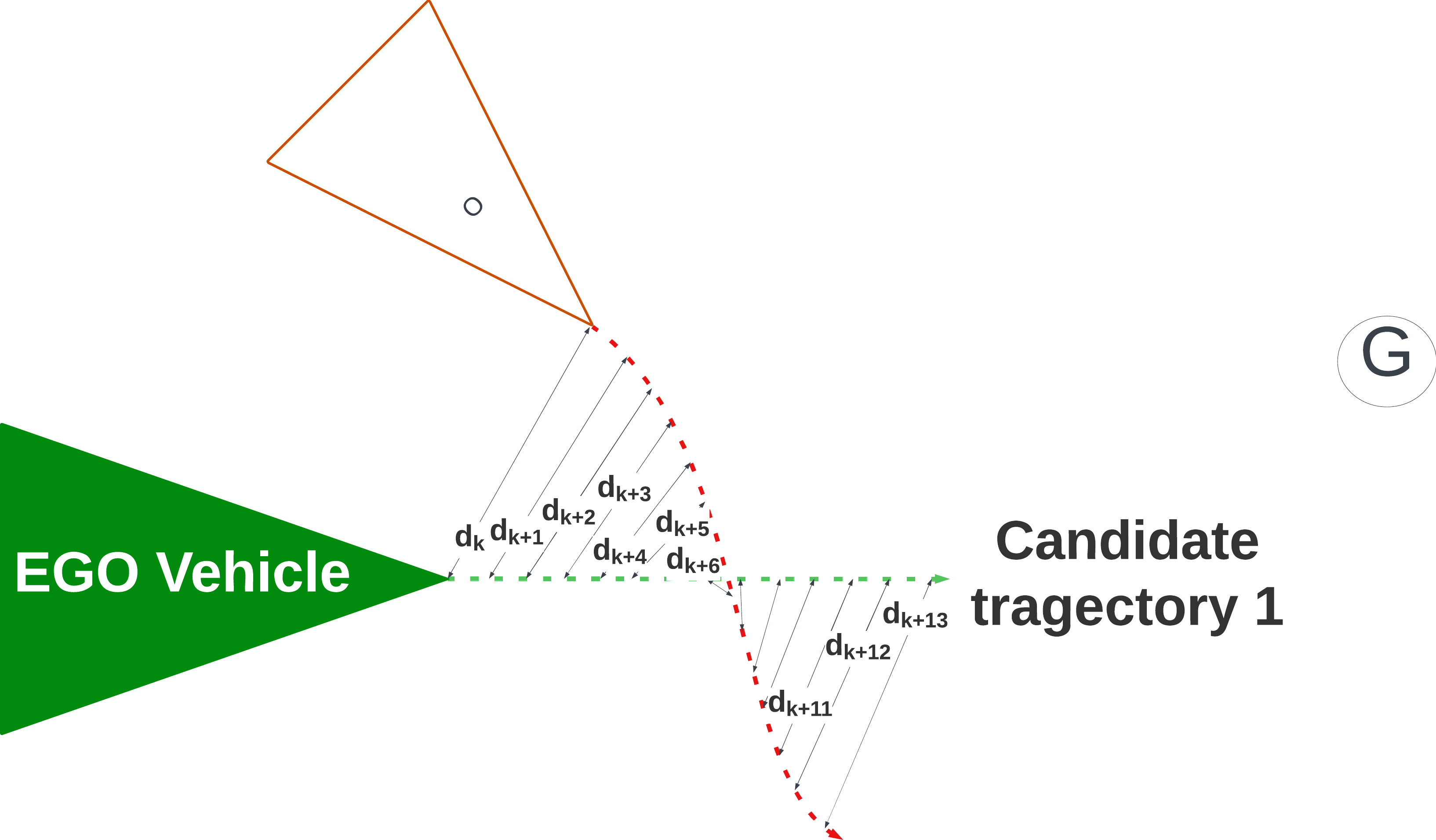}
         \label{fig:obstacle:b}
    \end{subfigure}
    \hspace{10mm}
    \begin{subfigure}[b]{0.5\textwidth}
         \centering
         \caption{candidate trajectory free of collision}
        \includegraphics[width=0.45\textwidth]{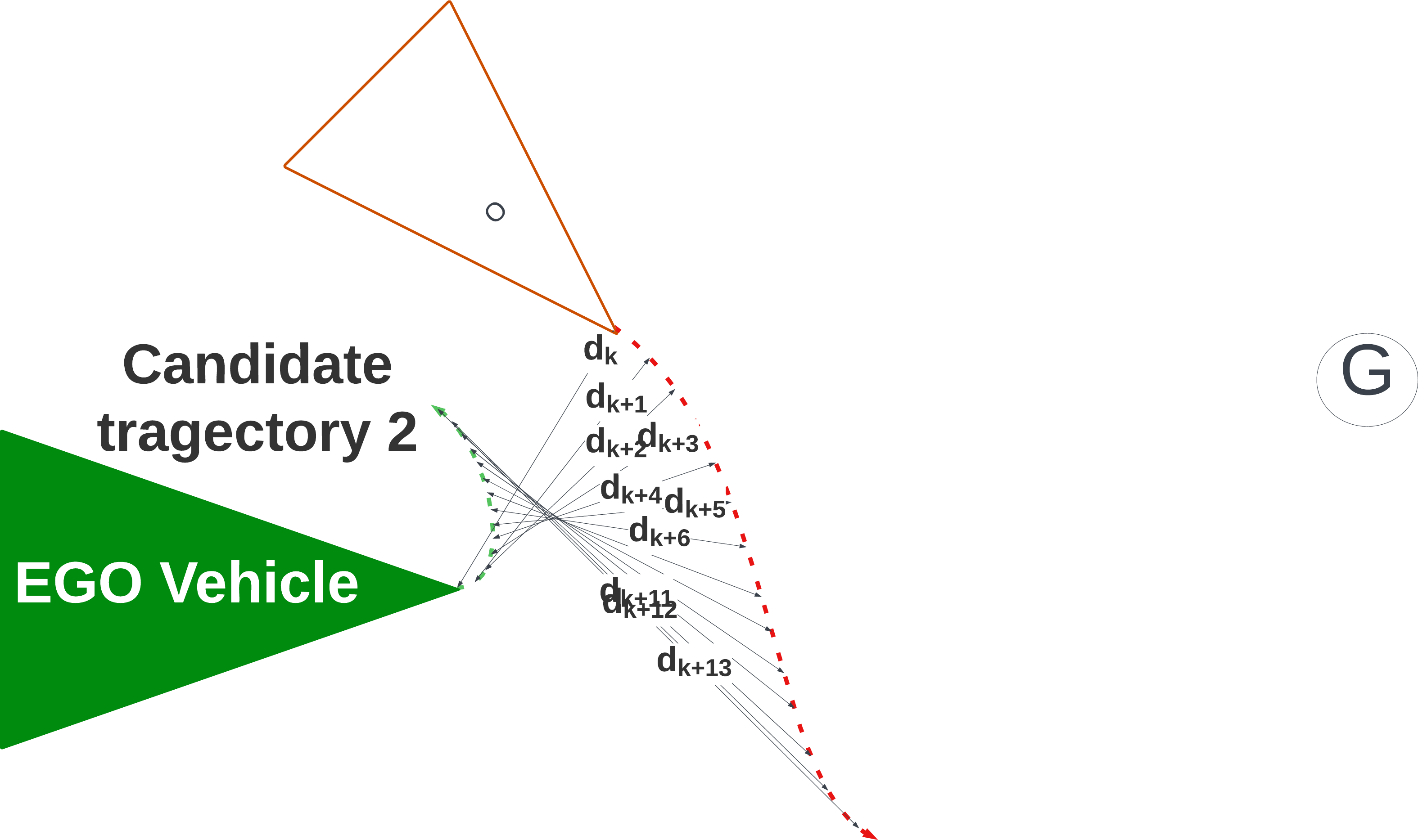}
         \label{fig:obstacle:c}
    \end{subfigure}
\caption{\textit{Scenario showing two vehicles in a collision path}}
\label{fig:obstacle}
\end{figure}

An illustration is provided in \autoref{fig:obstacle} to depict the above algorithm  where the velocity estimation is used to generate the trajectory of the obstacles. In \autoref{fig:obstacle:a} candidate trajectories are generated to which their cost will be evaluated in \autoref{fig:obstacle:b} we see a candidate trajectory that would cause a collision if it was chosen in the classical \glsxtrshort{dwa} this trajectory would be chosen. In \autoref{fig:obstacle:c} is one example that is obstacle free according to the new method but would be considered not feasible using the classical approach the obstacle cost considers the future position of the robots according to their velocities. In this new approach, \textit{d} represents the position difference between the robot and the obstacle projected in time. the smallest \textit{d} is taken and inverted as the cost

%% file: sections/system_architecture.tex
\section{System Architecture}
\label{sec:architecture}
In this section, the overall system architecture of the proposed solution will be discussed. The system is designed to perform real-time object detection, tracking, and predictive collision avoidance using a 2D LiDAR sensor. The architecture consists of several interconnected modules, each responsible for a specific task. The main components of the system architecture are as follows:

    

Here the steps mentioned in \autoref{sec:methods} are modularized in the following manner
\subsection{Data Processing} 
\begin{figure}[H]
    \centering
    
    \includegraphics[width=0.4\textwidth]{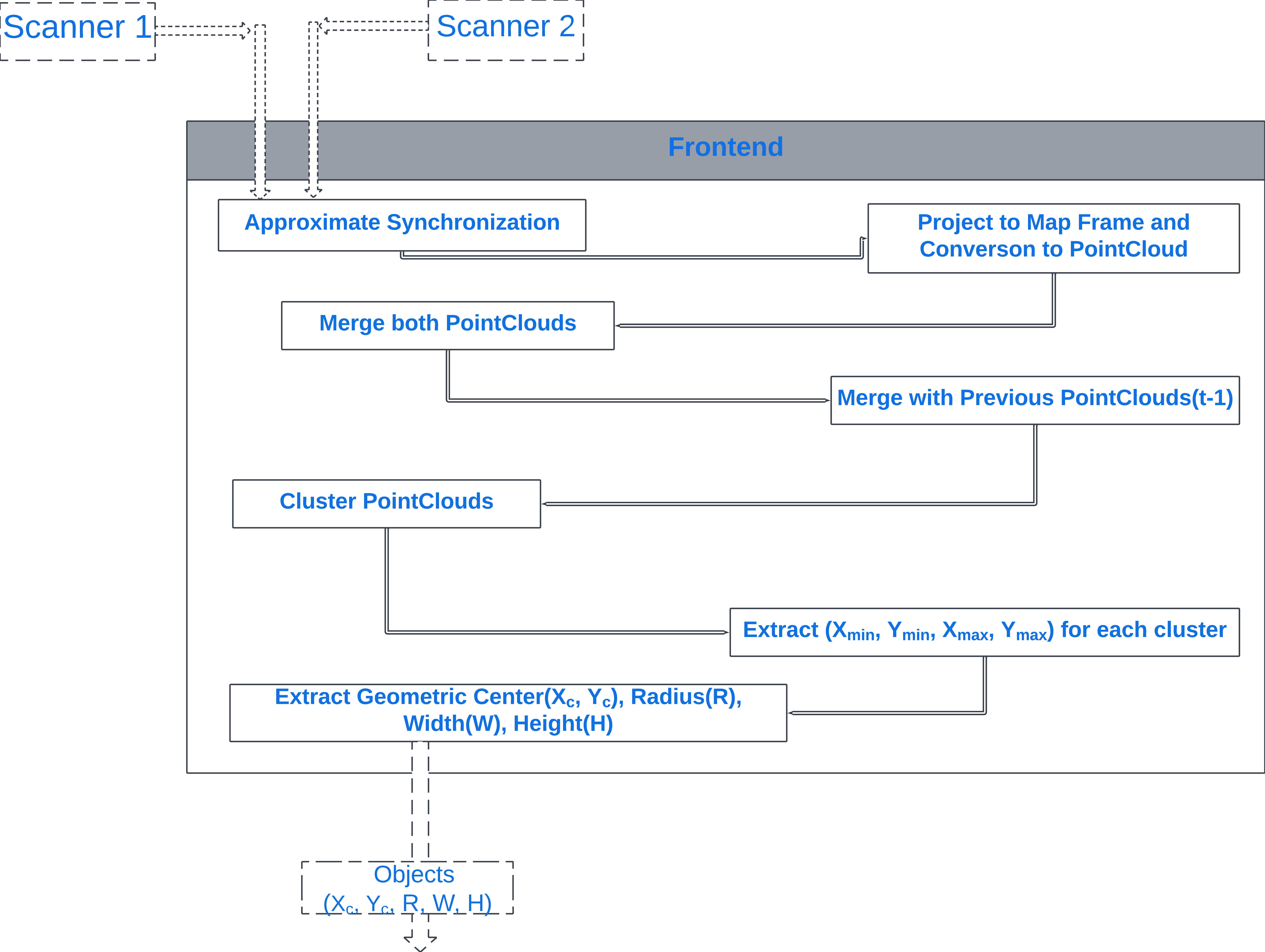}
    \caption{\textit{Architecture of Frontend for Extracting Objects from two 2D lidar scanners} }
    \label{fig:frontend}
\end{figure}

This acts as the front-end of the system taking in lidar scans
\begin{itemize}
    \item \textbf{Data Acquisition:} This module is responsible for acquiring raw data from the 2D LiDAR sensor(s). The data consists of range and angle measurements, which are preprocessed and converted from polar coordinates into Cartesian coordinates (x, y). Additional noise filtering can be done if necessary.
    \item \textbf{Object Clustering:} The pre-processed data are passed to the object segmentation module, which groups the data points into clusters representing individual objects. Clustering is performed using the Euclidean clustering algorithm based on k-d trees, as described in \autoref{alg:kdtree-clustering}.

    \item \textbf{Object Representation:} Once the objects have been segmented, each cluster's geometric center and radius are calculated using the equations presented in \autoref{eq:center}. These geometric properties are then used to represent the objects as rectangles in the global map frame.
    
\end{itemize}
An overall representation of front-end is shown in \autoref{fig:frontend}.

\subsection{Tracking}
\begin{itemize}
    \item \textbf{Initialization:} The object representation module provides the input for the object tracking module, which uses the Ensemble Kalman Filter (\glsxtrshort{enkf}) for state estimation, as detailed in \autoref{alg:enkf}. The state vector for each object consists of position and velocity components in the x- and y-directions.
    \item \textbf{Data Association:} Here in the back-end, the tracker associates new measurements with existing data; for this, multiple methods were used and tested, including greedy nearest neighbor and \glsxtrshort{gnn}(Hungarian) method. 
    \item \textbf{State clearing:} objects(obstacles) that do not get associated for a certain period are removed. This keeps the state from overgrowing.
    \item \textbf{Perturbation drawing} for both the initialization and prediction (forecast) stages of the tracker, the ensembles are drawn from a normal distribution $\mathcal{N}(0,  \epsilon )$ where $\epsilon$ is estimated from the radius of the tracked object for prediction and a constant value for the update.
\end{itemize}

\begin{figure}[H]
    \centering
    
    \includegraphics[width=0.4\textwidth]{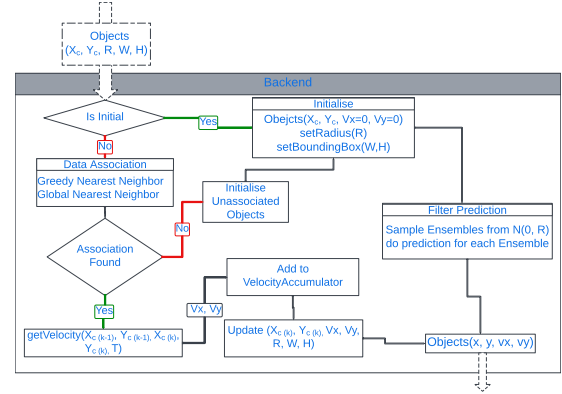}
    \caption{\textit{Architecture of Tracker Backend Ensemble Kalman Filter} }
    \label{fig:backend}
\end{figure}

An overall representation of the above tracking steps is shown in \autoref{fig:backend}.
\subsection{Collision Predictive Control}
\begin{figure}[H]
    \centering
    
    \includegraphics[width=0.45\textwidth]{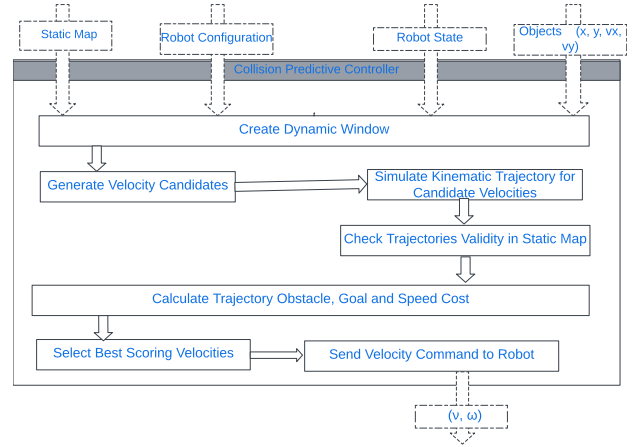}
    \caption{\textit{Architecture of Predictive Dynamic Collision Controller} }
    \label{fig:dwa}
\end{figure}

Based on the tracked objects, the predictive collision avoidance module calculates possible collision risks and generates safe, collision-free trajectories for the robot. The Dynamic Window Approach is employed to predict the motion of the robot for different velocity commands and compare and evaluate each trajectory to that of the obstacle's trajectory according to \autoref{alg:obstaclecost} in real-time. Then, the best velocity command is sent to the lower-level controller. The overall architecture is depicted in \autoref{fig:dwa}

The system architecture is modular, and it allows easy adaptation and integration of additional components, such as different sensors, object representation methods, or tracking algorithms. This design ensures that the system is robust and flexible, capable of addressing various requirements and real-world scenarios. The complete architecture of the system is included in \autoref{fig:rosgraph}. This \glsxtrshort{ros2} graph shows all the nodes and the communication between them for five robots.

%% file: sections/experiment_setup.tex
\section{Experimental Setup}
\label{sec:setup}
\subsection{Simulation Environment}
The simulation environment was set up using Gazebo\cite{Koenig2004} and Stage\cite{vaughan2008massively}, both open-source robotics simulators, to provide controlled environments for testing the proposed system. While Gazebo was used to simulate complex industrial settings, Stage was employed to test multiple robots simultaneously. \acrlong{ros2} was used for communication and control between the various components of the system.

The Gazebo simulated environment was designed to mimic various industrial settings with different configurations and complexities. The LiDAR sensor used in the simulation was based on the specifications of a real-world 2D LiDAR sensor. A forklift model was used in the simulation to represent the dynamic obstacles that are commonly encountered in industrial environments.

The forklift model was created by converting the \acrshort{cad} file, in \acrshort{stp} format, to a \acrshort{urdf} file using Blender with the Phobos add-on \cite{phobos}. This allowed for the seamless integration of the forklift model into the Gazebo simulation.

In the Stage simulator with \acrshort{ros2}, multiple Pioneer robots were used for testing the proposed multi-robot tracking and collision avoidance algorithms. The Pioneer robot is a widely used mobile platform in research and education, and its integration with the Stage simulator enabled the evaluation of the proposed methods in scenarios with multiple robots operating concurrently.

\begin{figure}[H]
    \begin{subfigure}[t]{0.49\linewidth}    
        \centering
        \includegraphics[width=1\linewidth]{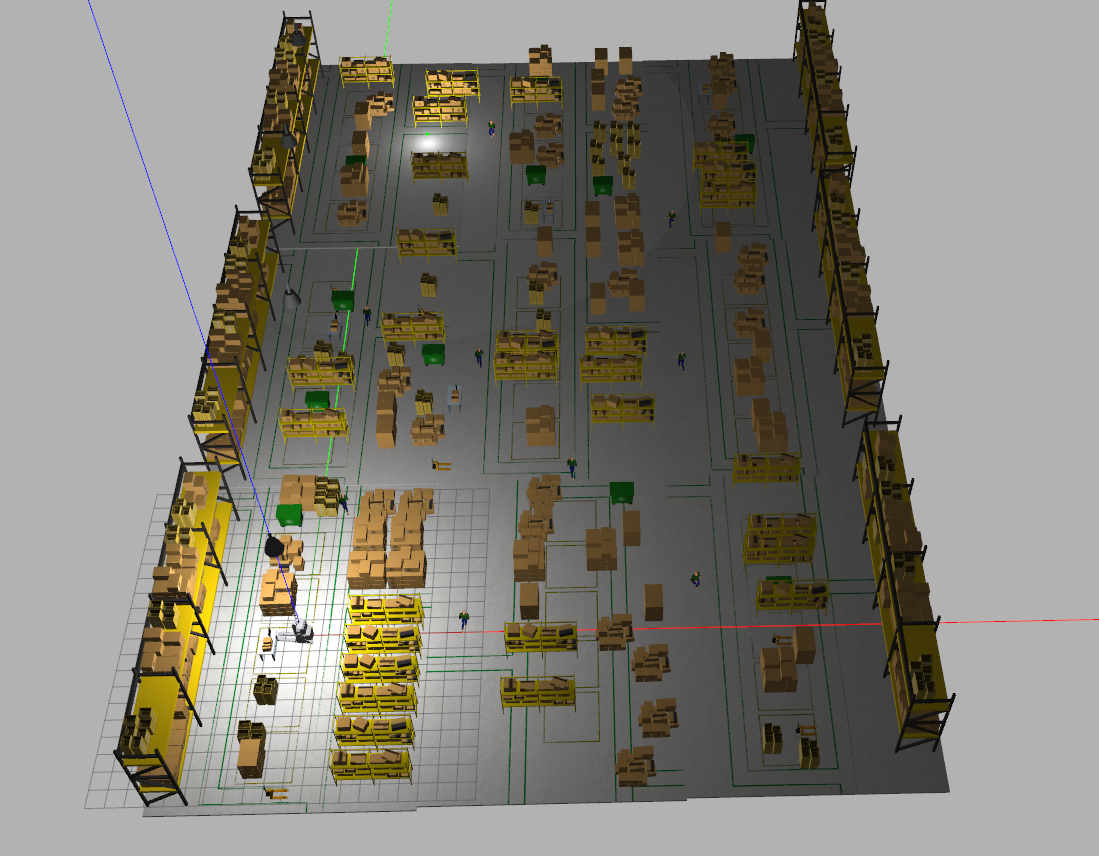}
        \caption{\textit{Gazebo simulation environment \cite{dynamicwarehouse}}}
        \label{fig:gazebo_environment}
    \end{subfigure}
    \begin{subfigure}[t]{0.49\linewidth}
        \centering
        \includegraphics[width=1\linewidth]{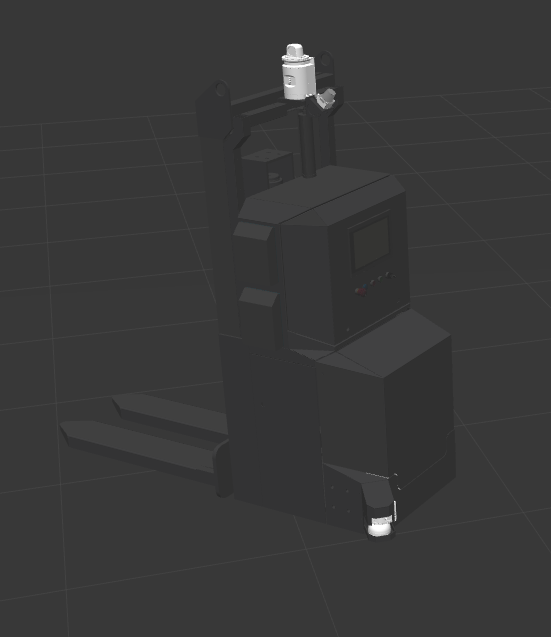}
        \caption{\textit{Simulated forklift in the Gazebo environment using the generated \acrshort{urdf}}}
        \label{fig:simulated_robot}
    \end{subfigure}
\end{figure}
\begin{figure}[H]
    \centering
    \ContinuedFloat
    \begin{subfigure}[t]{0.49\linewidth}
        \centering
        \includegraphics[width=1\linewidth]{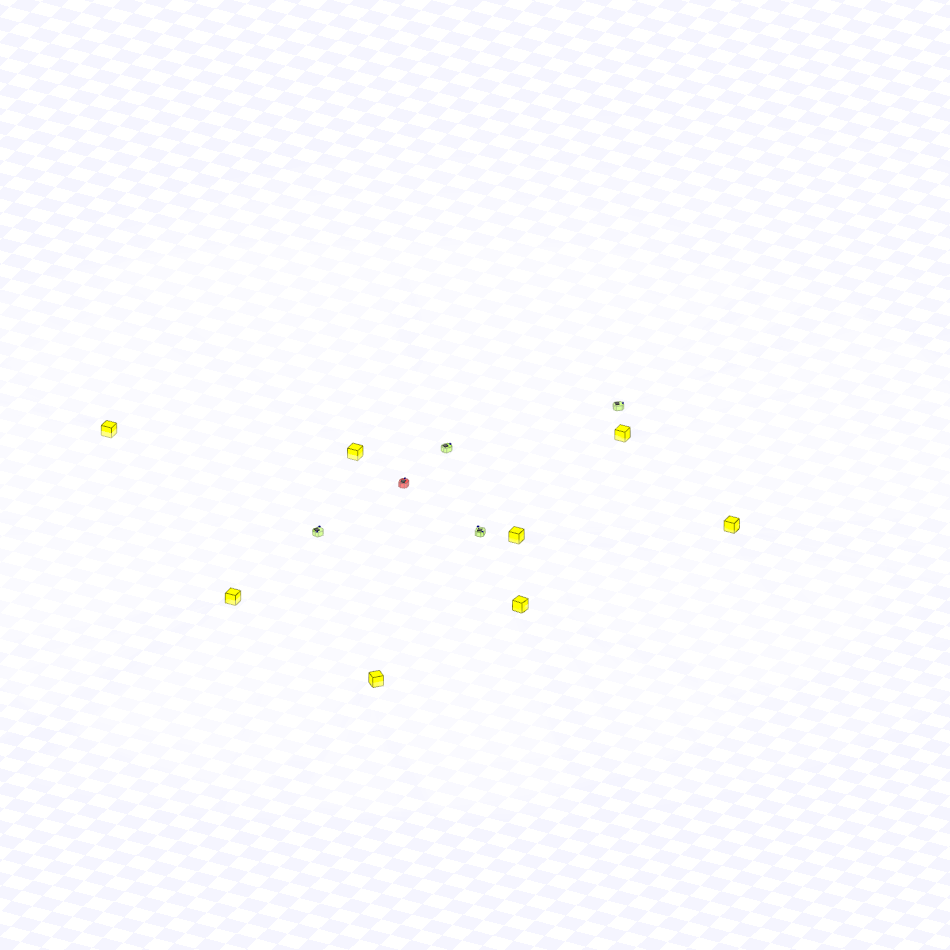}
        \caption{\textit{Stage simulation environment}}
        \label{fig:simulated_robot}
    \end{subfigure}
    \begin{subfigure}[t]{0.49\linewidth}
        \centering
        \includegraphics[width=1\linewidth]{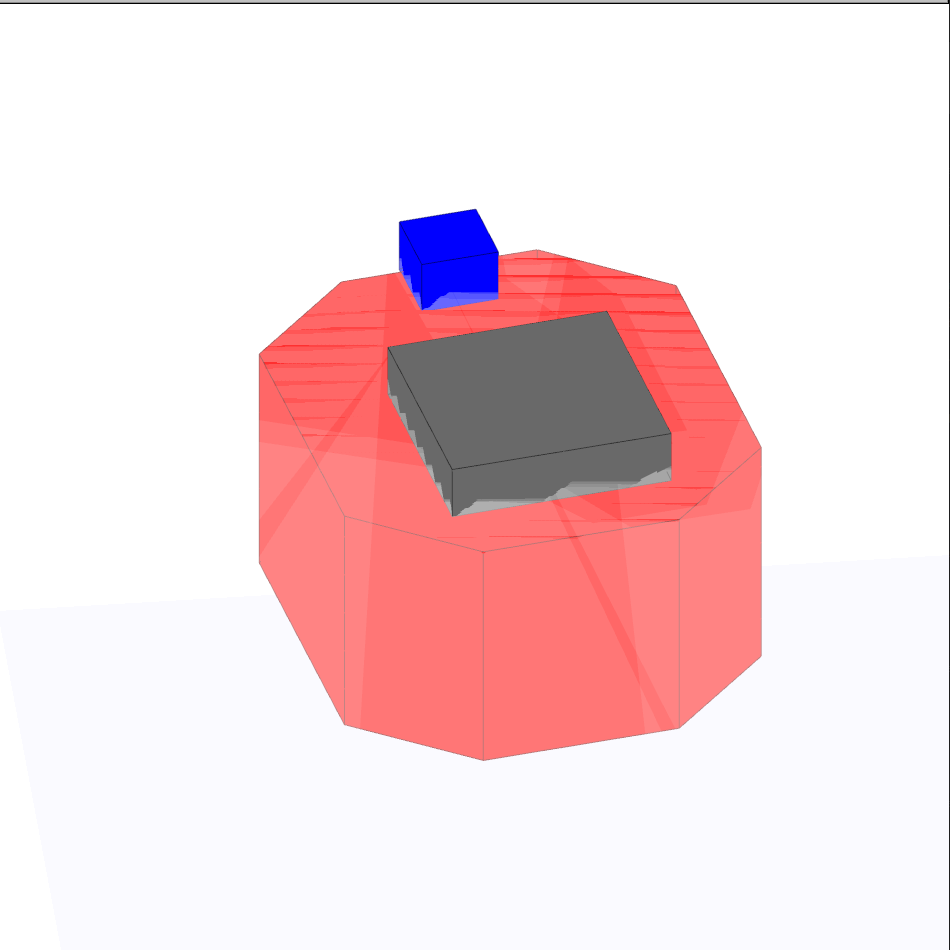}
        \caption{\textit{Simulated pioneer robot in stage environment}}
        \label{fig:simulated_robot}
    \end{subfigure}
        \caption{\textit{Gazebo and Stage simulation setup}}
\end{figure}

\subsection{Real-World Environment}
For real-world data collection, a forklift was used in an industrial environment to gather data from a 2D LiDAR sensor. The experimental setup was designed to test the proposed system in real-world conditions, including different obstacle configurations, dynamic objects, and various operating scenarios. \acrshort{ros2} was used for communication and control between the system components, while the \acrshort{udp} was utilized for communicating with the forklift's \acrshort{plc}.

\begin{table}[H]
    \centering
    \caption{Specifications of the 2D LiDAR sensor used in the experiments}
    \label{tab:lidar_spec}
    \begin{tabular}{|l|l|}
        \hline
        Parameter & Value \\
        \hline
        Resolution (can be configured)	& 30 mm, 40 mm, 50 mm,  \\
        & 70 mm, 150 mm, 200 mm\\
        Scanning angle & $275^o$\\
        Angular resolution & $0.39^o$\\
        Response time &	$\geq 95 ms$\\
        \hline
    \end{tabular}
\end{table}

\begin{figure}[H]
    \begin{subfigure}[b]{0.49\linewidth}
         \centering
         \includegraphics[width=1\linewidth]{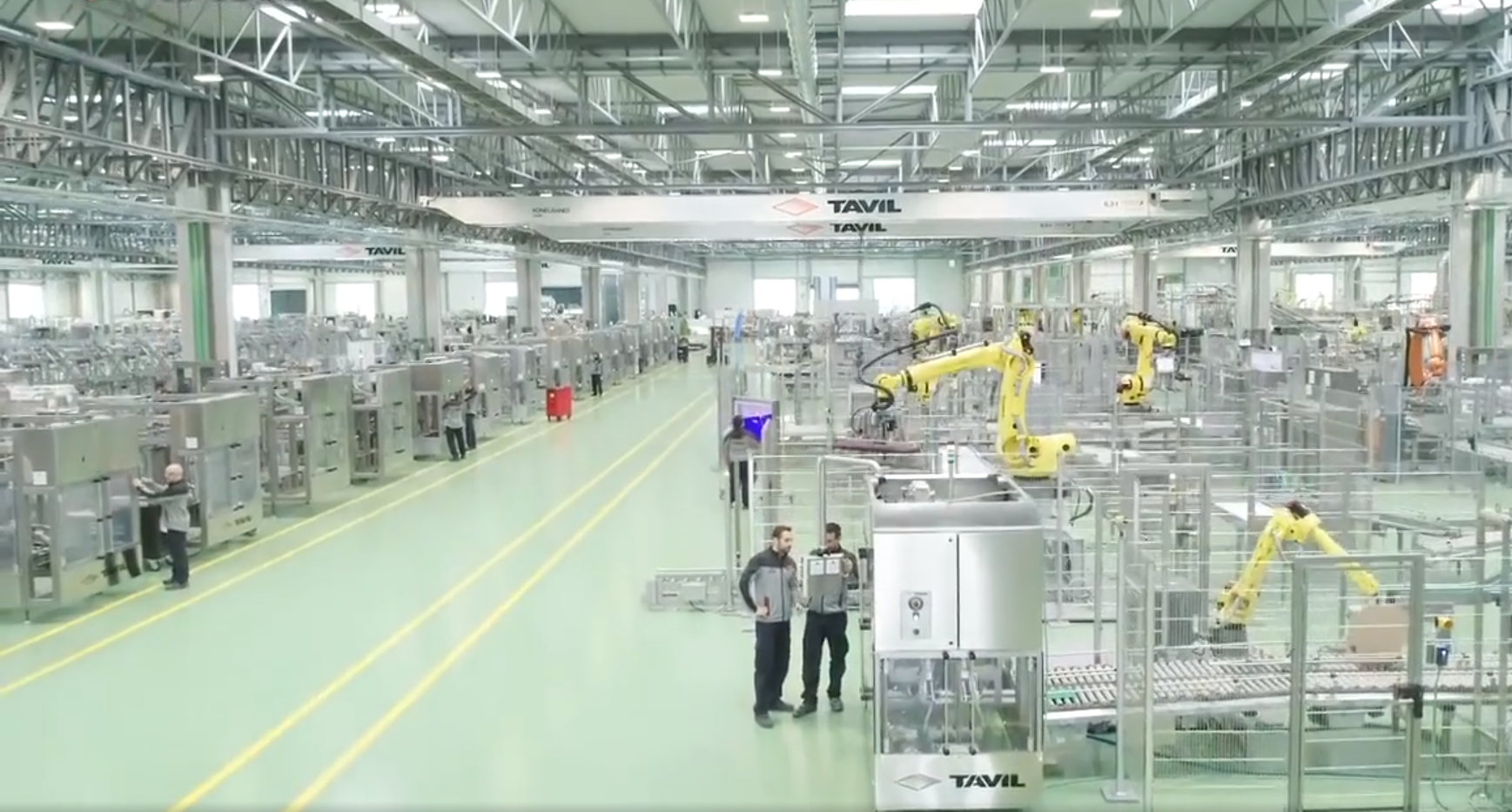}
         \caption{Sample Image of The Industrial Testing Environment }
         \label{fig:cad:a}
    \end{subfigure}
    \hfill
    \begin{subfigure}[b]{0.49\linewidth}
         \centering
         \includegraphics[width=1\linewidth]{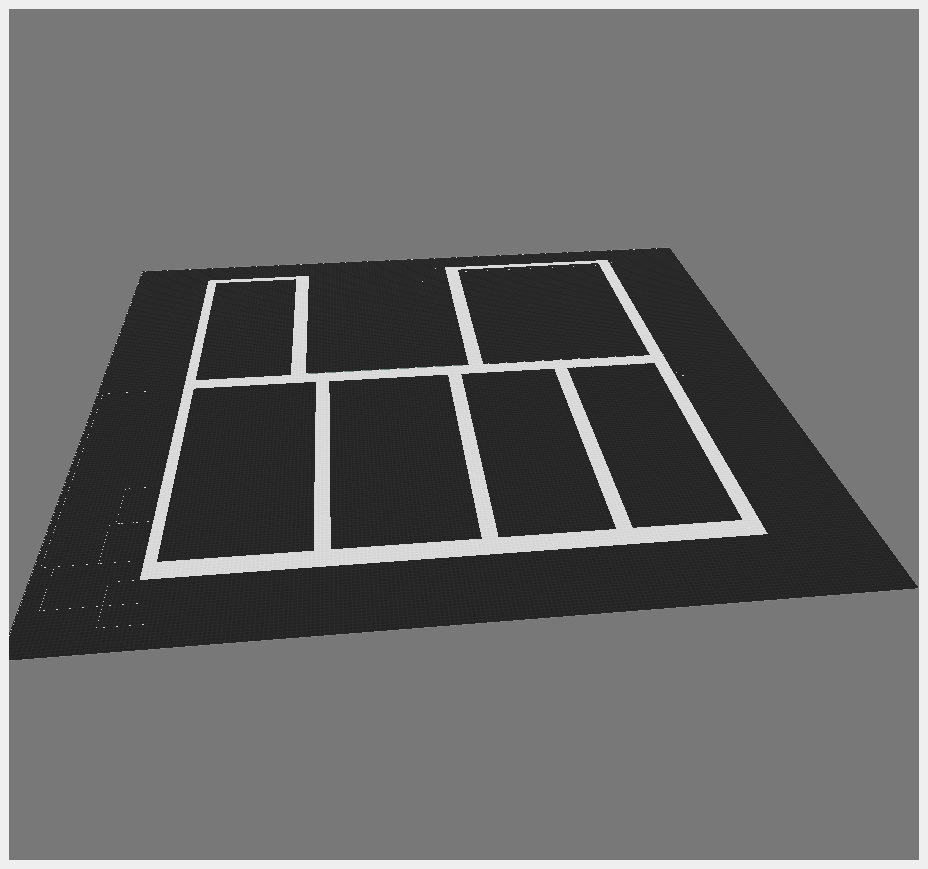}
         \caption{Static map of the Area 200x200m}
         \label{fig:forklift_spec:a}
    \end{subfigure}
    \caption{Real-world industrial environment used for data collection and testing}
    \label{fig:real_world_environment}
\end{figure}

\begin{figure}[H]
    \begin{subfigure}[b]{0.49\linewidth}
         \centering
         \includegraphics[width=1\linewidth]{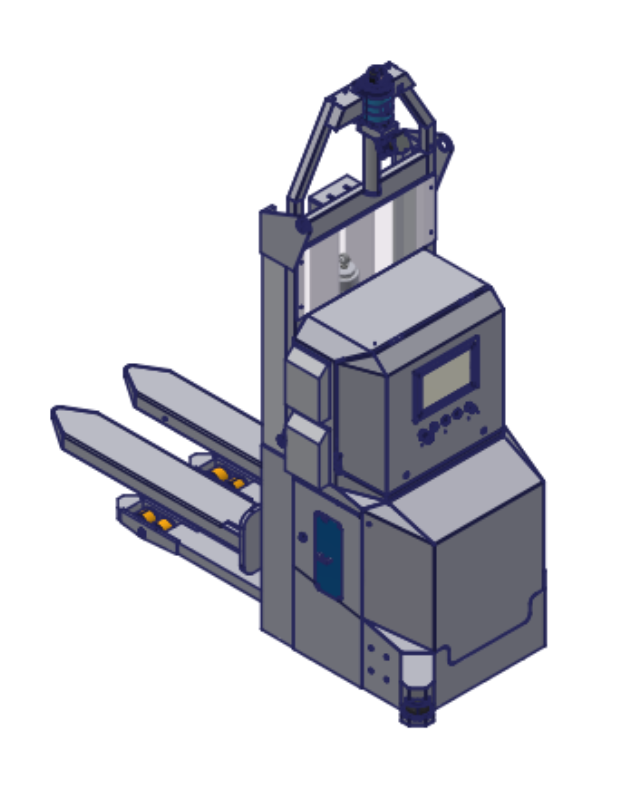}
         \caption{CAD Design of forklift}
         \label{fig:cad:a}
    \end{subfigure}
    \begin{subfigure}[b]{0.49\linewidth}
         \centering
         \includegraphics[width=1\linewidth]{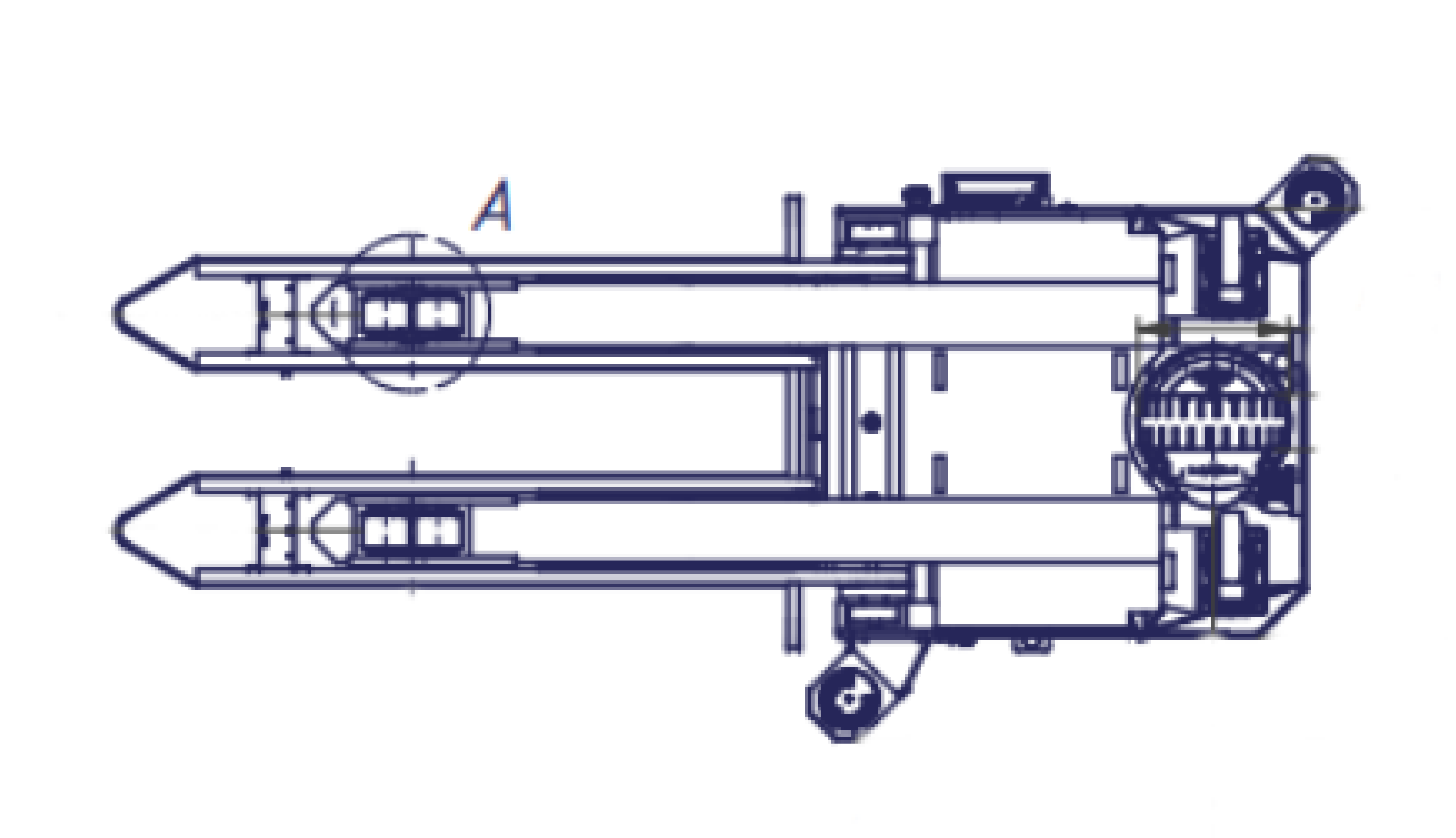}
         \caption{Wireframe of the forklift bottom view}
         \label{fig:forklift_spec:a}
    \end{subfigure}
    \begin{subfigure}[b]{0.5\linewidth}
         \centering
         \includegraphics[width=1\linewidth]{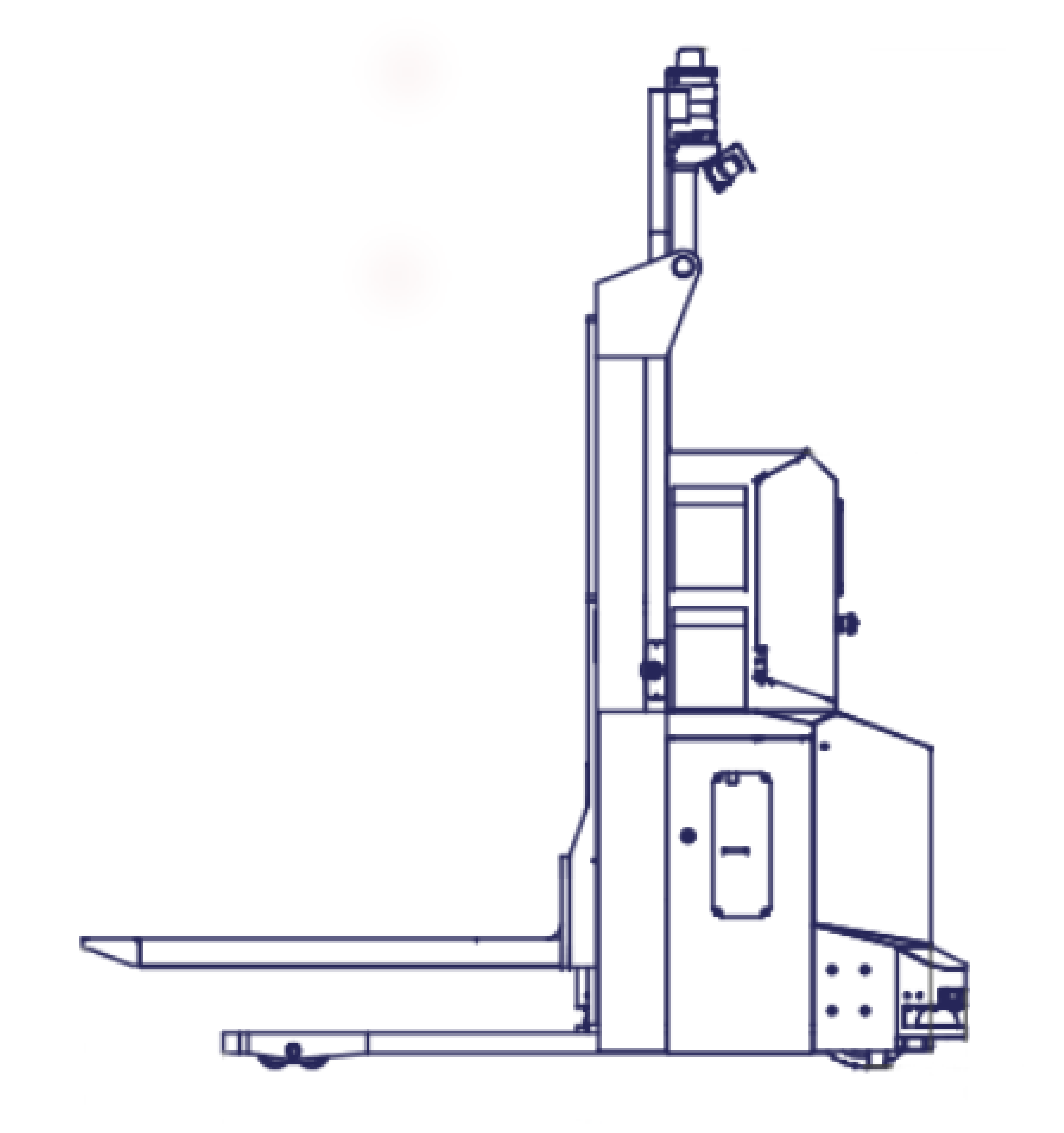}
         \caption{Wireframe of the forklift side view}
         \label{fig:forklift_spec:b}
    \end{subfigure}
    \caption{Real forklift used in the experiments}
    \label{fig:real_robot}
\end{figure}
As shown in the experimental setup, the wire-frame images in \autoref{fig:real_robot} of the robot used in this study were provided by \textbf{TAVIL IND}, where the thesis internship was conducted.

\begin{figure}[H]
    \centering
    \includegraphics[width=0.35\textwidth]{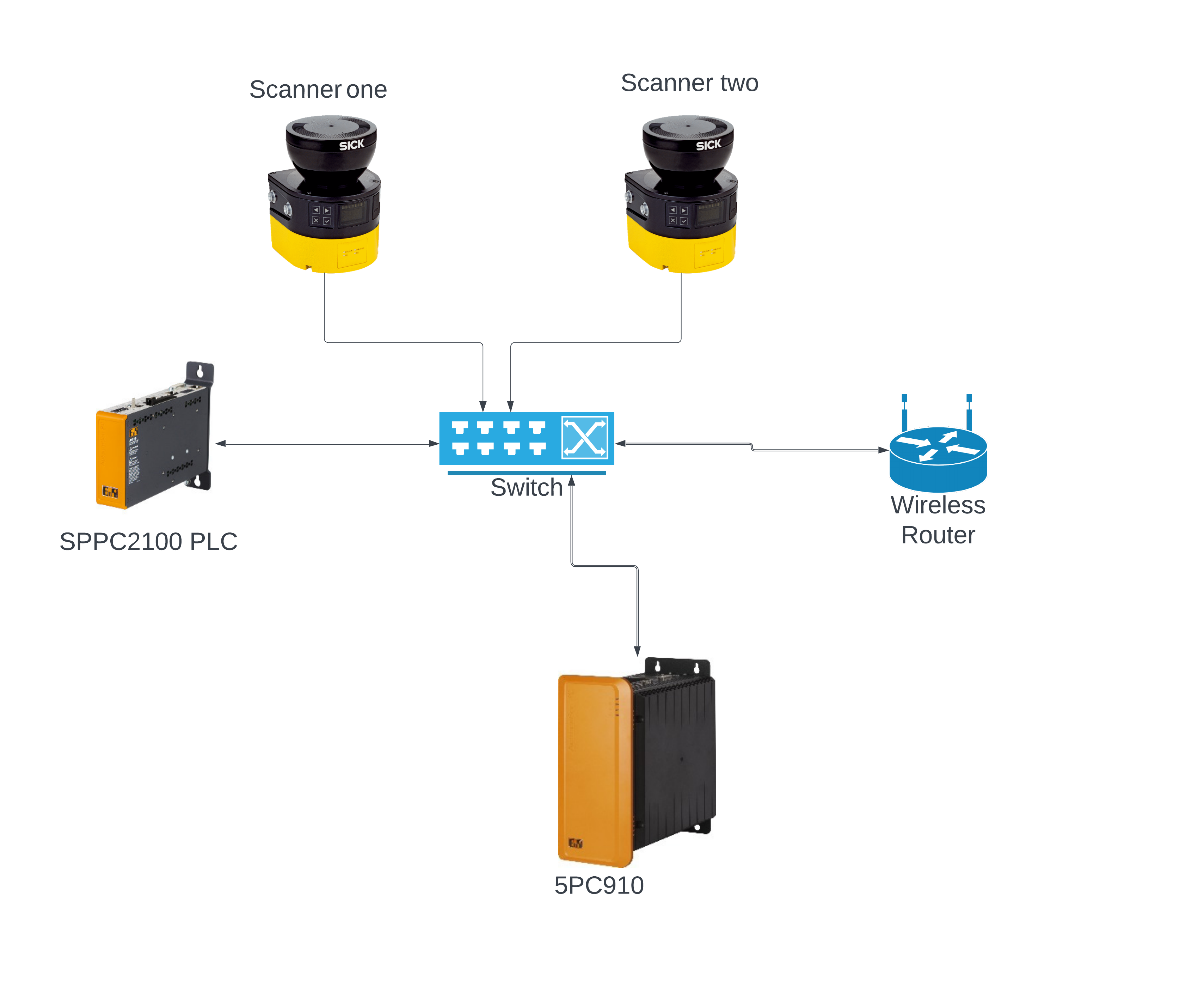}
    \caption{Hardware Configuration}
    \label{fig:hardware}
\end{figure}
\autoref{fig:hardware} shows the hardware configuration used in this study for testing in an industrial environment. The setup consists of a Programmable Logic Controller (PLC), a personal computer (PC), a network switch, and a wireless access point (AP). The \acrshort{plc} is responsible for controlling and coordinating the various components of the autonomous system. The PC serves as the main computational unit, running the developed algorithms and handling communication with the \acrshort{plc} and other devices. The network switch facilitates communication between the PC, \acrshort{plc}, and other devices, while the wireless AP provides connectivity for remote monitoring and control. This hardware configuration ensures efficient and reliable operation of the autonomous system during both simulation and real-world experiments.

The collected data from both simulation and real-world environments were used to evaluate the performance of the proposed system. Various metrics, such as accuracy, processing time, and robustness, were considered in the evaluation.

%% file: sections/results.tex
\section{Result and Discussion}
\label{sec:results}
In this section, the performance of the multi-object tracking and predictive collision avoidance system is evaluated and discussed. The experiments were conducted in both simulated and real-world environments. The results are presented in terms of the effectiveness of the object detection, tracking algorithms, and collision avoidance performance under various scenarios.

\subsection{Object Detection and Tracking Performance}
The Results of Object detection and representation were depicted in \autoref{fig:bounding_box} extracted from simulation and \autoref{fig:industrial_test} shows the results from real-world setup. Both showcase fairly accurate representations of tracked objects.

\autoref{fig:enkf_result} shows a comparison of position and velocity plots for the ensemble Kalman filter for two objects (vehicles), providing a visual representation of the tracking performance in different scenarios. From the plots, it is evident the filter is able to accurately track the positions of the objects in the environment.

The position plots show that the ensemble Kalman filter closely follows the true positions of the objects. However, the velocity plots reveal that the ensemble Kalman filter is capable of providing more reliable velocity estimates for the objects in the environment. This is particularly important for predictive collision avoidance, as accurate velocity estimates are crucial for determining the likelihood of future collisions and planning appropriate avoidance maneuvers.

\begin{figure}[H]
         \centering
    \begin{subfigure}[t]{0.32\textwidth}
         \centering
         \includegraphics[width=1\textwidth, height=0.7\textwidth]{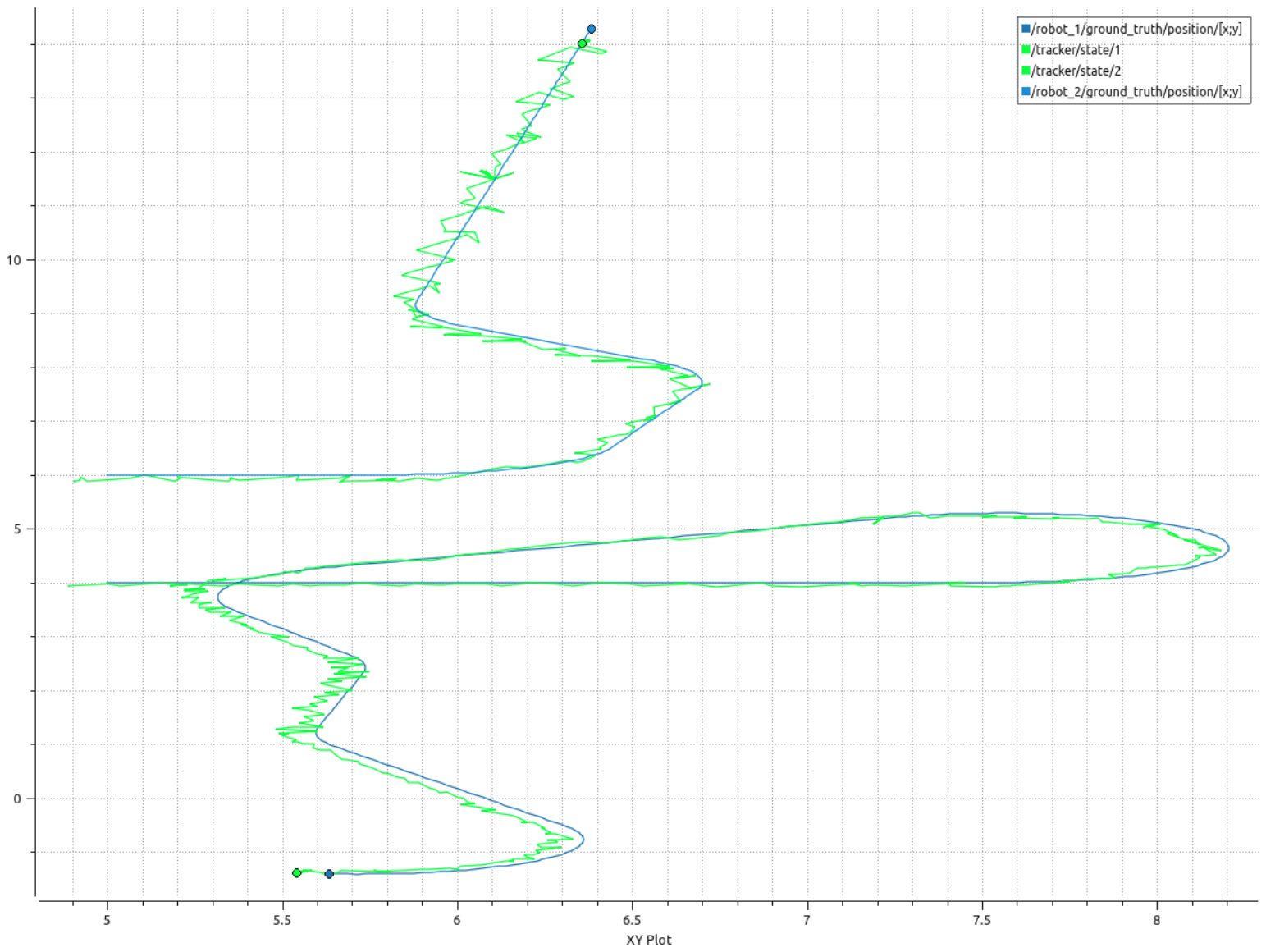}
         \caption{\textit{Tracked Objects(one and two) estimated position(x,y)(green) with ground truth(blue) in meters}}
    \end{subfigure}
\end{figure}
\begin{figure}[H]
    \ContinuedFloat
         \centering
    \begin{subfigure}[t]{0.32\textwidth}
         \centering
         \includegraphics[width=1\textwidth, height=0.7\textwidth]{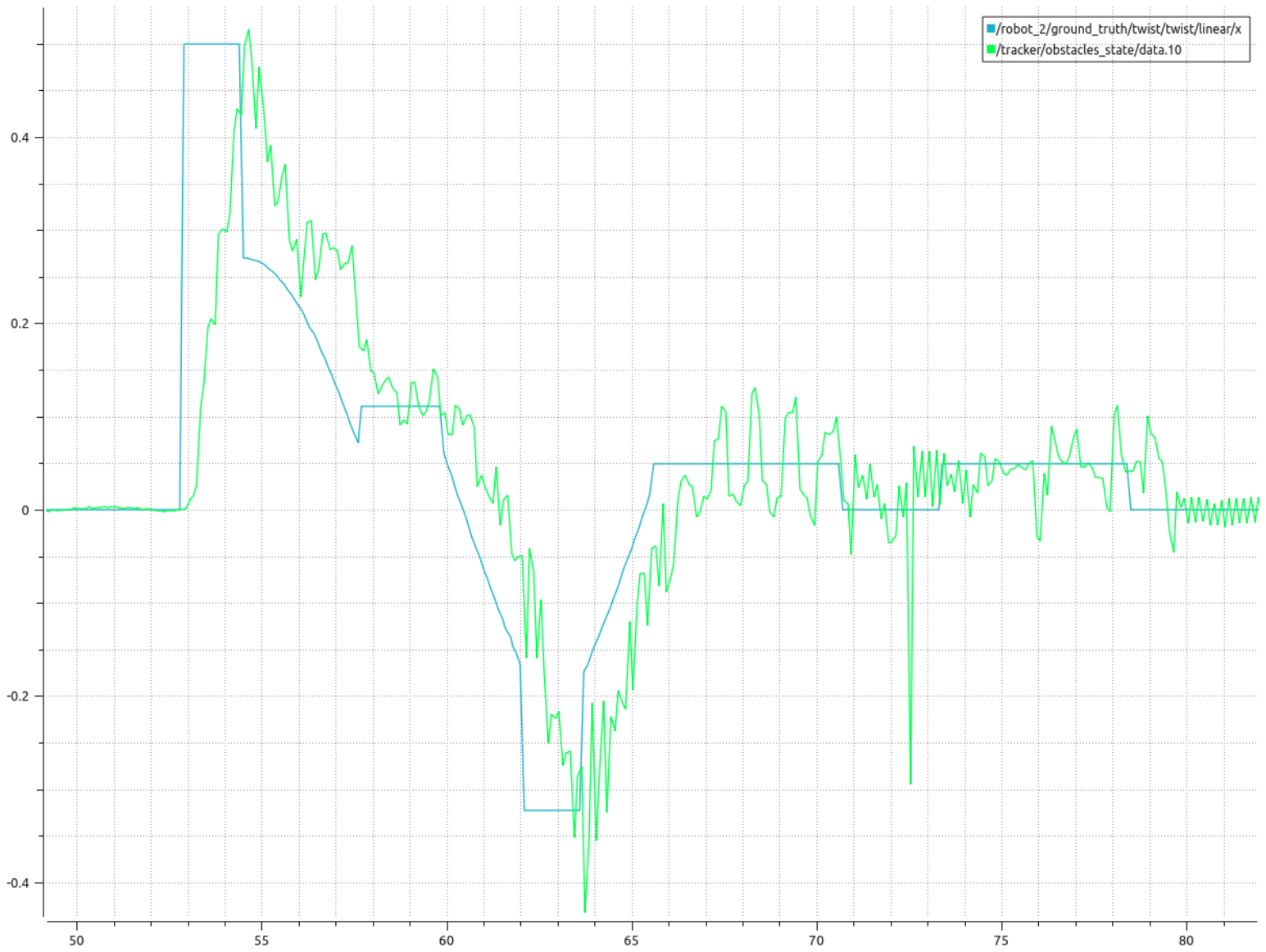}
         \caption{\textit{Tracked object one estimated velocity($\dot{x}, \dot{y}$)(green) with ground truth(blue) in m/s}}
    \end{subfigure}
\end{figure}
\begin{figure}[H]
    \ContinuedFloat
         \centering
    \begin{subfigure}[t]{0.32\textwidth}
         \centering
         \includegraphics[width=1\textwidth, height=0.7\textwidth]{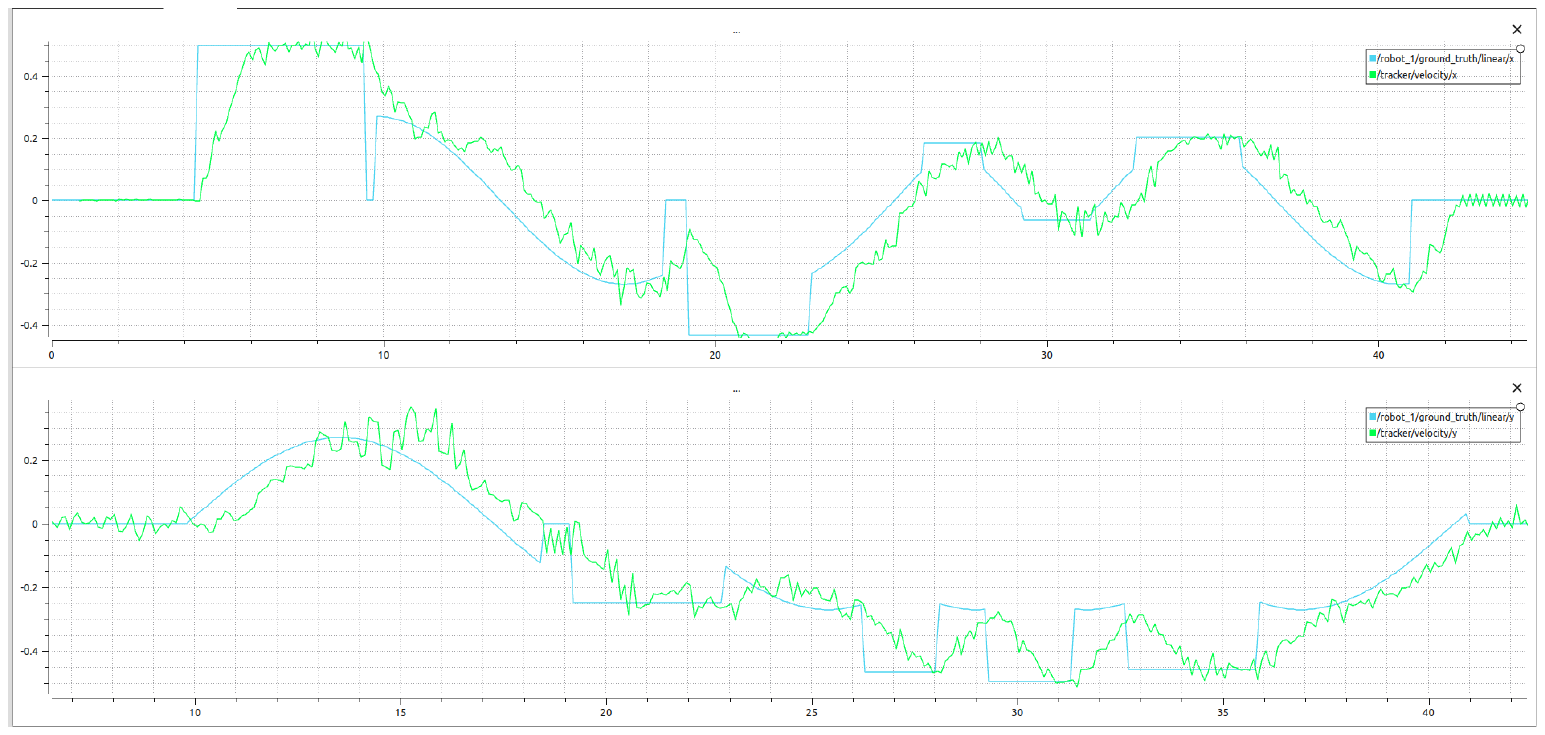}
         \caption{\textit{Tracked object two estimated velocity($\dot{x}, \dot{y}$)(green) with ground truth(blue) in m/s}}
    \end{subfigure}
    \caption{\textit{Result of Ensemble Kalman tracker with greedy data association in Stage Simulator}}
    \label{fig:enkf_result}
\end{figure}

Overall, the results of this analysis suggest that the ensemble Kalman filter is a promising solution for multi-object tracking and predictive collision avoidance in complex environments. Its ability to better handle non-linearities and uncertainties in the system dynamics can lead to improved tracking performance and more reliable collision predictions, making it a valuable tool for ensuring the safety and efficiency of autonomous systems.

\subsection{Predictive Collision Avoidance Performance}
\begin{figure}[H]
    \begin{subfigure}[t]{0.48\linewidth}
         \centering
         \includegraphics[width=1.0\textwidth]{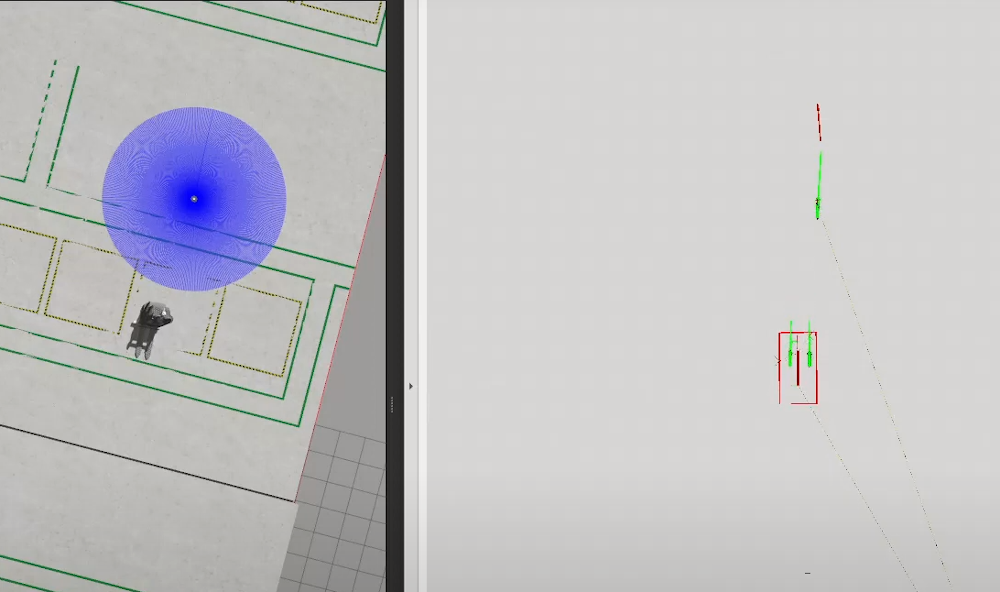}
         \caption{\textit{robot going in a straight line at a safe distance}}
    \end{subfigure}
    \begin{subfigure}[t]{0.48\linewidth}
         \centering
         \includegraphics[width=1.0\textwidth]{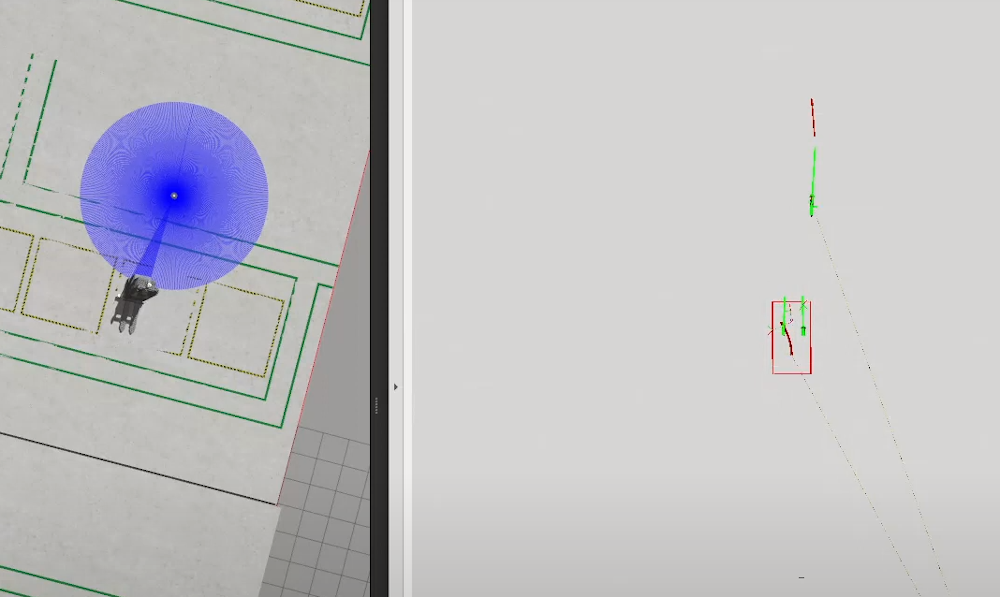}
         \caption{\textit{robot in front stops moving ego robot starts to change direction}}
    \end{subfigure}
    \begin{subfigure}[t]{0.48\linewidth}
         \centering
         \includegraphics[width=1.0\textwidth]{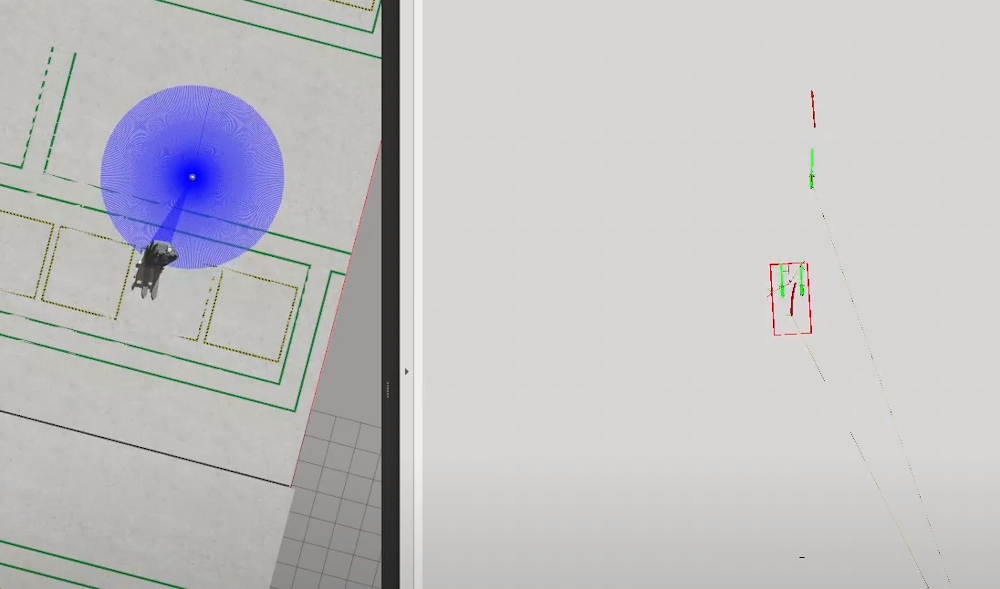}
         \caption{\textit{robot in front starts to move ego robot starts to change back direction to straight line}}
    \end{subfigure}
    \caption{\textit{Results of first scenario. Path predicted by the tracker(green), red box(robot footprint), and the red path(scored by the controller) \href{https://www.youtube.com/watch?v=1jSHrXle1zQ&ab_channel=FieldRobotics}{full video} }}
    \label{fig:case_1}
\end{figure}

Multiple scenarios were considered for testing the predictive collision avoidance algorithm. Here three scenarios will be discussed. In the first scenario, shown in \autoref{fig:case_1}, an autonomous robot is in front of the ego robot, both moving in the same direction (please check the link to the video in the caption). In this case, the ego robot should neither stop nor change direction, provided that the robot in front maintains a constant velocity and a safe distance.

\begin{figure}[H]
    \begin{subfigure}[t]{0.48\linewidth}
         \centering
         \includegraphics[width=1.0\textwidth]{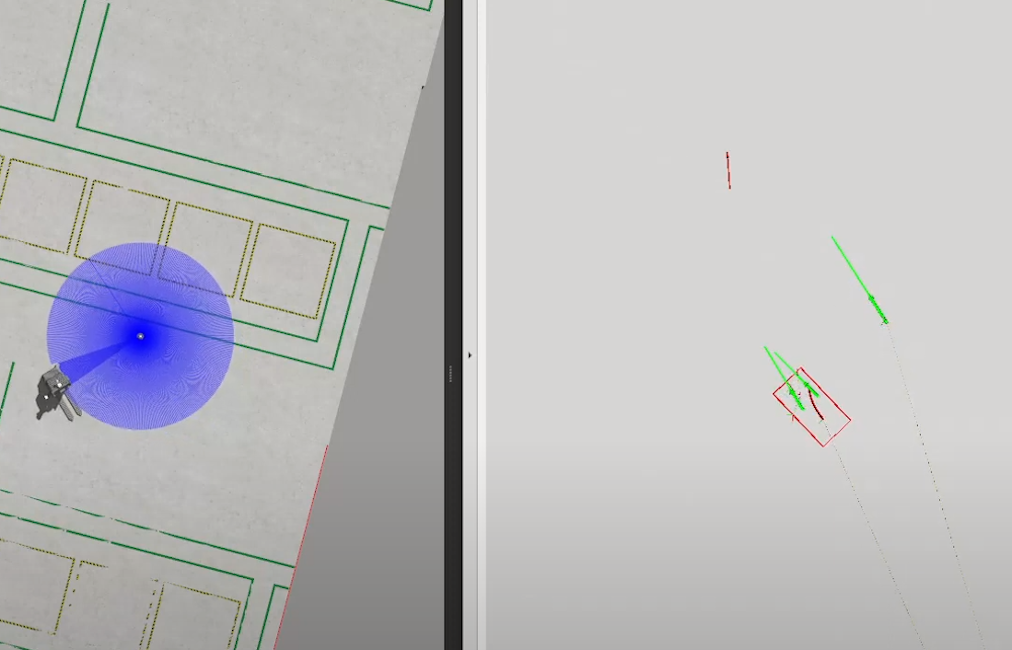}
         \caption{\textit{robot not affected by the other vehicle to the side}}
    \end{subfigure}
    \begin{subfigure}[t]{0.48\linewidth}
         \centering
         \includegraphics[width=1.0\textwidth]{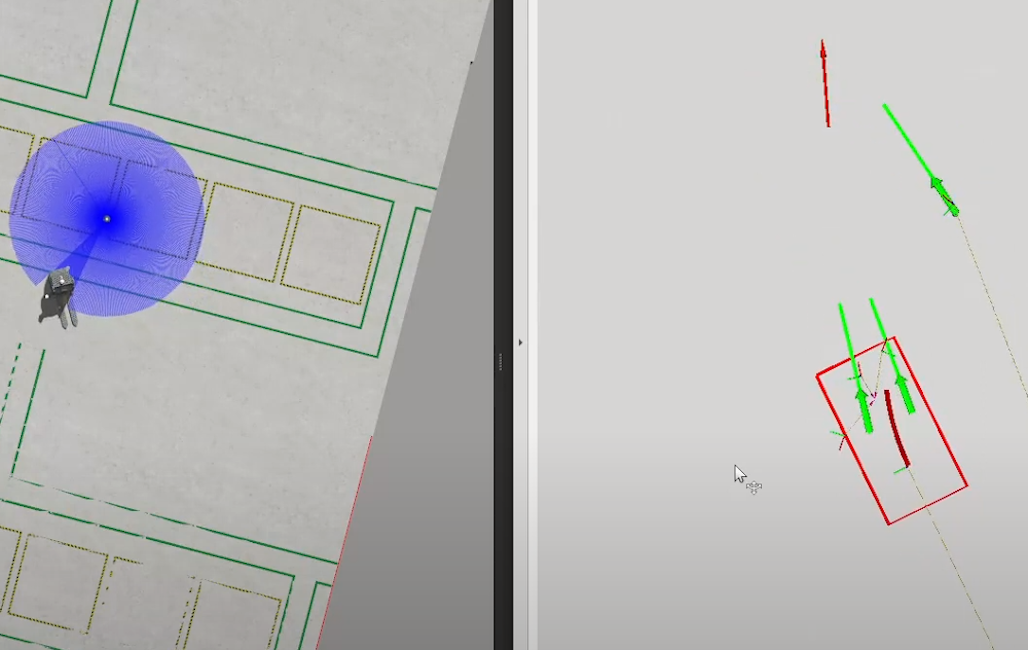}
         \caption{\textit{robot continues to choose the path that looks close to the other vehicle}}
    \end{subfigure}
    \begin{subfigure}[t]{0.48\linewidth}
         \centering
         \includegraphics[width=1.0\textwidth]{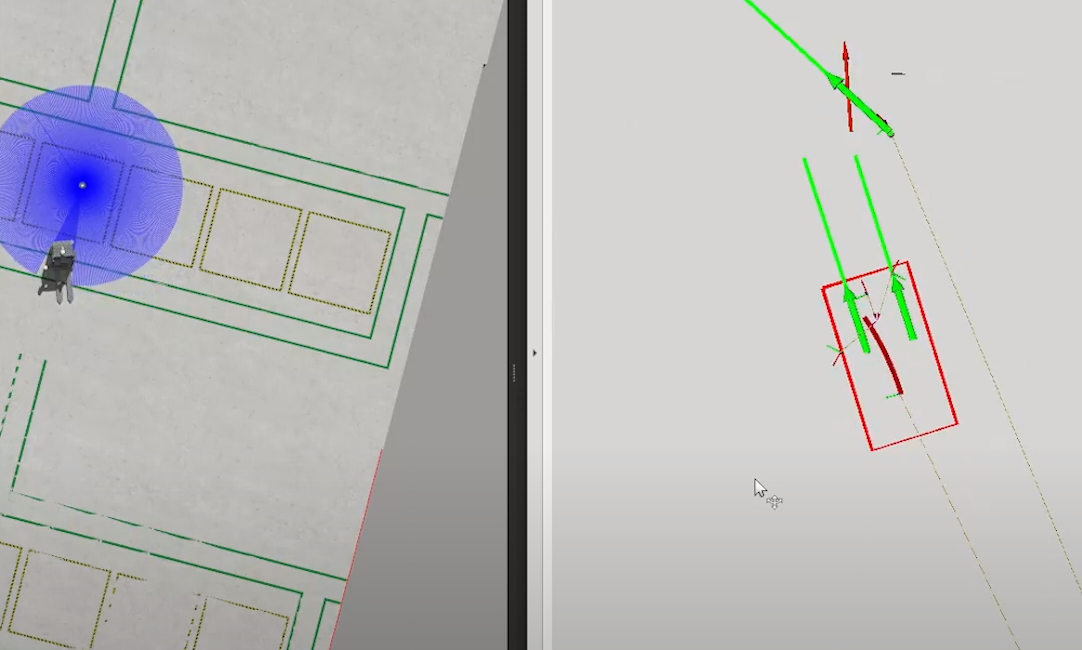}
         \caption{\textit{robot chooses to divert since the other robot was too close to the safe distance threshold}}
    \end{subfigure}
    \begin{subfigure}[t]{0.48\linewidth}
         \centering
         \includegraphics[width=1.0\textwidth]{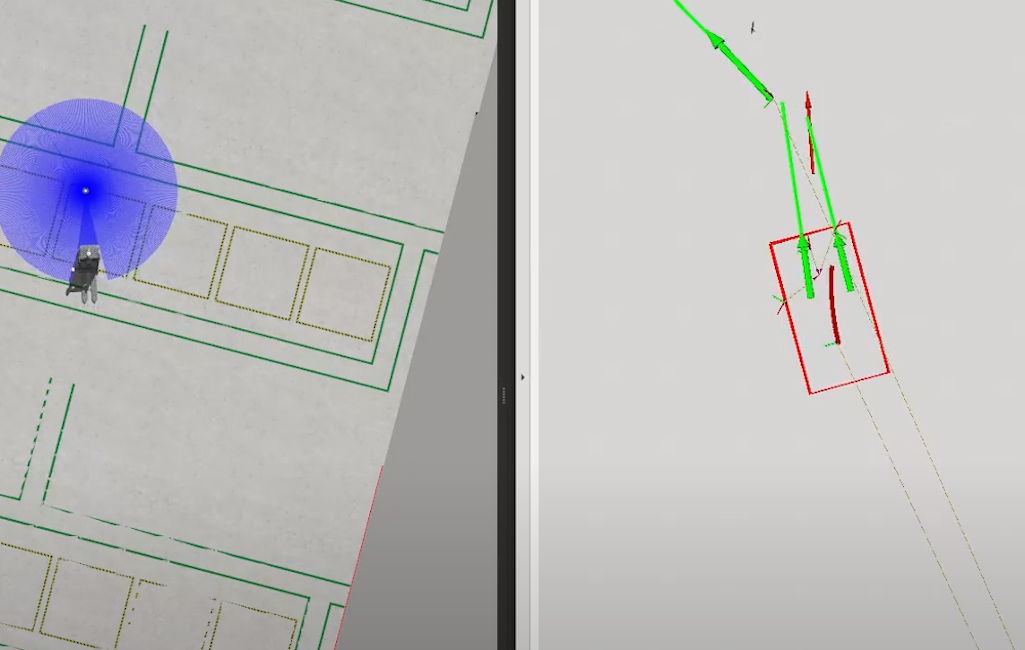}
         \caption{\textit{robot continues back to the optimal path}}
    \end{subfigure}
    \caption{\textit{Results of second scenario. Path predicted by the tracker(green), red box(robot footprint), and the red path(scored by the controller) \href{https://www.youtube.com/watch?v=i_dFthUnMCg&ab_channel=FieldRobotics}{full video} }}
    \label{fig:case_2}
\end{figure}

The second scenario, in \autoref{fig:case_2} which was illustrated in \autoref{fig:obstacle}, features a vehicle crossing the ego robot's path. In this situation, the controller selects a velocity that avoids collision by steering toward the current position of the other vehicle since that position will be unoccupied in the future.

The second scenario involves another vehicle moving in the opposite direction to the ego vehicle, which would result in a head-on collision. In this case, the ego robot initially chooses to reverse in order to avoid the collision, as shown in \autoref{fig:case_31}. To examine the algorithm's performance, a penalty was added to discourage reverse motion whenever possible. Consequently, the ego vehicle diverges to the side to avoid the collision, as depicted in \autoref{fig:case_32}. Both outcomes are expected and demonstrate the adaptability of the algorithm.

\begin{figure}[H]
    \begin{subfigure}[t]{0.48\linewidth}
         \centering
         \includegraphics[width=1.0\textwidth]{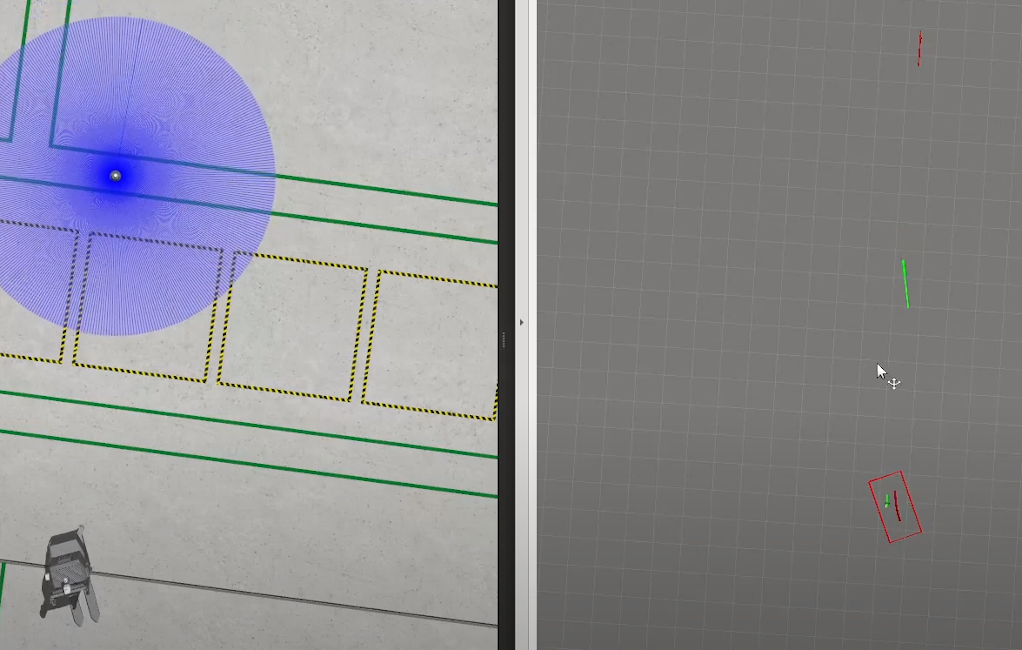}
         \caption{\textit{robot not affected by the other vehicle Starts to move}}
    \end{subfigure}
    \begin{subfigure}[t]{0.48\linewidth}
         \centering
         \includegraphics[width=1.0\textwidth]{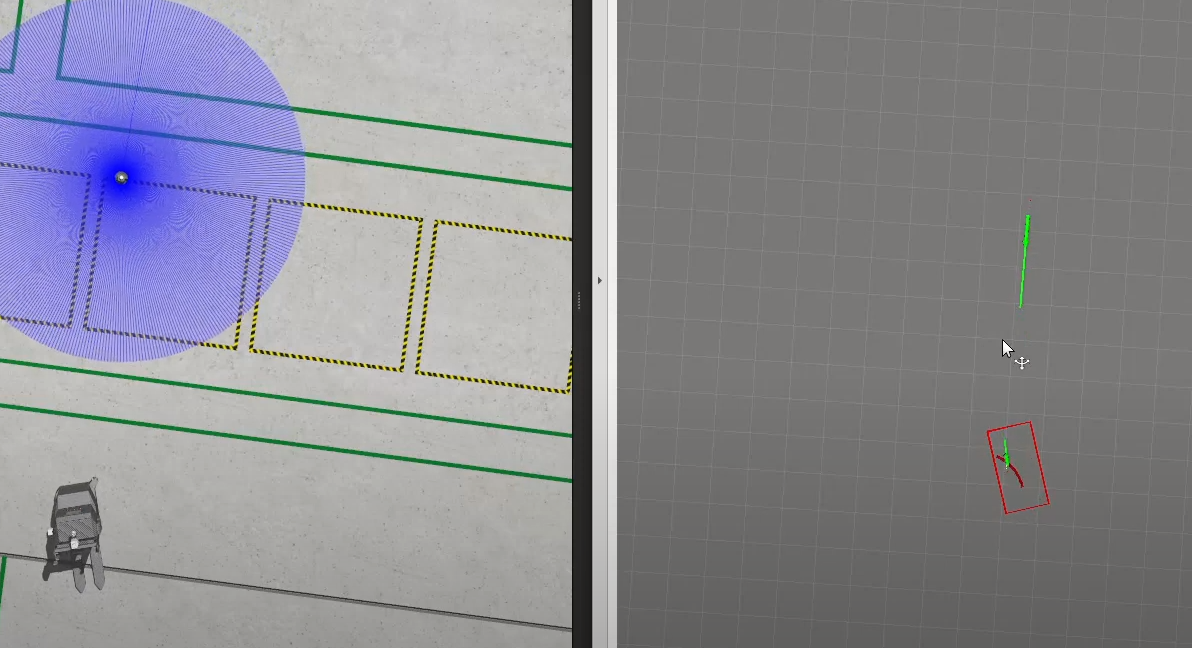}
         \caption{\textit{robot starts to move to left to avoid ultimate collision}}
    \end{subfigure}
\end{figure}
\begin{figure}[H]
    \centering
    \ContinuedFloat
    \begin{subfigure}[t]{0.48\linewidth}
         \centering
         \includegraphics[width=1.0\textwidth]{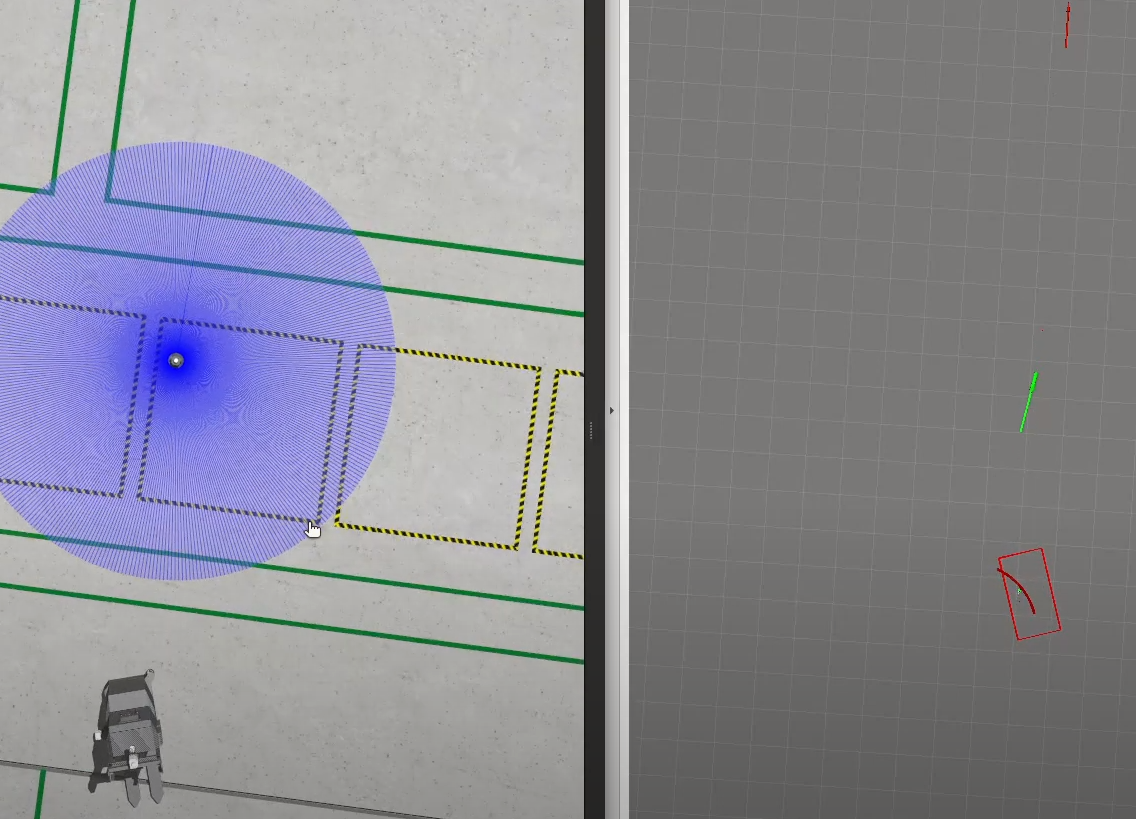}
         \caption{\textit{robot continues moving to left to avoid ultimate collision}}
    \end{subfigure}
    \begin{subfigure}[t]{0.48\linewidth}
         \centering
         \includegraphics[width=1.0\textwidth]{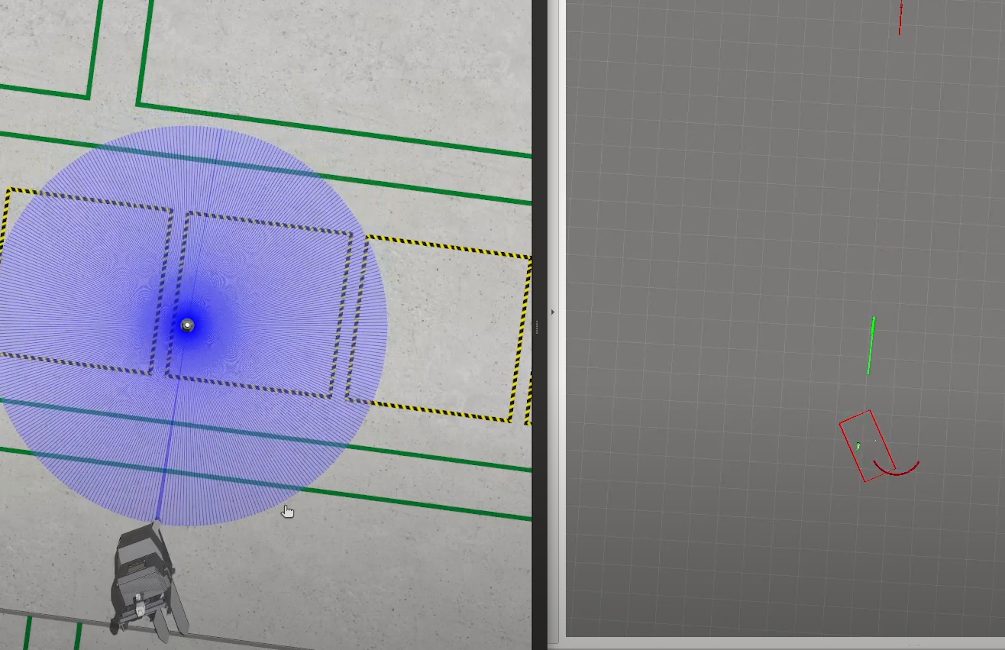}
         \caption{\textit{robot changes to reverse to avoid collision}}         
    \end{subfigure}
\end{figure}
\begin{figure}[H]
    \centering
    \ContinuedFloat
    \begin{subfigure}[t]{0.48\linewidth}
         \centering
         \includegraphics[width=1.0\textwidth]{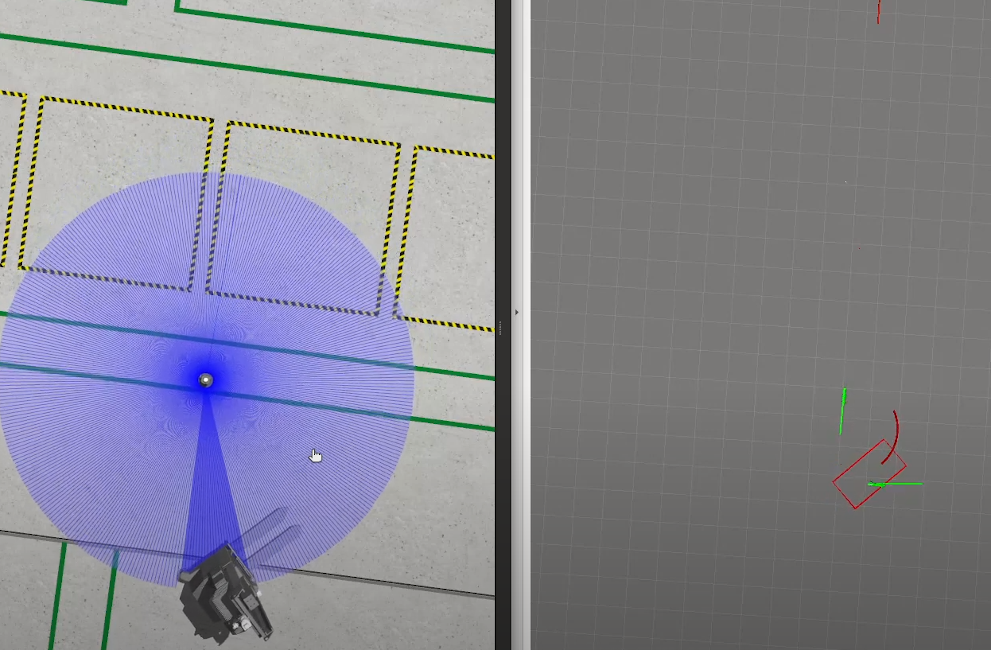}
         \caption{\textit{robot continues to go in reverse}}
    \end{subfigure}
    \hfill
    \begin{subfigure}[t]{0.48\linewidth}
         \centering
         \includegraphics[width=1.0\textwidth]{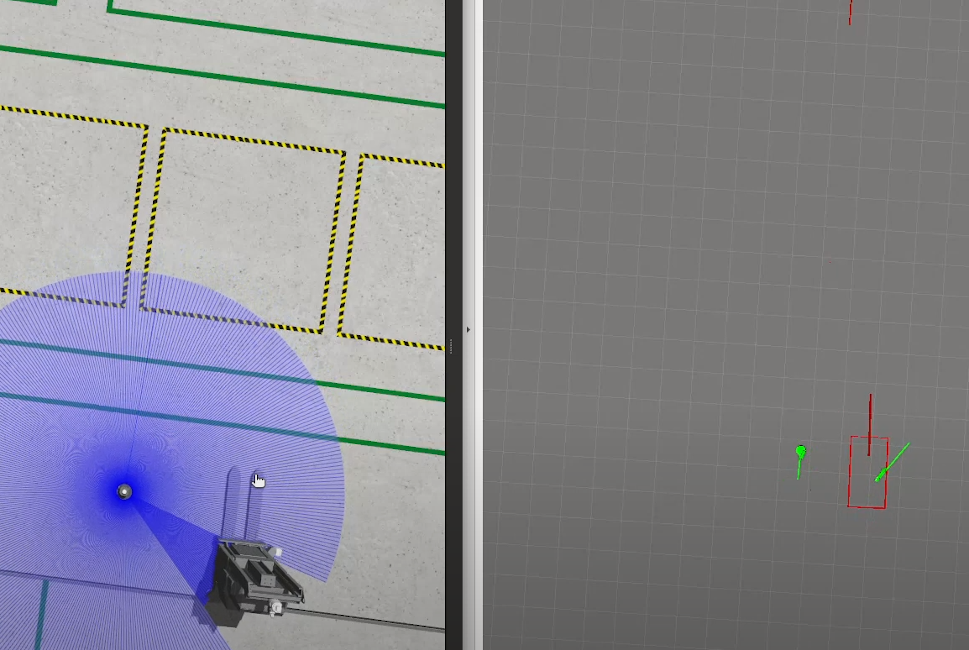}
         \caption{\textit{robot continues going to goal collision avoided}}
    \end{subfigure}\\
    \caption{\textit{Results of the modified \glsxtrshort{dwa} applied to the third scenario. Path predicted by the tracker(green), red box(robot footprint), and the red path(scored by the controller) with a small reverse direction penalty \href{https://youtu.be/qhcZLr5zXbk}{full video} }}
    \label{fig:case_31}
\end{figure}

\begin{figure}[H]
    \begin{subfigure}[t]{0.48\linewidth}
         \centering
         \includegraphics[width=1.0\textwidth]{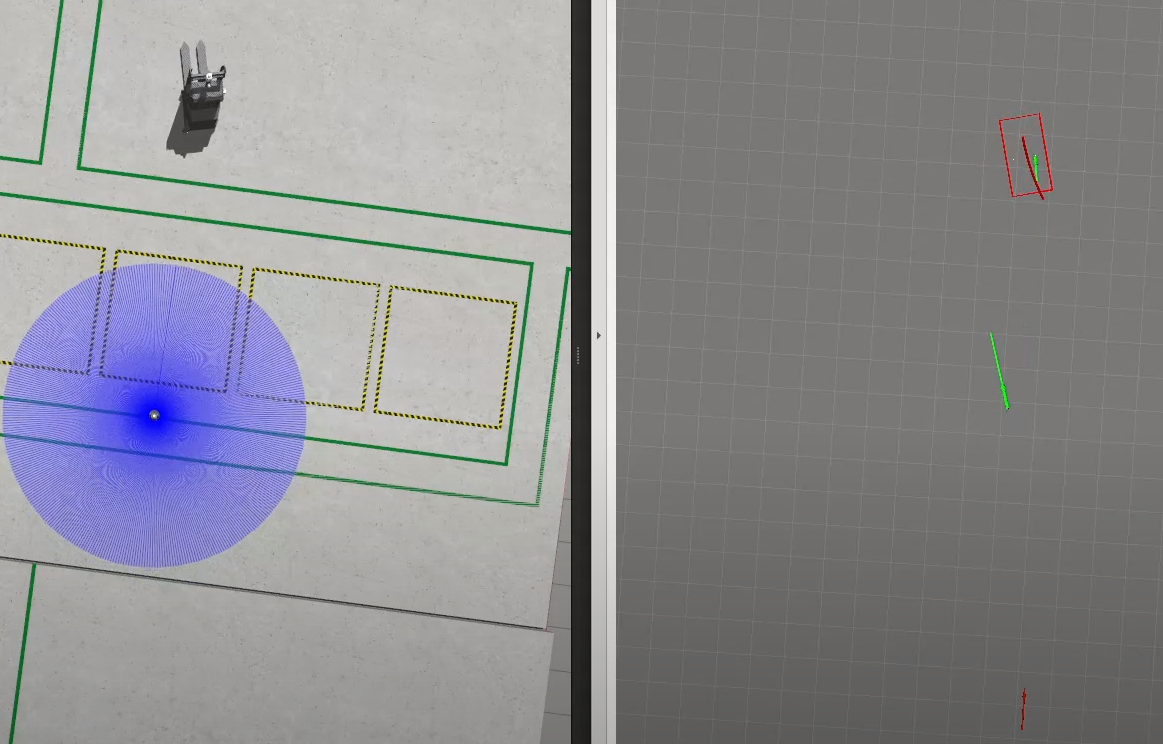}
         \caption{\textit{robot starts to move to left to avoid ultimate collision}}
    \end{subfigure}
    \begin{subfigure}[t]{0.48\linewidth}
         \centering
         \includegraphics[width=1.0\textwidth]{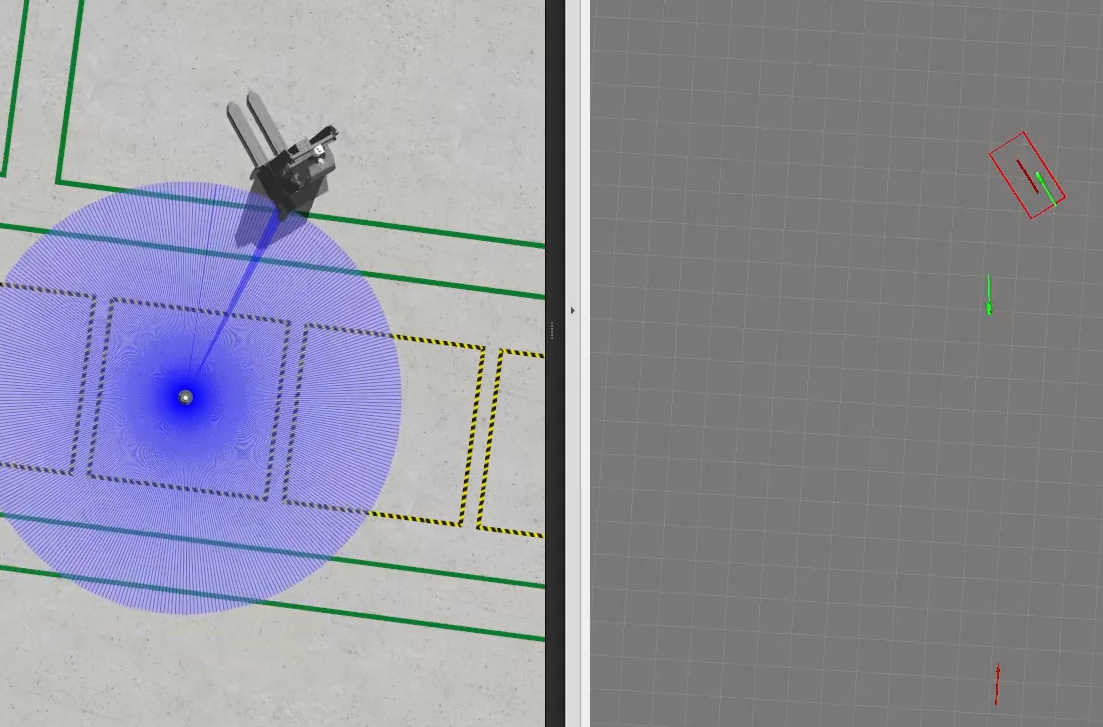}
         \caption{\textit{robot continues moving to left to avoid ultimate collision}}
    \end{subfigure}
\end{figure}
\begin{figure}[H]
    \centering
    \ContinuedFloat
    \begin{subfigure}[t]{0.48\linewidth}
         \centering
         \includegraphics[width=1.0\textwidth]{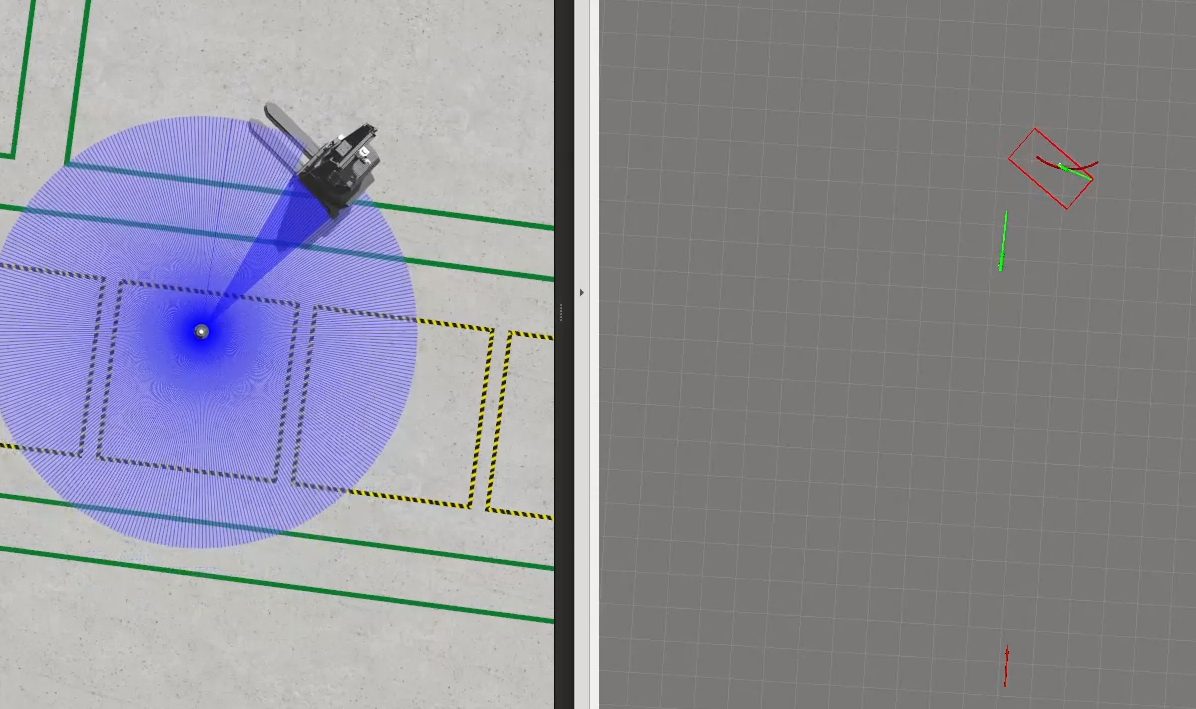}
         \caption{\textit{robot continues moving to left to avoid ultimate collision}}     
    \end{subfigure}
    \begin{subfigure}[t]{0.48\linewidth}
         \centering
         \includegraphics[width=1.0\textwidth]{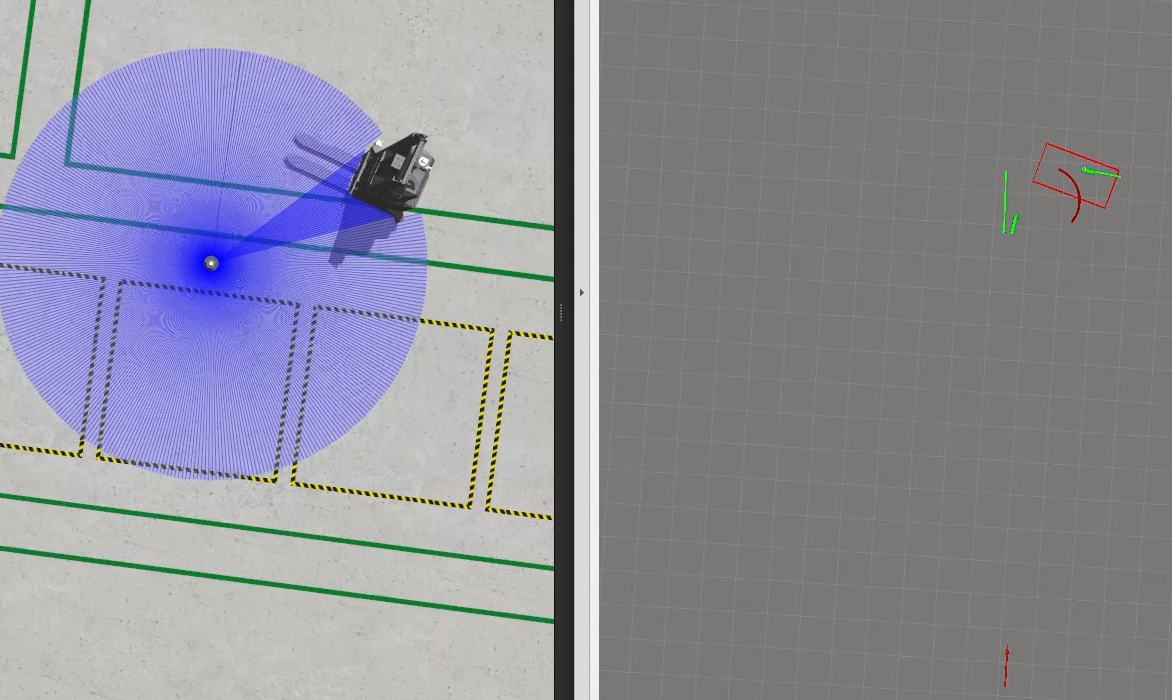}
         \caption{\textit{robot continues going to goal collision avoided}}
    \end{subfigure}
    \begin{subfigure}[t]{0.48\linewidth}
         \centering
         \includegraphics[width=1.0\textwidth]{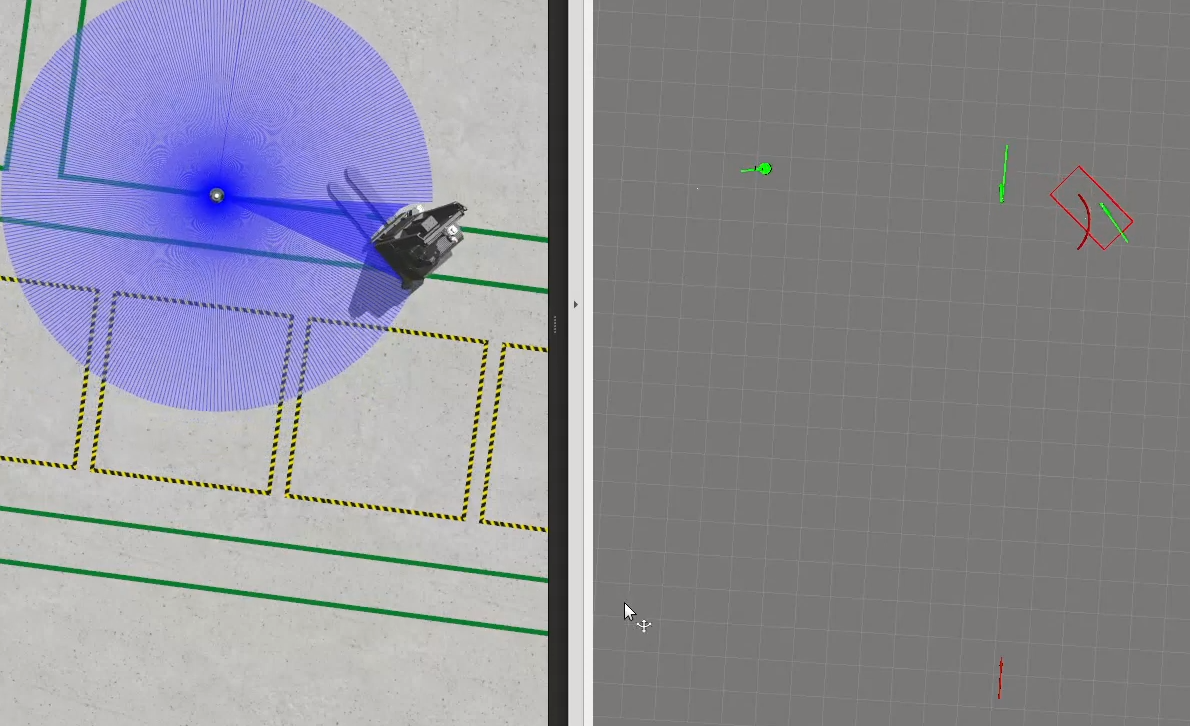}
         \caption{\textit{robot continues going to goal collision avoided}}
    \end{subfigure}
    \hfill
    \begin{subfigure}[t]{0.48\linewidth}
         \centering
         \includegraphics[width=1.0\textwidth]{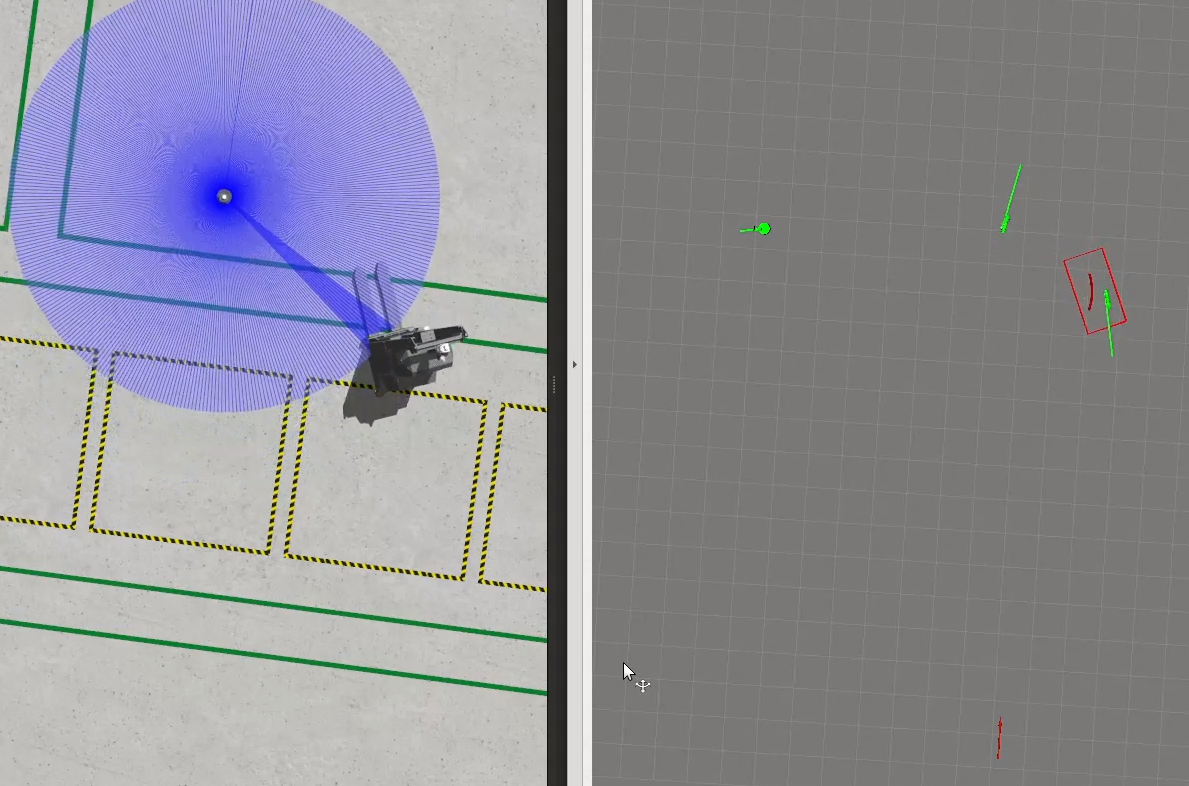}
         \caption{\textit{robot continues going to goal collision avoided}}
    \end{subfigure}
    \caption{\textit{Results of third scenario. Path predicted by the tracker(green), red box(robot footprint), and in red path scored by the controller with a large reverse direction penalty \href{https://youtu.be/ERIKxRJEduA}{full video} }}
    \label{fig:case_32}
\end{figure}

\subsubsection{Time Response}
The response time of a controller plays a crucial role in mobile robotics, as it directly affects the system's overall performance. A fast controller response time ensures that the robot can react quickly to environmental changes, allowing it to avoid obstacles and navigate safely and efficiently. In applications such as manufacturing, logistics, and agriculture, where the robot operates in dynamic environments with moving objects, a rapid controller response time is particularly critical to ensure both the safety of the robot and its surroundings, as well as the efficiency of the overall operation.

\begin{figure}[H]
\centering
\includegraphics[width=0.8\linewidth]{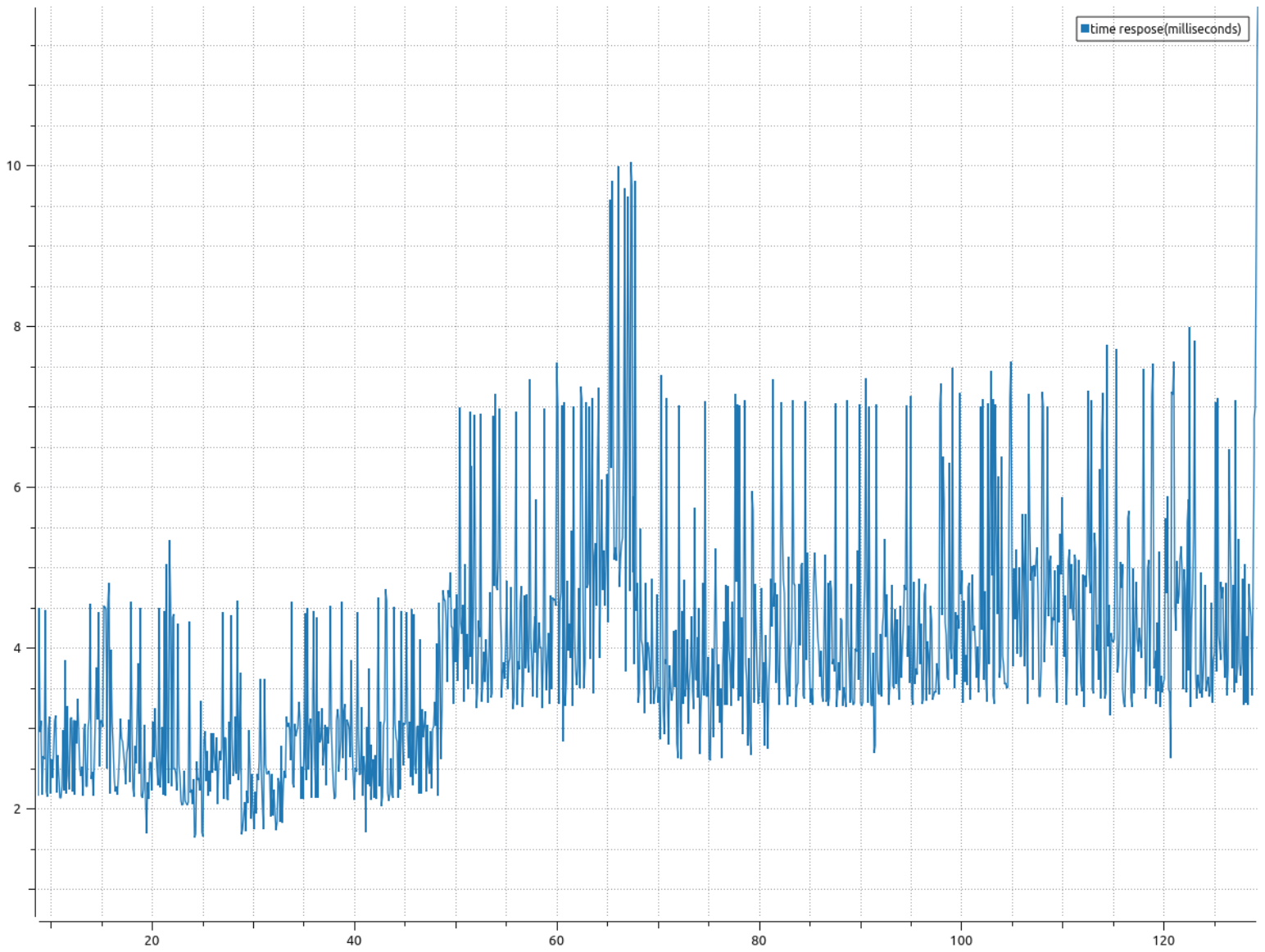}
\caption{\textit{Controller time response plot in milliseconds.}}
\label{fig:controller_time_response}
\end{figure}

This study measured and analyzed the controller response time to evaluate its performance. As depicted in \autoref{fig:controller_time_response}, the controller response time was consistently found to be less than 10 ms. This fast response time demonstrates the effectiveness of the developed algorithms and techniques in providing real-time trajectory sampling and collision avoidance for autonomous mobile robots.

\begin{figure}[H]
    \centering
    \includegraphics[width=0.8\linewidth]{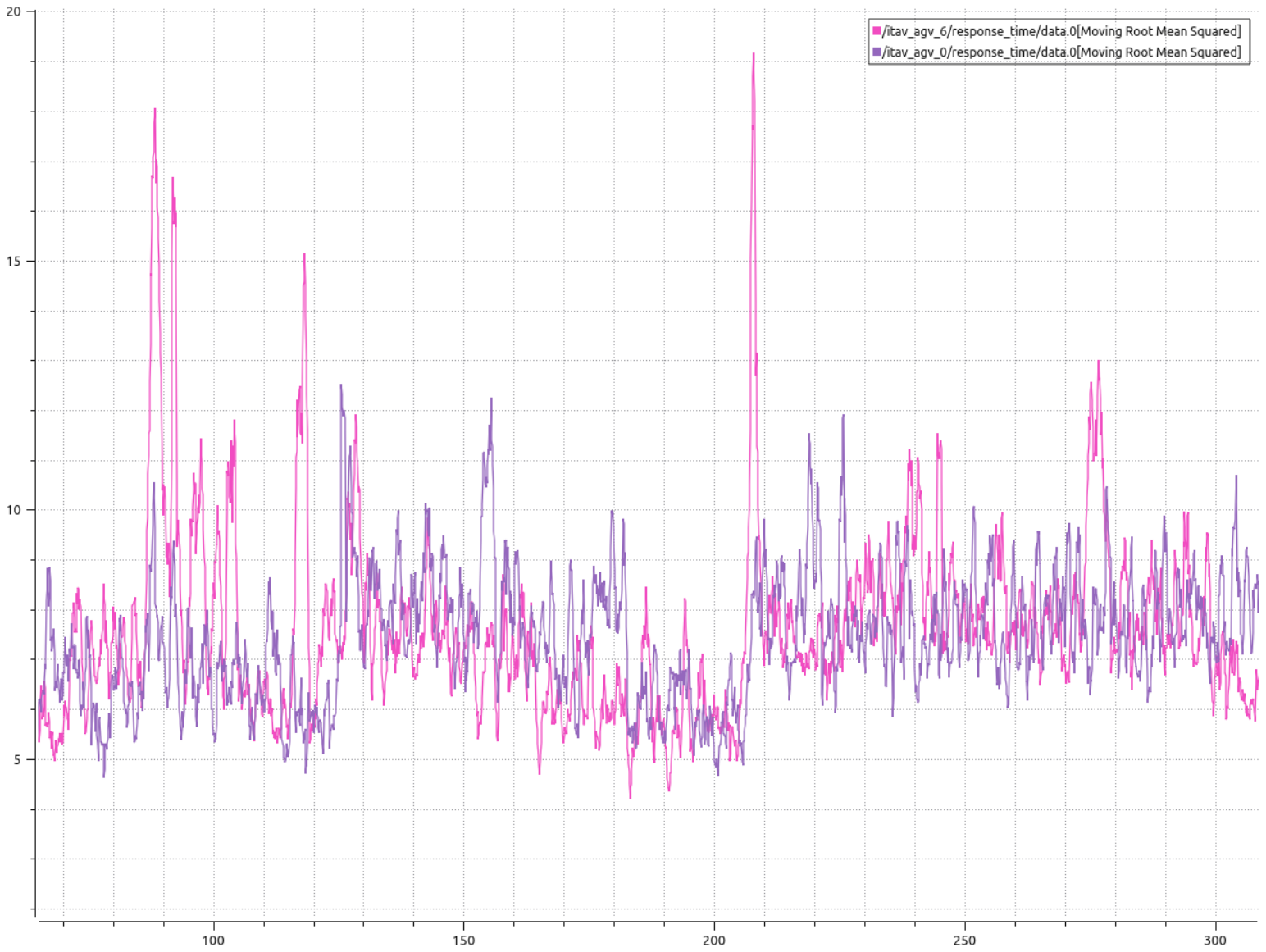}
    \caption{\textit{time response(milliseconds) for two robots running indpendently i.e. reminder this is running in one device} }
    \label{fig:time_response_gazebo_two}
\end{figure}

\begin{figure}[H]
    
    \begin{subfigure}[t]{0.45\linewidth}
         \centering
         \includegraphics[width=1.0\linewidth]{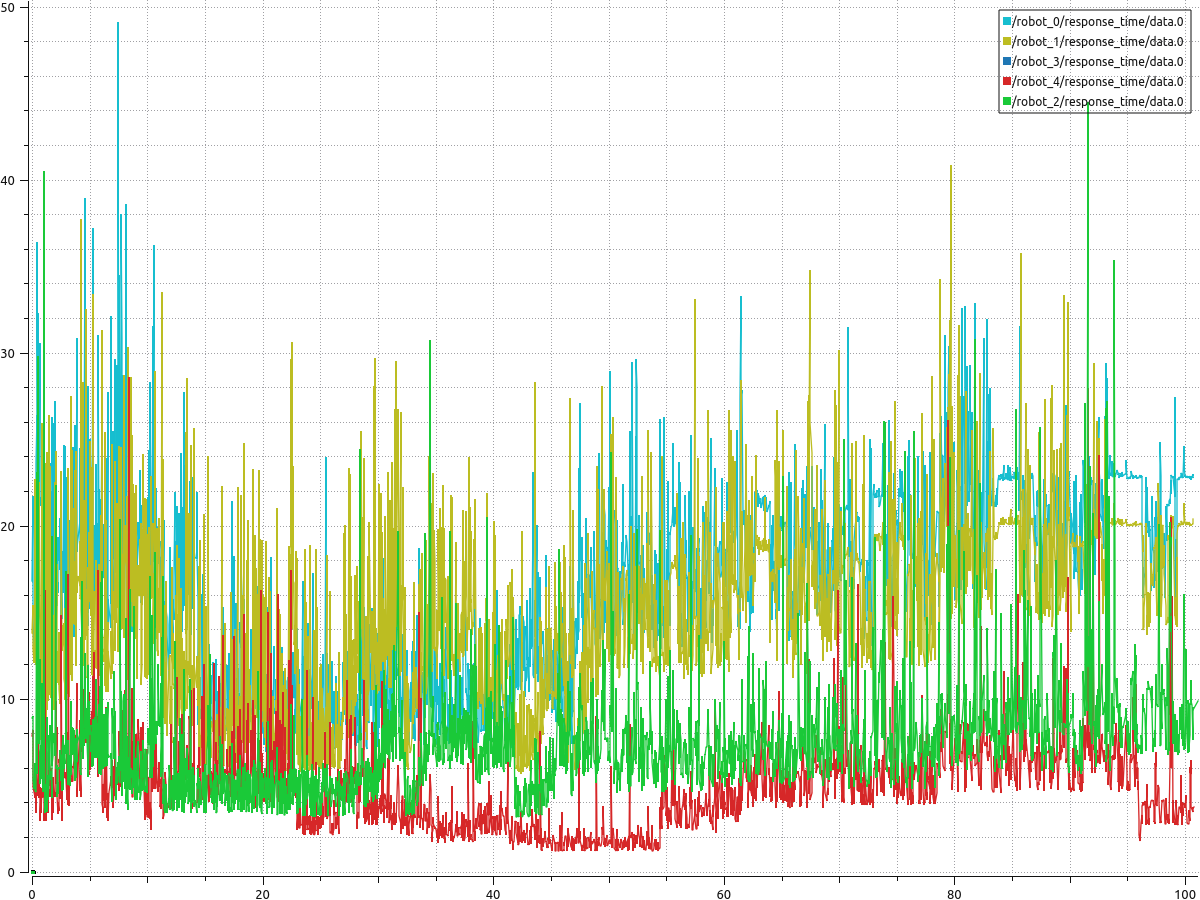}
         \caption{\textit{Five robots using the same hardware resource controller and their time response(milliseconds) }}
    \end{subfigure}
    \hfill
    \begin{subfigure}[t]{0.45\linewidth}
         \centering
         \includegraphics[width=1.0\linewidth]{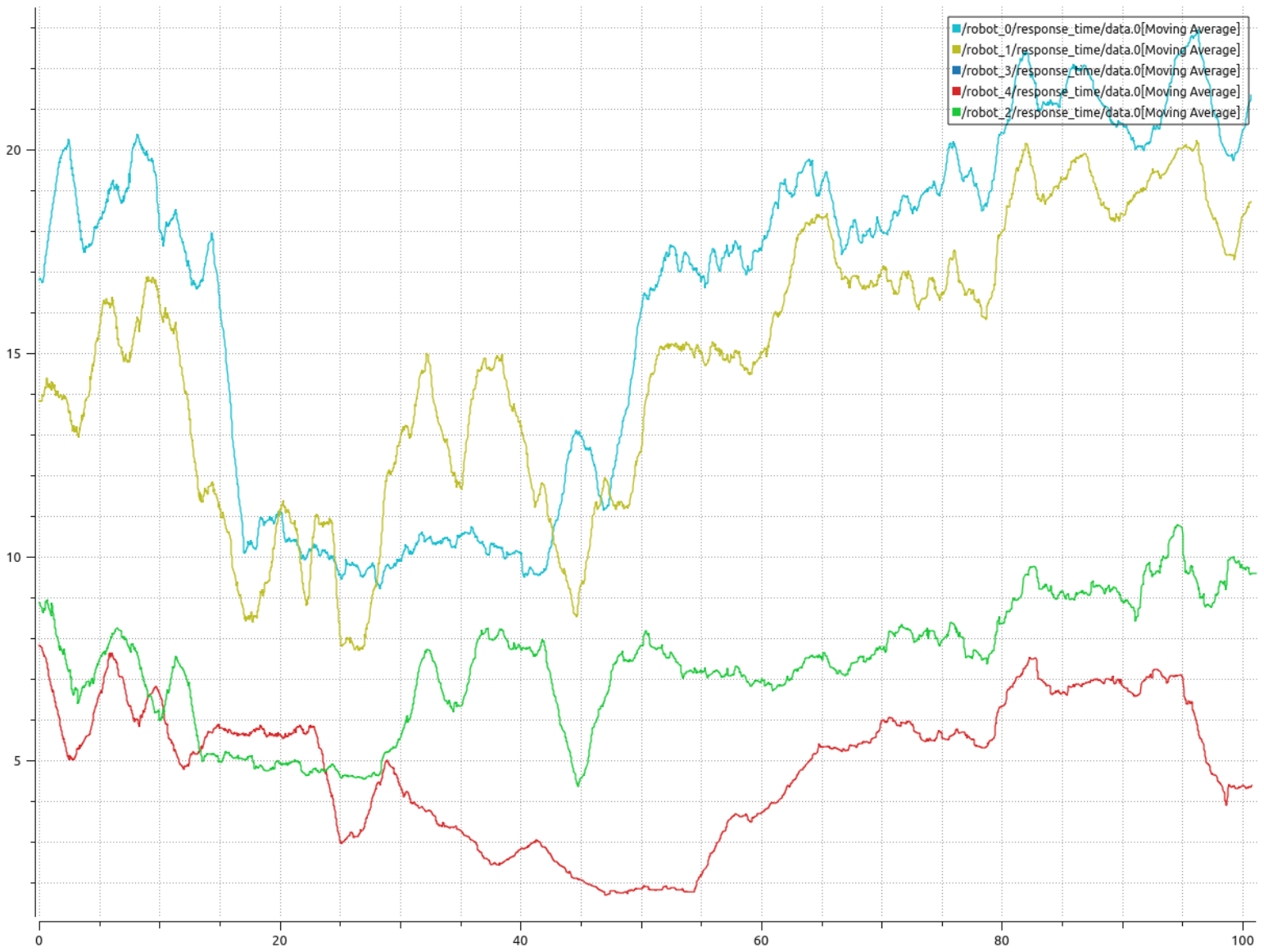}
         \caption{\textit{moving average applied to the left plot for a better view of the result}}
    \end{subfigure}
    \caption{\textit{Result of Five Pioneer robots in Stage simulator to showcase the Controller time response}}
    \label{fig:stage_five_robots}
\end{figure}

The quick response time of the controller is a significant advantage for the proposed multi-object tracking and predictive collision avoidance system, as it contributes to the safe and efficient operation of autonomous mobile robots in complex and dynamic environments. Multiple tests were done in the gazebo and stage to showcase the controller response time. In the gazebo, using two robots in the same hardware resource \autoref{fig:time_response_gazebo_two} shows the time response of the controllers, and another test in Stage simulator \autoref{fig:stage_five_robots} shows slower controller response due to hardware constraint. In both scenarios, we see the controllers for each robot working under the time required for mobile robots. But since the same hardware resources are being shared, we see an increased delay in the controllers. This is expected as each robot is expected to have its own hardware. An empty map of size 200x200m with a resolution of 0.05 was used; this is a very large map. In a practical scenario, the robot doesn't need the whole map hence the speed can be further improved by using a smaller local map.

\subsection{Real-world Experiments}
To validate the performance of the predictive collision avoidance algorithm in a real-world environment with a more complex and dynamic obstacle, scenario one was tested using a human moving in front of the ego robot. In this scenario, the human moves in the same direction as the ego robot, occasionally stopping and starting again. The ego robot is expected to adjust its velocity accordingly to maintain a safe distance from the human and avoid collisions.

The real-world test was conducted, and the results can be accessed using this \href{https://youtu.be/1w-0Eyuem6w}{link}. The video demonstrates the effectiveness of the predictive collision avoidance algorithm in maintaining a safe distance from the human, adapting its velocity to the human's unpredictable stops and starts. This indicates that the algorithm is able to accurately predict the future positions of the human and adjust the ego robot's velocity accordingly to prevent collisions.

During real-world tests, the ego robot demonstrated smooth and stable operation, without abrupt stops or sudden changes in direction. This is a strong indication of the system's ability to handle real-world conditions and adapt to the uncertainties associated with real-world sensor data and the unpredictable behavior of humans.

The real-world test results for scenario one with a human in front provide valuable insights into the practicality and effectiveness of the proposed multi-object tracking and predictive collision avoidance system. The successful performance of the algorithm in a real-world environment with a human obstacle supports its potential to enhance the safety and efficiency of autonomous mobile robots operating in complex and dynamic environments.

%% file: sections/limits.tex
\section{Limitations and Future Work}
\label{sec:limits}
This thesis focused on multi-object tracking and predictive collision avoidance using local planning for autonomous systems operating in complex and dynamic environments. However, there are some limitations and areas for future work that could further enhance the effectiveness and applicability of the proposed methods.

One limitation of this study is the reliance on local planning, which may lead the robot to stray far from the objective or not follow an optimal path. In the future, the findings of this thesis can be extended to global planners, which would enable the robot to follow a more efficient trajectory toward its goal while still considering the dynamism of obstacles.

Another aspect that could be improved in future work is the exploration of other controllers besides the \glsxtrshort{dwa}, such as Model Predictive Controllers (MPC). Incorporating MPC would potentially result in smoother paths and more precise control for the robot, leading to overall improvements in navigation performance.

Additionally, this thesis focused primarily on the use of lidar sensors for multi-object tracking and predictive collision avoidance. Future work could explore the integration of additional sensor modalities, such as radar, ultrasonic sensors, or vision-based systems, to further enhance the robustness and performance of the proposed algorithms. Incorporating multiple sensor types could help mitigate the effects of sensor noise and improve the reliability of the system in diverse and challenging environments.

An improvement on the tracker can be achieved by dynamically updating the motion model of the robots from previous observations to include other models and not just only holonomic ones. This enhancement would allow the tracker to better adapt to different robot types and improve the overall tracking performance.

Furthermore, the developed methods could be applied and tested in a broader range of applications, including different types of autonomous vehicles, drones, and robots in various industrial, commercial, and public settings. This would help to validate the generalizability and adaptability of the proposed methods to different scenarios and use cases.

In summary, this thesis has laid a solid foundation for future research in multi-object tracking and predictive collision avoidance. By addressing the limitations and exploring the potential areas for future work mentioned above, it is expected that the safety and efficiency of autonomous systems in complex and dynamic environments can be further improved.

%% file: sections/conclusion.tex
\section{Conclusion}
\label{sec:conclusion}
This thesis presented a comprehensive study on multi-object tracking and predictive collision avoidance for autonomous systems operating in complex and dynamic environments. The primary focus was on industrial AGVs, such as forklifts, to ensure safety and efficiency in their operation. A thorough literature review was conducted, followed by the development and evaluation of novel algorithms for multi-object tracking and predictive collision avoidance.

The proposed methods were tested in both simulated and real-world scenarios, with the latter involving the use of lidar sensors in an industrial setting. The results demonstrated the effectiveness of the developed algorithms, even in the presence of increased noise from real-world lidar data. The ensemble Kalman filter was shown to have great performance, accuracy, and computational efficiency. Furthermore, the system demonstrated promising results in multi-robot environments with decentralized control and a fast controller time response, which is crucial for real-time applications. Different test results for Gazebo, Stage, and the real world can be found \href{https://www.youtube.com/playlist?list=PLuvKaBdrGFpR7MCaRiTQHtrFh83bqK-sK}{here.}

Despite the promising results, some limitations and challenges remain. The assumptions made regarding the dynamic behavior of objects and the challenges related to the implementation of the proposed algorithms on real-world systems need to be addressed in future work. Additionally, the use of only lidar sensors in this research may be expanded to include other types of sensors for improved performance and robustness.

In conclusion, this thesis has made significant contributions to the field of multi-object tracking and predictive collision avoidance, providing valuable insights and practical solutions for enhancing the safety and efficiency of autonomous mobile robots. Future research directions include refining the developed algorithms, exploring the use of additional sensor modalities, and further validating the effectiveness of the proposed methods in more diverse and challenging real-world scenarios.

%% file: sections/appendex.tex
\appendix
\section{\glsxtrshort{ros2} Graphs and Experimental Results }
\label{sec:appendix}
\begin{figure}[H]
    \centering
    \includegraphics[width=0.45\textwidth]{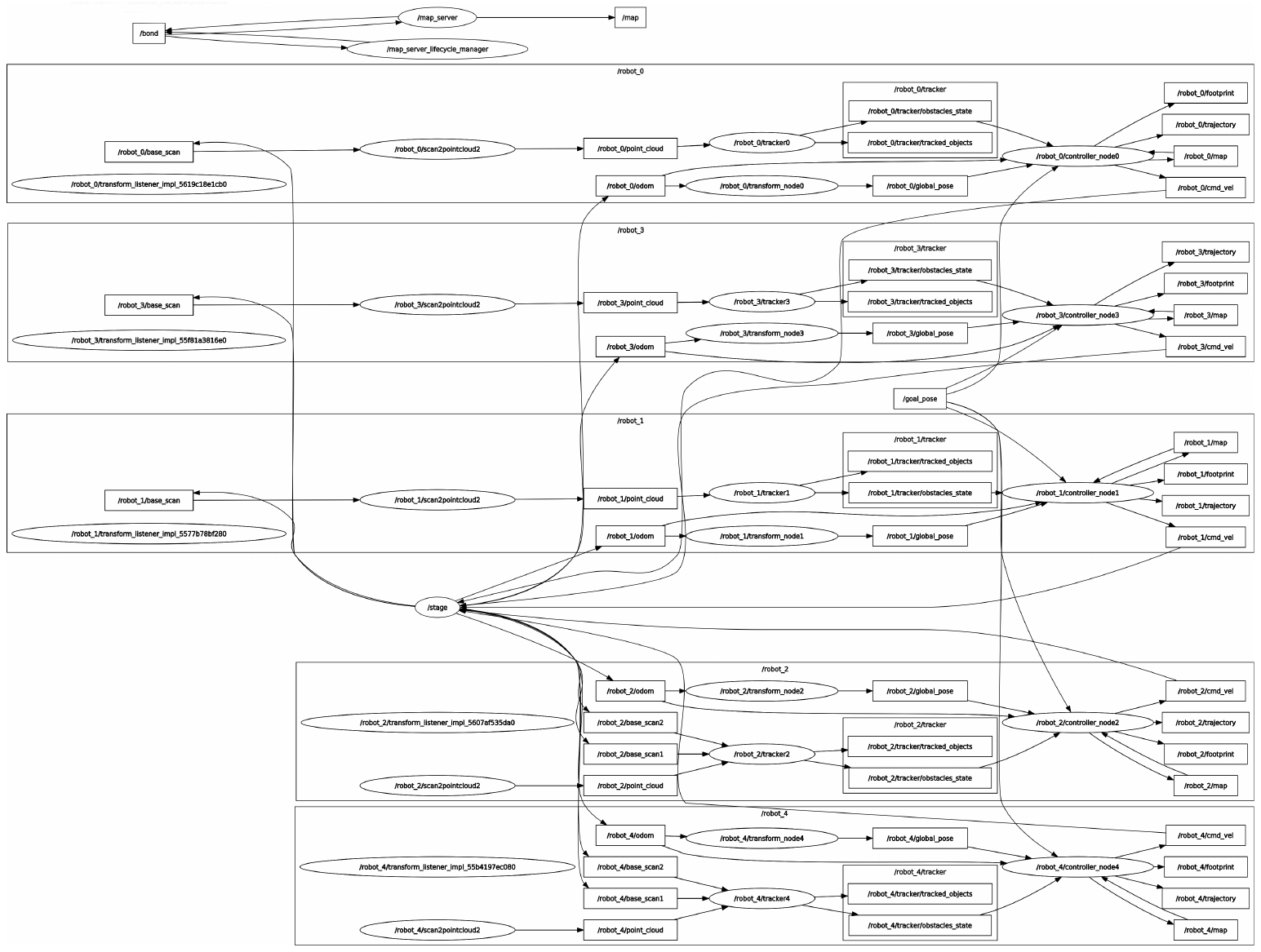}
    \caption{\textit{\glsxtrshort{ros2} graph for 5 robots in stage simulator using the implemented tracker and controller } }
    \label{fig:rosgraph}
\end{figure}

\begin{figure}[H]
    \centering
    \includegraphics[width=0.4\textwidth]{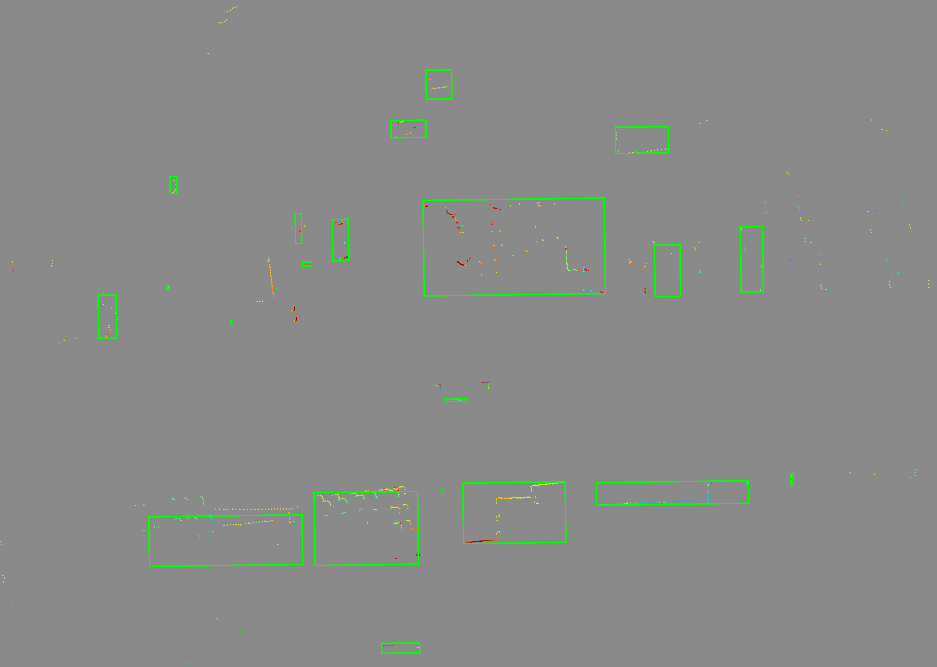}
    \caption{\textit{tracking result in an Industrial environment} }
    \label{fig:industrial_test}
\end{figure}

\begin{figure}[H]
    \centering
    \includegraphics[width=0.45\textwidth]{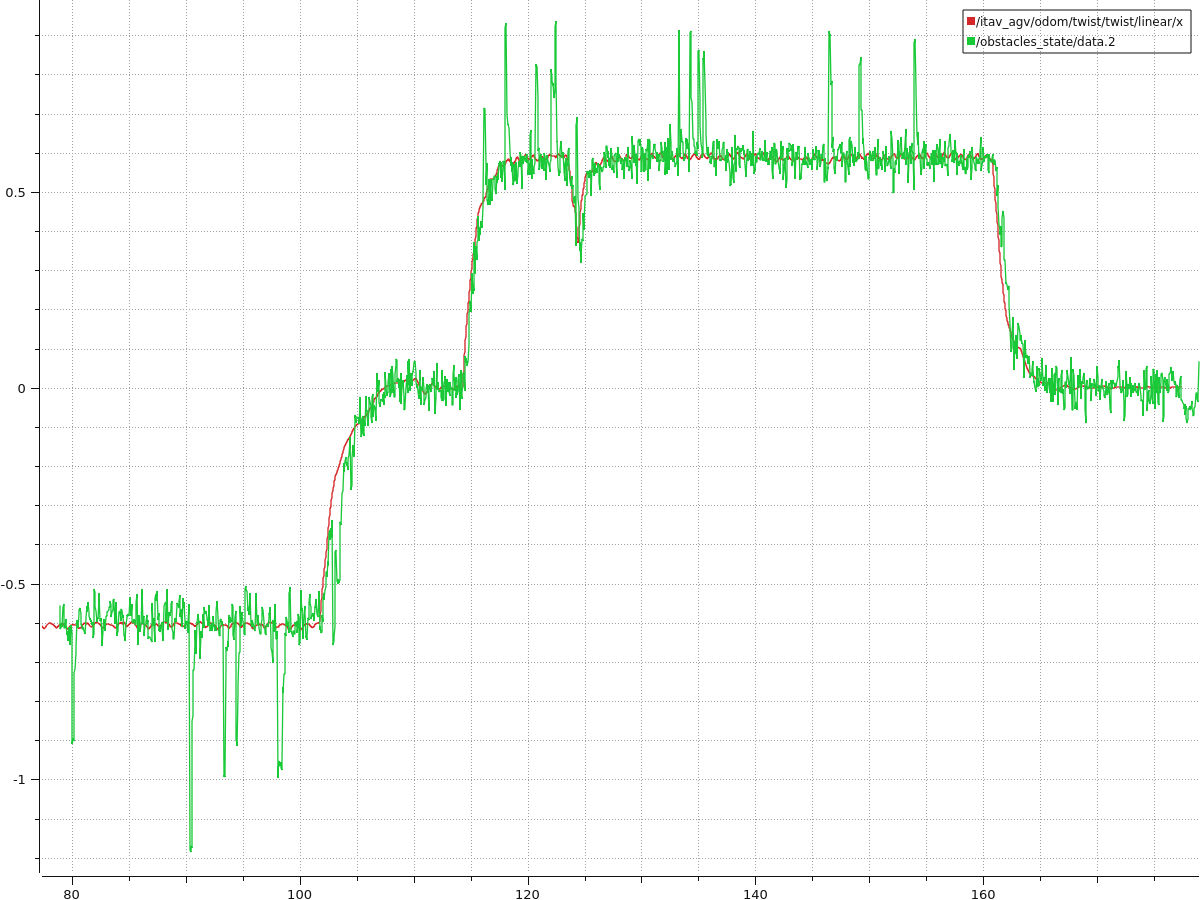}
    \caption{\textit{\glsxtrshort{enkf} velocity estimation without rolling mean average} }
    \label{fig:no_moving_average}
\end{figure}

    